\documentclass[11pt]{article}

\usepackage[T1]{fontenc}
\usepackage[utf8]{inputenc}
\usepackage[english]{babel}
\usepackage{microtype}

\usepackage{amsmath}
\usepackage{amssymb}
\usepackage{amsthm}

\usepackage{newtxtext}
\usepackage{newtxmath}

\usepackage{xcolor}

\usepackage{graphicx}
\graphicspath{{./}{./figures/}}

\usepackage{tikz}
\usetikzlibrary{shadows,shapes.geometric}

\usepackage{float}

\usepackage{pifont}

\usepackage{booktabs}
\usepackage{multirow}
\usepackage{array}

\usepackage{pdflscape}

\usepackage[ruled,vlined,linesnumbered]{algorithm2e}
\SetKwInput{KwInput}{Input}
\SetKwInput{KwOutput}{Output}

\usepackage[a4paper,margin=1in]{geometry}

\usepackage[section]{placeins}   
\usepackage{csquotes}

\usepackage{hyperref}

\usepackage[
  backend=biber,
  style=numeric,
  sorting=none,
  maxbibnames=99,
  giveninits=true,
]{biblatex}
\title{\textbf{The GEST-Engine: From Event Graphs to Synthetic Video}\\[4pt]
  \large A Full Technical Report}
\author{Nicolae Cudlenco$^{1,3}$ \and Mihai Masala$^{2}$ \and Marius Leordeanu$^{1,2}$\\[8pt]
  \small $^{1}$Institute of Mathematics of the Romanian Academy, Bucharest, Romania\\
  \small $^{2}$National University of Science and Technology Politehnica Bucharest, Romania\\
  \small $^{3}$B\"uchi Labortechnik AG, Flawil, Switzerland\\[4pt]
  \small \texttt{\{nicolae.cudlenco,leordeanu\}@gmail.com}, \texttt{mihaimasala@gmail.com}}
\date{}

\begin{document}

\maketitle

\begin{abstract}
We present the GEST-Engine, a complete system that goes from natural-language
text to fully-annotated multi-actor video across all modalities. At its core is
an \emph{explicit world model}: rather than encoding state as a learned latent
like implicit neural generators, the engine maintains a complete, inspectable
representation of the world --- which actors exist, where they are, what they
are doing, which objects they hold, and how events relate in time and space ---
expressed as a formal Graph of Events in Space and Time (GEST) and realized
deterministically inside the open world of a commercial game engine controlled
through an open-source multiplayer scripting framework. GESTs are produced
either procedurally or by an \emph{agentic} text-to-GEST system in which an LLM
Director plans a story through tool calls that a programmatic state backend
validates, so every generated specification is executable by construction. A
GEST then enters a four-stage execution pipeline --- graph parsing and
validation (episode selection via set-cover with backtracking), entity and
action grounding (a chain-ID system that binds abstract entities to concrete 3D
objects), temporal orchestration (Allen-style constraints resolved with a
Floyd--Warshall transitive closure), and execution and capture. During a single
simulation pass, a unified in-client capture path and a set of collectors emit
frame-aligned RGB video, dense per-pixel depth, texture-level instance
segmentation, per-actor eighteen-joint skeletal pose, per-frame pairwise
spatial-relation graphs, screen-space 2D bounding boxes, event-to-frame
temporal mappings, and natural-language descriptions --- all at zero marginal
annotation cost. We further describe an in-game world editor, a runtime
capability-extraction mechanism, a text-generation pipeline, and a production
system that renders corpora at scale across parallel virtual machines. Because
every generated frame traces back to a semantic specification, the engine
guarantees object permanence, multi-actor coordination, and temporal
consistency by construction --- properties that make its output valuable as
training data, evaluation benchmarks, and diagnostic tools for video
understanding.
\end{abstract}


\section{Related Work}
\label{sec:engine_related}

The GEST-Engine can be understood as an \emph{explicit world model}: a system that maintains a complete, inspectable representation of the state of the world --- which actors exist, where they are, what they are doing, what objects they hold, and how events relate to each other in time and space --- and uses this representation to generate visual observations deterministically. This positions it in contrast to the \emph{implicit world models} that have gained prominence in recent years, where the world state is encoded as a latent vector learned from data. Foundational work on latent dynamics models \cite{ha2018worldmodels}, DreamerV3 \cite{hafner2023dreamerv3}, and Genie \cite{bruce2024genie} has shown that neural networks can learn to predict how environments evolve over time, while large-scale video generation models such as Sora \cite{brooks2024sora} and VEO 3 \cite{veo3modelcard} produce photorealistic footage from text descriptions. However, as confirmed by a companion experimental evaluation, these implicit models struggle with object permanence, multi-actor coordination, and temporal consistency --- precisely the properties that an explicit world model guarantees by construction. The two paradigms are complementary \cite{liu2025worldsurvey,lecun2022path}: explicit models like ours generate the structured data that implicit models require for training and evaluation.

\textbf{Synthetic simulation platforms.} Several synthetic platforms have been developed for computer vision research, each targeting a specific domain. CARLA \cite{dosovitskiy2017carla} provides a rich driving simulator with LiDAR, depth, and semantic segmentation, but does not model human actors performing narrative activities. BEDLAM \cite{black2023bedlam} generates photorealistic body motion with perfect pose annotations and has demonstrated that purely synthetic training data can achieve state-of-the-art results on real-world human pose benchmarks --- an encouraging precedent for sim-to-real transfer. However, each BEDLAM sequence contains a single activity with no inter-actor coordination or event-level structure. AI2-THOR \cite{kolve2017ai2} provides an interactive 3D environment for embodied AI with rich object affordances, but focuses on navigation and manipulation tasks rather than multi-actor narrative generation.

VirtualHome \cite{puig2018virtualhome} is the most closely related system. It executes programs of household activities in a simulated environment, providing object state changes and action-level annotations. VirtualHome Action Genome \cite{qiu2023virtualhome} extends it by generating spatio-temporal scene graphs from simulations. However, both versions are limited in ways that our system is not. VirtualHome programs are single-actor sequential scripts without temporal orchestration across multiple actors. The environments are closed interiors --- there is no open world, no outdoor environments, and no ability to switch between contexts (e.g., going from inside a house to a garden and back). VirtualHome does not produce per-frame pairwise spatial relation graphs or event-to-frame temporal mappings.

\begin{table}[tb]
\centering
\caption{Simulation capacity comparison across synthetic platforms. For our system, we report both the currently configured subset and the full potential of the GTA San Andreas platform. AI2-THOR numbers include ProcTHOR extensions. VirtualHome numbers from v2.3 documentation.}
\label{tab:platform_comparison}
\begin{tabular}{@{}lccccc@{}}
\toprule
\textbf{Platform} & \textbf{Contexts} & \textbf{Actions} & \textbf{Objects} & \textbf{Actors} & \textbf{Multi-actor} \\
\midrule
CARLA \cite{dosovitskiy2017carla}           & 10 maps       & driving      & 70+ vehicles  & pedestrians & no      \\
BEDLAM \cite{black2023bedlam}               & Unreal scenes & motion cap.  & --            & 271 bodies  & no      \\
AI2-THOR \cite{kolve2017ai2}                & 120 rooms     & nav.+manip.  & 1{,}633       & 1 agent     & no      \\
VirtualHome \cite{puig2018virtualhome}      & 50 homes      & 42           & 357/home      & 1--2        & limited \\
\midrule
\textbf{Ours} (configured)                  & 11            & 46           & 45 types      & 312 skins   & yes     \\
\textbf{Ours} (platform)                    & 70+           & 2{,}500+ anim. & 733        & 312 skins   & yes     \\
\bottomrule
\end{tabular}
\end{table}

\textbf{Game engines for vision research.} Commercial game engines have been used as data sources for computer vision research. Richter et al.\ \cite{richter2016playing} extracted per-pixel semantic labels from Grand Theft Auto V, demonstrating that combining synthetic and real training data outperforms real data alone for semantic segmentation. This established the viability of game engines as annotation-free data sources, but captured only driving scenes without narrative progression or multi-actor coordination. The use of game engines extends beyond research: in professional film and television production, engines such as Unreal Engine 5 are already used for virtual cinematography and previsualization \cite{jia2025evolutionary}, suggesting that the gap between game-rendered and real footage will continue to narrow.

\textbf{Positioning of our system.} Our system operates within the open world of GTA San Andreas, which provides 733 objects across 32 categories, over 2,500 individual animations across 101 animation categories, 312 actor skins, and more than 70 contexts --- including houses, gyms, casinos, restaurants, gardens, forests, rivers, fields, and city streets. This means the system is not confined to a single interior: actors can move between indoor and outdoor environments within the same story, with the engine managing context switches transparently. The world editor tool (Section~\ref{sec:world_editor}) can be used to configure any location in the game as a simulation environment, and any game animation can be mapped to a GEST action, making the system extensible far beyond the environments we have mapped so far. Table~\ref{tab:platform_comparison} compares the simulation capacity of existing platforms with both the currently configured subset of our system and the full potential of the underlying game world. Beyond scale, the GEST-Engine supports multi-actor temporal coordination with explicit constraints (before, after, same\_time), produces dense multi-modal annotations at zero marginal cost, and takes a formal event graph as input --- making every generated video traceable back to its semantic specification. None of the existing platforms combines all of these properties within a single framework.

\section{Overview and Evolution}
\label{sec:engine_overview}

The idea of using a graph of events in space and time as a specification for video generation originated with the observation that if we could programmatically control every entity in a 3D world --- spawning actors, placing objects, triggering animations, and recording the results --- we could produce training data at a scale and with an annotation density that no manual labeling effort could match. The GEST formalism~\cite{masala2023gest} provided the formal framework; what remained was to build a system that could take such a graph as input and realize it as a video.

\subsection{The Platform: GTA San Andreas and Multi Theft Auto}

Building a virtual world from scratch requires the effort of a large team of software developers, extensive resources, and years of development. Instead, we chose to build on top of an existing game: Grand Theft Auto: San Andreas (2004), one of the most commercially successful video games of its era. To programmatically interact with the game's internals --- characters, objects, animations, camera, and world state --- we used Multi Theft Auto (MTA), an open-source multiplayer modification framework that exposes the game engine through a Lua scripting interface.

We chose this platform for several practical reasons. First, MTA is fully open-source (GPLv3), which meant we could modify and extend the engine without licensing constraints. Second, the hardware requirements are minimal --- the game runs on virtually any modern computer, including systems without dedicated GPUs, and multiple instances can be run in parallel on commodity hardware. This property proved essential for production-scale video generation (Section~\ref{sec:production}). Third, and most importantly for our purposes, the game world provides an extraordinarily rich set of pre-existing assets that would have taken years to create from scratch. Fourth, the MTA community has actively maintained the platform for over two decades, providing stable APIs, extensive documentation, and a mature ecosystem of tools.

\begin{figure}[htbp]
\centering
\includegraphics[width=\textwidth]{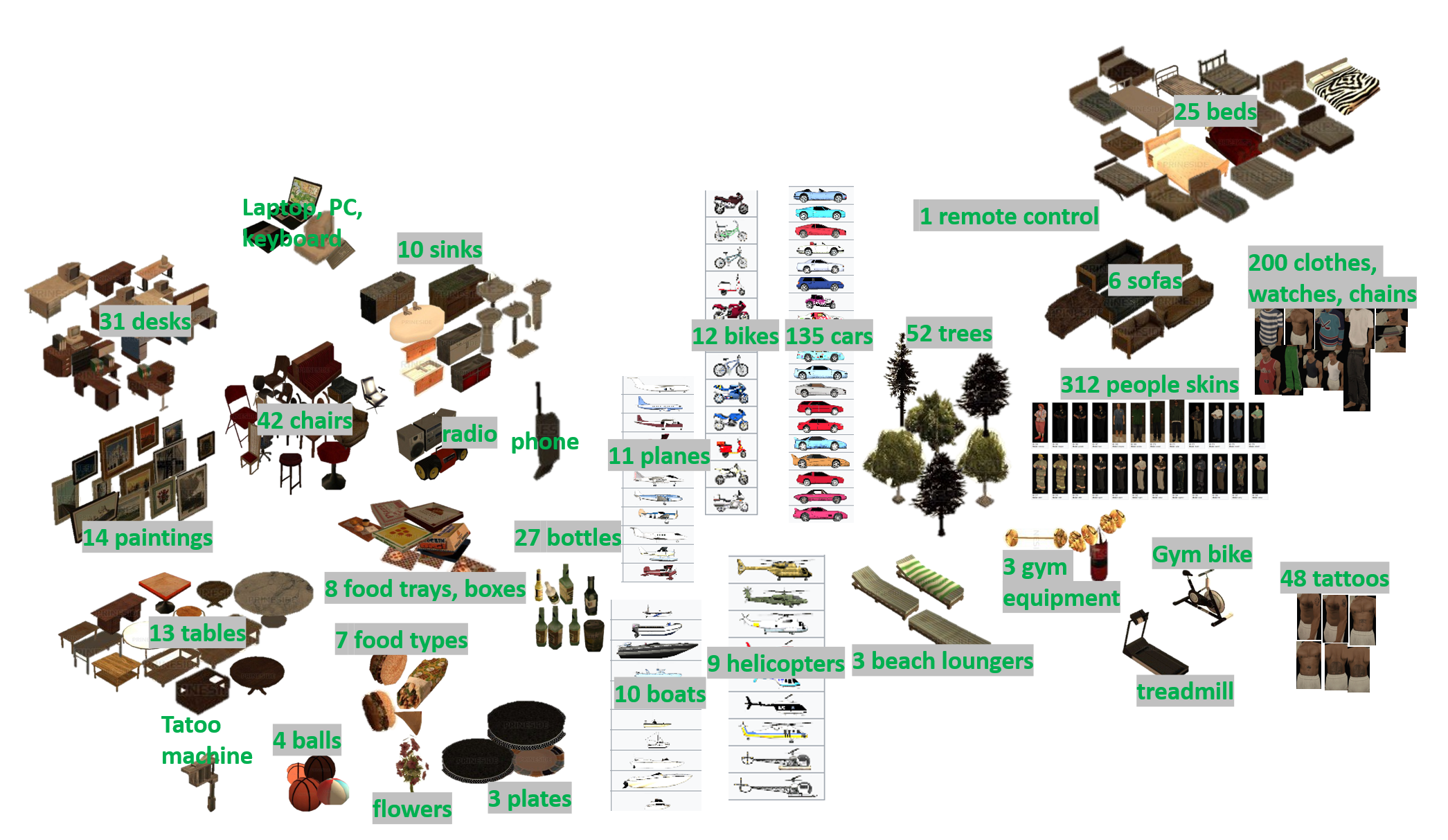}
\caption{\textbf{Available objects in GTA San Andreas via MTA.} A total of 733 object models clustered into 32 categories, ranging from household furniture and food to vehicles, vegetation, and gym equipment.}
\label{fig:mta_objects}
\end{figure}

GTA San Andreas is a vast and realistic open world. As shown in Figure~\ref{fig:mta_objects}, the game provides a total of 733 object models that can be clustered into 32 categories --- from furniture (desks, chairs, sofas, beds) and household items (food, drinks, laptops, phones) to vehicles (bikes, cars, boats, helicopters), vegetation (trees, flowers), and gym equipment (treadmills, barbells, punching bags). The animation system offers over 2,500 individual animations organized into 101 animation categories, covering everything from sitting and eating to dancing, exercising, smoking, and social interactions. The character system provides 312 actor skins representing people of different ages, genders, races, and clothing styles. The world itself contains more than 70 distinct contexts --- houses, apartments, gyms, casinos, restaurants, bars, nightclubs, offices, classrooms, gardens, beaches, forests, rivers, fields, and city streets --- each with its own geometry, lighting, and object placement.

\begin{figure}[htbp]
\centering
\includegraphics[width=\textwidth]{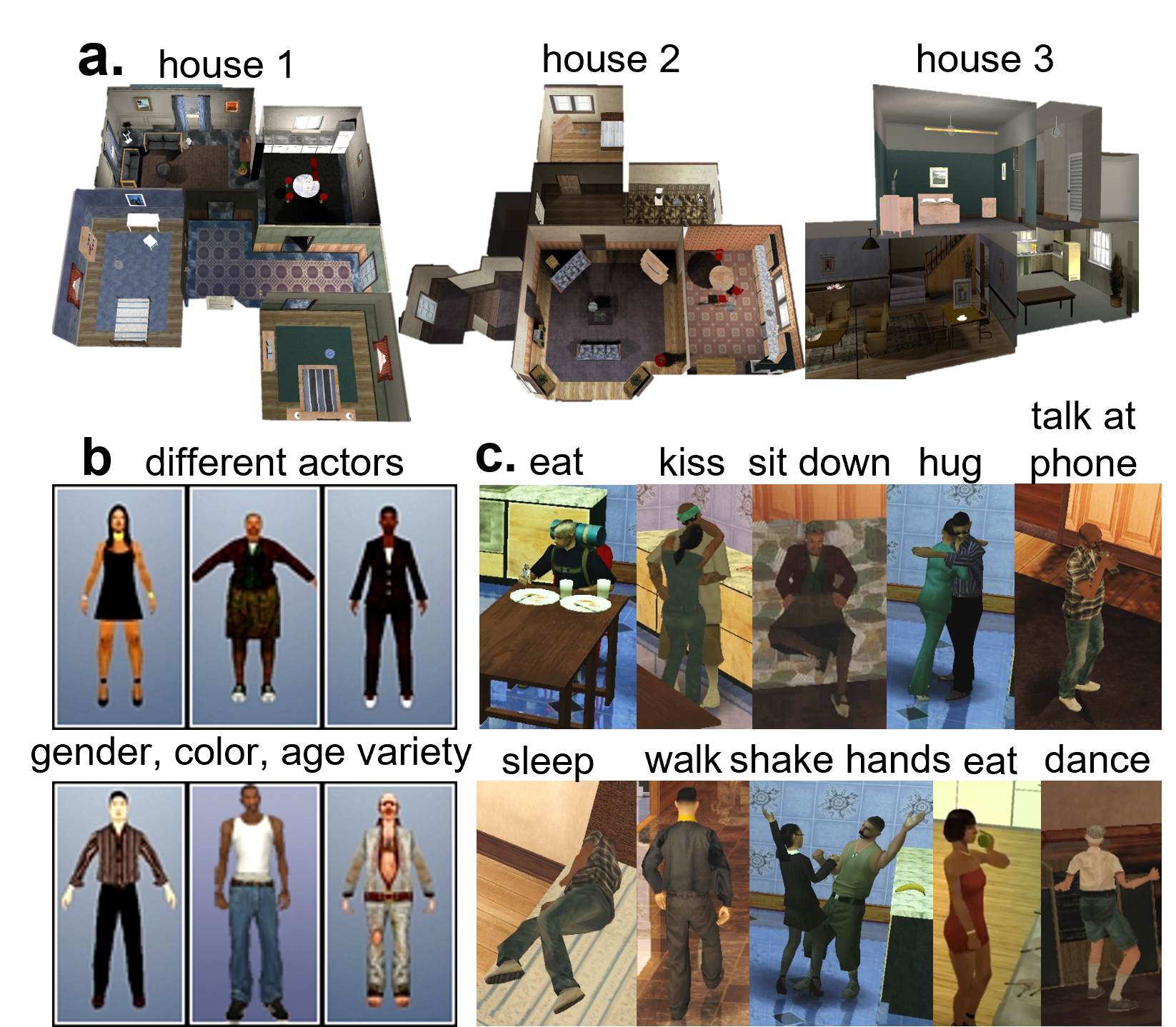}
\caption{\textbf{Examples of contexts, actors, and actions in MTA San Andreas.} (a) Sample contexts: residential interiors, gyms, gardens, and outdoor environments. (b) A selection of the 312 available actor skins. (c) Examples of actions available to actors in the simulation.}
\label{fig:mta_contexts}
\end{figure}

Figure~\ref{fig:mta_contexts} shows examples of contexts, actors, and actions available in the simulation environment. An important difference between GTA San Andreas and the original single-player game is that in MTA, the game's built-in AI is disabled --- there are no pedestrians walking around, no traffic, no ambient characters. Every entity in the simulation is placed and controlled programmatically by our system, which gives us complete knowledge of and authority over the state of the world at every moment. This is what enables the deterministic, annotation-free data generation that is the central capability of the engine.

Of the full set of available assets, our current system has mapped 11 environments with the World Editor tool (Section~\ref{sec:world_editor}): three residential houses, three gyms, a classroom, two offices, a garden, and a common area. We have implemented 46 granular action types across seven categories (movement, social, object interaction, consumption, physical activity, device, and personal) and configured 45 object types (12 spawnable, 28 fixed, 5 interactive). However, the World Editor can be used to configure any location in the game as a simulation environment, and any game animation can be mapped to a GEST action --- the system is designed to be extended incrementally as new environments and actions are needed.

\subsection{System Evolution}

The system evolved substantially over six years of continuous development, growing from a single-room prototype with one actor performing random actions to a full four-stage pipeline capable of orchestrating complex multi-actor narratives across multiple environments with dense multi-modal annotations. We summarize the major phases of this evolution here; the technical details of each component are presented in the subsequent sections.

The initial proof-of-concept (2020) consisted of hardcoded episodes where locations, objects, and action sequences were wired directly in Lua code. A single actor would be spawned in a house, move randomly between predefined points of interest, and perform whatever actions were available at each location. A Logger module recorded natural language descriptions of the actions as they occurred, producing paired video and text data. This early system already demonstrated the core insight: because we had full control over the simulation, every aspect of the generated video --- who did what, where, when, and with which objects --- was known by construction.

The first major architectural improvement came with the introduction of the World Editor (mid-2020), which replaced hardcoded episodes with data-driven JSON definitions. An in-game tool allowed us to walk through environments and define regions, points of interest, objects, actions, cameras, and pathfinding graphs interactively, saving the results to JSON files that could be loaded at runtime. This separated the content (what environments exist and what can happen in them) from the logic (how the simulation executes), making the system extensible without code changes.

In mid-2021, the direction of the graph flipped: instead of only logging events from random simulation, the system began loading predefined GESTs and executing them deterministically --- the transition from graph-as-output to graph-as-input described in Appendix~\ref{app:gest_json}. This was the pivotal moment that transformed the system from a random video generator into a controllable, specification-driven engine. Multi-actor interactions (handshakes, conversations, giving objects) followed, and in early 2022, cross-episode context switching was introduced, allowing actors to move between indoor and outdoor environments within a single story through a MetaEpisode wrapper that coordinates teleportation, camera fading, and action pausing across 3D contexts.

The introduction of formal temporal orchestration (2024--2025) through Allen's interval algebra \cite{allen1983maintaining} and Floyd-Warshall-based constraint satisfaction was the second major leap. The ActionsOrchestrator and EventPlanner replaced the earlier sequential execution model with a constraint-aware scheduler capable of coordinating concurrent multi-actor activities --- for instance, having one actor eat while another simultaneously talks on the phone in a different room, with explicit temporal ordering between their subsequent actions.

The multi-modal artifact collection pipeline (2025) added frame-aligned dense annotations --- RGB video via GPU-accelerated Desktop Duplication, instance segmentation via an HLSL shader with FNV-1a texture hashing, per-frame pairwise spatial relation graphs, event-to-frame temporal mappings, and natural language descriptions --- all captured deterministically at zero marginal cost during a single simulation pass. Production hardening (2025--2026) addressed chain-ID consistency across action sequences, race conditions in concurrent action dispatch, and scalable batch orchestration across 25 parallel virtual machines. An initial version of the engine was described in earlier work~\cite{masala2023gest}. The remainder of this report describes the full system.

The GEST-Engine represents a substantial engineering effort. The system comprises approximately 38,000 lines of Lua for the engine core, 8,500 lines of C++ for the native screenshot module and pathfinding, 53,000 lines of Python for the procedural generator and the agentic text-to-GEST system (Section~\ref{sec:agentic}), and 600 lines of JavaScript for the interactive GEST visualizer --- an Electron application built on Cytoscape.js that renders GEST specifications as interactive directed graphs for visual inspection of event nodes, temporal edges, and actor assignments --- totalling approximately 100,000 lines of code across all components. This entire codebase was developed by a single researcher over the course of the project. Our choice to build on an existing game rather than creating a virtual world from scratch proved critical to making this feasible: by leveraging the assets, rendering, physics, and animation systems already present in GTA San Andreas, we could focus our engineering effort entirely on the graph orchestration, temporal coordination, and annotation collection layers. For comparison, VirtualHome \cite{puig2018virtualhome} --- the most closely related system --- was developed by a team of seven researchers with nine contributors to the public codebase over a comparable timespan (2018--2024, based on public commit history), and despite this larger team, it remains limited to single-actor sequential programs in closed indoor environments without temporal orchestration or dense spatial annotations.

\subsection{Architecture Overview}

Figure~\ref{fig:engine_architecture} shows the architecture of the current system. A GEST specification in the JSON format described in Appendix~\ref{app:gest_json} enters the pipeline along with a configuration file that specifies which graphs to execute, whether artifact collection is enabled, and various debug flags. The pipeline consists of four stages:

\begin{landscape}
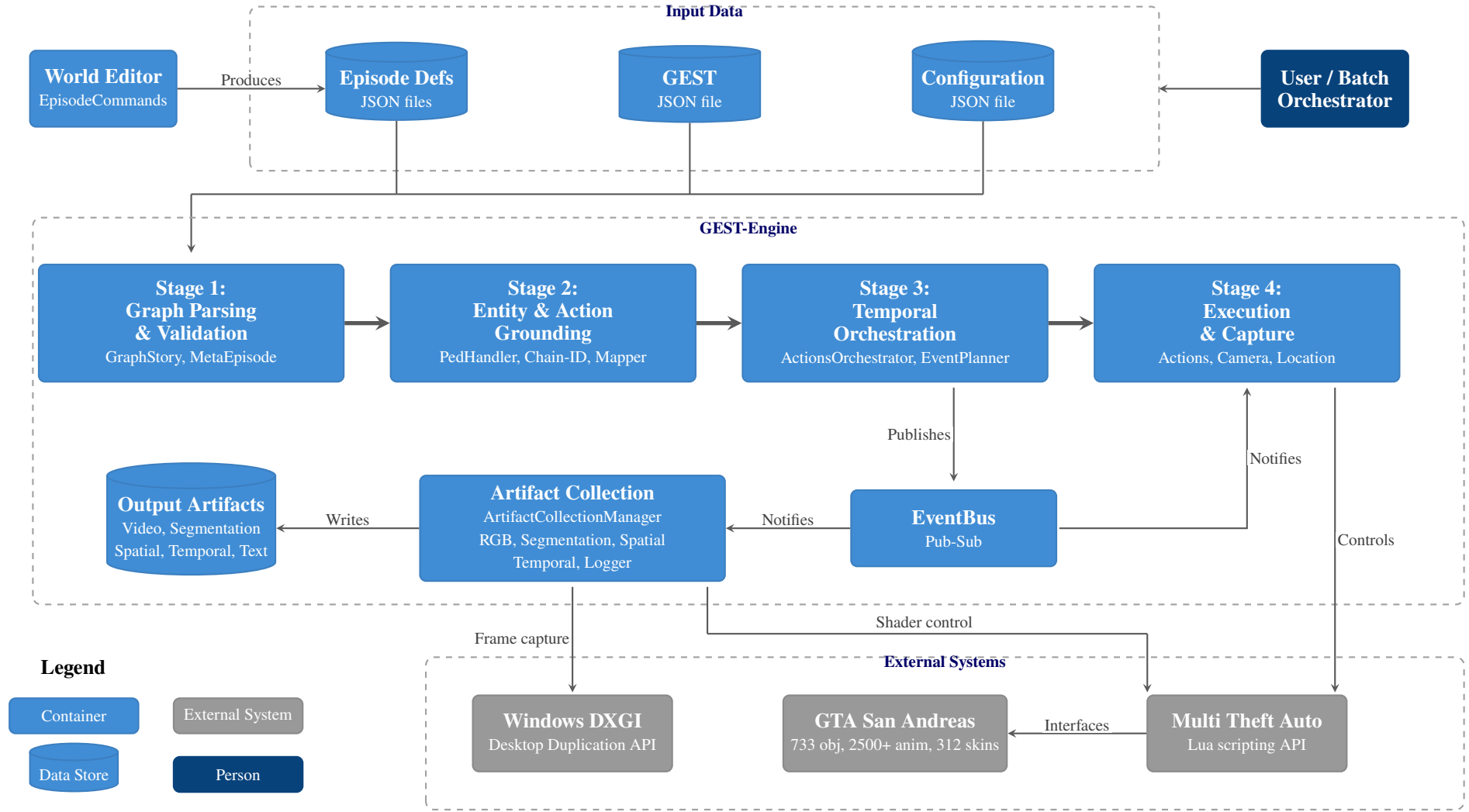
\begin{figure}[p]
\centering
\begin{tikzpicture}[
    container/.style={rectangle, rounded corners=3pt, draw={rgb,255:red,60;green,127;blue,192}, fill={rgb,255:red,64;green,141;blue,213},
                      text=white, thick, minimum width=3.6cm, minimum height=1.6cm,
                      align=center, font=\small},
    datastore/.style={cylinder, shape border rotate=90, draw={rgb,255:red,60;green,127;blue,192}, fill={rgb,255:red,64;green,141;blue,213},
                      text=white, thick, minimum width=2.4cm, minimum height=1.3cm,
                      aspect=0.15, align=center, font=\small},
    external/.style={rectangle, rounded corners=3pt, draw={rgb,255:red,138;green,138;blue,138}, fill={rgb,255:red,153;green,153;blue,153},
                     text=white, thick, minimum width=3.4cm, minimum height=1.3cm,
                     align=center, font=\small},
    tool/.style={rectangle, rounded corners=3pt, draw={rgb,255:red,60;green,127;blue,192}, fill={rgb,255:red,64;green,141;blue,213},
                 text=white, thick, minimum width=2.5cm, minimum height=1.3cm,
                 align=center, font=\small},
    output/.style={cylinder, shape border rotate=90, draw={rgb,255:red,60;green,127;blue,192}, fill={rgb,255:red,64;green,141;blue,213},
                   text=white, thick, minimum width=2.6cm, minimum height=1.4cm,
                   aspect=0.15, align=center, font=\small},
    person/.style={rectangle, rounded corners=3pt, draw={rgb,255:red,8;green,66;blue,123}, fill={rgb,255:red,8;green,66;blue,123},
                   text=white, thick, minimum width=2.5cm, minimum height=1.3cm,
                   align=center, font=\small},
    arr/.style={->, >=stealth, thick, color=gray!70!black},
    lbl/.style={font=\scriptsize, color=gray!50!black, fill=white, inner sep=1pt},
    boundary/.style={draw={rgb,255:red,161;green,161;blue,161}, dashed, rounded corners=4pt, thick},
]

\node[tool] (worldeditor) at (0.5, 9) {\textbf{World Editor}\\[-1pt]{\scriptsize EpisodeCommands}};
\node[datastore] (episodejson) at (5.5, 9) {\textbf{Episode Defs}\\[-1pt]{\scriptsize JSON files}};
\node[datastore] (gestjson) at (10.5, 9) {\textbf{GEST}\\[-1pt]{\scriptsize JSON file}};
\node[datastore] (config) at (15.5, 9) {\textbf{Configuration}\\[-1pt]{\scriptsize JSON file}};

\draw[boundary] (3, 7.6) rectangle (18.5, 10.4);
\node[font=\scriptsize\bfseries, color=blue!40!black] at (10.75, 10.3) {Input Data};

\node[person] (user) at (21.5, 9) {\textbf{User / Batch}\\[-1pt]\textbf{Orchestrator}};
\draw[arr] (user) -- (18.5, 9);

\draw[arr] (worldeditor) -- (episodejson) node[lbl, midway, above] {Produces};

\draw[thick, color=gray!70!black] (episodejson.south) -- (5.5, 7.2);
\draw[thick, color=gray!70!black] (gestjson.south) -- (10.5, 7.2);
\draw[thick, color=gray!70!black] (config.south) -- (15.5, 7.2);
\draw[thick, color=gray!70!black] (5.5, 7.2) -- (15.5, 7.2);
\draw[thick, color=gray!70!black] (2.0, 7.2) -- (5.5, 7.2);
\draw[arr] (2.0, 7.2) -- (2.0, 6.2);

\draw[boundary] (-0.7, 0.2) rectangle (23.7, 6.8);
\node[font=\scriptsize\bfseries, color=blue!40!black] at (11.5, 6.6) {GEST-Engine};

\node[container, minimum width=5.2cm, minimum height=2cm] (stage1) at (2, 5) {\textbf{Stage 1:}\\[-1pt]\textbf{Graph Parsing}\\[-1pt]\textbf{\& Validation}\\[-1pt]{\scriptsize GraphStory, MetaEpisode}};
\node[container, minimum width=5.2cm, minimum height=2cm] (stage2) at (8, 5) {\textbf{Stage 2:}\\[-1pt]\textbf{Entity \& Action}\\[-1pt]\textbf{Grounding}\\[-1pt]{\scriptsize PedHandler, Chain-ID, Mapper}};
\node[container, minimum width=5.2cm, minimum height=2cm] (stage3) at (14, 5) {\textbf{Stage 3:}\\[-1pt]\textbf{Temporal}\\[-1pt]\textbf{Orchestration}\\[-1pt]{\scriptsize ActionsOrchestrator, EventPlanner}};
\node[container, minimum width=5.2cm, minimum height=2cm] (stage4) at (20, 5) {\textbf{Stage 4:}\\[-1pt]\textbf{Execution}\\[-1pt]\textbf{\& Capture}\\[-1pt]{\scriptsize Actions, Camera, Location}};

\draw[arr, line width=2pt] (stage1) -- (stage2);
\draw[arr, line width=2pt] (stage2) -- (stage3);
\draw[arr, line width=2pt] (stage3) -- (stage4);

\node[output, minimum width=2.8cm] (outartifacts) at (2, 1.5) {\textbf{Output Artifacts}\\[-1pt]{\scriptsize Video, Segmentation}\\[-1pt]{\scriptsize Spatial, Temporal, Text}};
\node[container, minimum width=5.2cm, minimum height=1.8cm] (artifacts) at (8.5, 1.5) {\textbf{Artifact Collection}\\[-1pt]{\scriptsize ArtifactCollectionManager}\\[-1pt]{\scriptsize RGB, Segmentation, Spatial}\\[-1pt]{\scriptsize Temporal, Logger}};
\node[container, minimum width=3.5cm, minimum height=1.3cm] (eventbus) at (15, 1.5) {\textbf{EventBus}\\[-1pt]{\scriptsize Pub-Sub}};

\draw[arr] (15, 3.9) -- (15, 2.3) node[lbl, midway, left] {Publishes};
\draw[arr] (16.8, 1.5) -- (20, 1.5) -- (20, 3.9) node[lbl, midway, right] {Notifies};
\draw[arr] (eventbus.west) -- (artifacts.east) node[lbl, midway, above] {Notifies};
\draw[arr] (artifacts) -- (outartifacts) node[lbl, midway, above] {Writes};

\node[external] (dxgi) at (8.5, -2) {\textbf{Windows DXGI}\\[-1pt]{\scriptsize Desktop Duplication API}};
\node[external] (gtasa) at (14, -2) {\textbf{GTA San Andreas}\\[-1pt]{\scriptsize 733 obj, 2500+ anim, 312 skins}};
\node[external] (mta) at (20, -2) {\textbf{Multi Theft Auto}\\[-1pt]{\scriptsize Lua scripting API}};

\draw[boundary] (6.0, -3.3) rectangle (23.7, -0.7);
\node[font=\scriptsize\bfseries, color=blue!40!black] at (14.85, -0.8) {External Systems};

\draw[arr] (21.5, 3.9) -- (21.5, -1.3) node[lbl, midway, right] {Controls};

\draw[arr] (8.5, 0.5) -- (8.5, -1.3) node[lbl, midway, left] {Frame capture};

\draw[arr] (mta) -- (gtasa) node[lbl, midway, above] {Interfaces};

\draw[thick, color=gray!70!black] (10.8, 0.5) -- (10.8, -0.3);
\draw[thick, color=gray!70!black] (10.8, -0.3) -- (18.3, -0.3);
\draw[arr] (18.3, -0.3) -- (18.3, -1.3);
\node[lbl] at (14.5, -0.1) {Shader control};

\node[font=\small\bfseries, anchor=west] at (-0.7, -0.9) {Legend};
\node[container, minimum width=2.2cm, minimum height=0.6cm, font=\scriptsize] at (0, -1.7) {Container};
\node[external, minimum width=2.2cm, minimum height=0.6cm, font=\scriptsize] at (2.8, -1.7) {External System};
\node[datastore, minimum width=1.5cm, minimum height=0.6cm, font=\scriptsize, aspect=0.2] at (0, -2.7) {Data Store};
\node[person, minimum width=2.2cm, minimum height=0.6cm, font=\scriptsize] at (2.8, -2.7) {Person};

\end{tikzpicture}
\caption{\textbf{System architecture of the GEST-Engine.} Input data feeds into the four-stage pipeline: \textbf{Stage~1} (Section~\ref{sec:stage_parsing}) parses and validates the GEST, \textbf{Stage~2} (Section~\ref{sec:stage_grounding}) grounds actors, objects, and actions to the 3D world, \textbf{Stage~3} (Section~\ref{sec:stage_orchestration}) orchestrates multi-actor execution and publishes events, \textbf{Stage~4} (Section~\ref{sec:stage_execution}) executes actions via Multi Theft Auto. The EventBus notifies both Stage~4 and the Artifact Collection system (Section~\ref{sec:artifact_collection}), which captures dense multi-modal annotations.}
\label{fig:engine_architecture}
\end{figure}
\end{landscape}

\begin{enumerate}
    \item \textbf{Graph Parsing and Validation} (Section~\ref{sec:stage_parsing}): the input GEST is parsed, its requirements (actors, objects, actions, locations) are extracted, and a valid set of episodes is selected through a set-cover algorithm with backtracking. For multi-location stories, a MetaEpisode wrapper aggregates multiple episodes and generates cross-episode movement actions.

    \item \textbf{Entity \& Action Grounding} (Section~\ref{sec:stage_grounding}): actors are created with gender-appropriate skins (random or specified) and assigned names --- either provided in the GEST or drawn randomly from a pool of approximately 1,000 male, 1,000 female first names, and 1,350 family names. Every required object and action sequence is mapped to all valid POI chains in the selected episodes through a chain-ID system, and actors are placed at their first event's location. The output is a set of one-to-many bidirectional maps that enable deferred runtime selection among valid alternatives.

    \item \textbf{Temporal Orchestration} (Section~\ref{sec:stage_orchestration}): the ActionsOrchestrator and EventPlanner coordinate multi-actor execution. A happens-before graph is constructed from the temporal constraints, Floyd-Warshall transitive closure derives a total ordering, and events are partitioned into temporal segments that execute concurrently within each segment and sequentially between segments.

    \item \textbf{Execution and Capture} (Section~\ref{sec:stage_execution}): actions are dispatched to actor handlers in the 3D environment. The PedHandler manages actor lifecycle, 55 action classes implement the full action taxonomy, the location system manages spatial semantics and POI selection, and the camera system provides automated cinematography with focus-based actor tracking.
\end{enumerate}

An EventBus decouples the pipeline stages from the artifact collection system through a publish-subscribe mechanism. Stage~3 publishes events (such as graph event start and graph event end signals) to the EventBus as it orchestrates the simulation. Two components subscribe to these events: Stage~4 (Execution \& Capture), which uses them for synchronization and camera control, and the Artifact Collection system (Section~\ref{sec:artifact_collection}), which uses them to capture frame-aligned annotations --- RGB video via GPU-accelerated Desktop Duplication, instance segmentation via an HLSL shader with FNV-1a texture hashing, per-frame pairwise spatial relation graphs, event-to-frame temporal mappings, and natural language descriptions --- all produced deterministically and simultaneously during a single simulation pass.

A deliberate architectural decision throughout the system is the progressive separation of game-agnostic logic from engine-specific bindings through an adapter pattern. The artifact collection subsystem already implements this cleanly: the Artifact\-Collection\-Manager, collection orchestration logic, and collector interfaces are fully game-agnostic, while MTA-specific implementations (e.g., MTA\-Native\-Screenshot\-Adapter, MTA\-Simulation\-Freeze\-Adapter, MTA\-Render\-Mode\-Controller) are isolated in dedicated adapter layers. The native C++ screenshot module similarly separates a reusable core (video encoding, frame buffer management, modality management) from the MTA-specific Desktop Duplication backend. The remaining parts of the system --- particularly the action classes, location system, and actor handlers --- still contain MTA-specific code interleaved with the orchestration logic. Completing the adapter pattern across all components is an ongoing effort; once finished, porting the system to a different 3D platform --- such as GTA V via FiveM, or a modern engine like Unreal or Unity --- would require implementing only the adapter layer for the new platform, without modifying the orchestration, planning, or collection logic. We discuss this further in Sections~\ref{sec:engineering} and~\ref{sec:future_gtav}.

The following sections describe each component in detail, proceeding through the pipeline in order.

\section{The World Editor}
\label{sec:world_editor}

One limitation of GTA San Andreas via MTA is that while we have programmatic access to the game's objects, animations, and characters, we do not have access to the geometry of the environment --- there is no API to query the layout of rooms, the positions of furniture, or the walkable paths through a building. This means we cannot procedurally generate simulation environments without additional data. To address this, we developed a World Editor: an in-game tool that allows us to walk through any environment in the game and interactively define all the spatial and semantic information needed for simulation.

The World Editor is implemented as a set of command-line commands available within the running MTA client. An operator walks through the game world and defines the following elements, which are serialized to a JSON file that can be loaded at runtime by the simulation engine:

\textbf{Regions.} A region is a polygon that represents the spatial boundary of a well-defined area --- for instance, a bedroom, a kitchen, a garden, or a hallway. Regions are defined by placing vertices at the corners of the area. A region corresponds to a location in the GEST, and the system ensures that actors are physically present in the correct region when executing actions associated with that location. Regions also determine which Points of Interest and objects belong to which spatial area, enabling location-aware action selection and providing spatial semantics for the procedurally generated text descriptions of the Logger. When an actor enters a region during simulation, the Logger automatically describes the room and its contents. Cameras are also defined within regions, and the camera system uses this information to select appropriate viewpoints and track actors as they move between rooms.

\textbf{Points of Interest (POIs).} A POI is a specific location within a region where an actor can perform actions. Each POI stores the precise 3D position and initial rotation needed for correct animation rendering --- for instance, to sleep on a bed, the actor must be positioned next to the bed on the correct side, facing the right direction. A POI can be associated with one or more actions, each optionally targeting a specific object. Actions at a POI can be defined with a specific chronological order (e.g., ``sit down'' must precede ``stand up''), or they can be chosen randomly from the available set (e.g., ``smoke'' or ``answer phone'' at a general-purpose POI). There are also special marks for actions: ``mandatory'' indicates an action that must be performed first when the POI is visited, and ``closed by'' indicates the final action in a group. A POI can be marked as ``interactions only'', meaning it is reserved exclusively for two-actor interactions (handshakes, conversations, giving objects) and will not be selected for solo actions. Some POIs contain a list of ``linked episodes'' --- other episodes that have a corresponding POI in the same location, allowing the system to switch between contexts (e.g., going from inside a house to the garden and back will route actors through the door in the hallway in the house interior and spawn them at the POI next to the door in the garden) while maintaining spatial consistency. During simulation, the location system uses the POI definitions to select where actors should go when executing actions associated with that location, and the pathfinding system computes routes to these POIs based on the pathfinding graph defined in the World Editor.

\begin{figure}[htbp]
\centering
\includegraphics[width=0.49\textwidth]{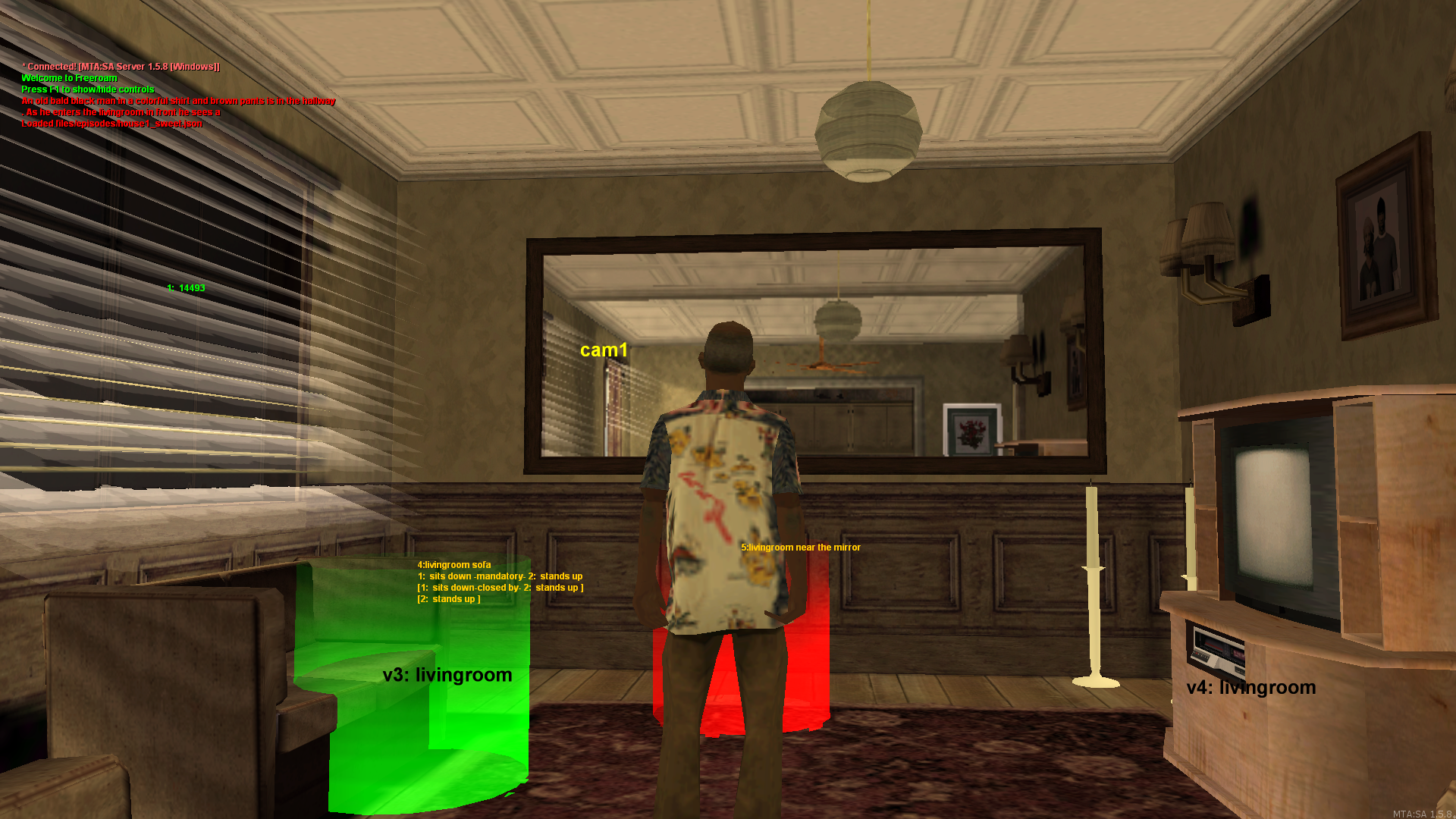}
\includegraphics[width=0.49\textwidth]{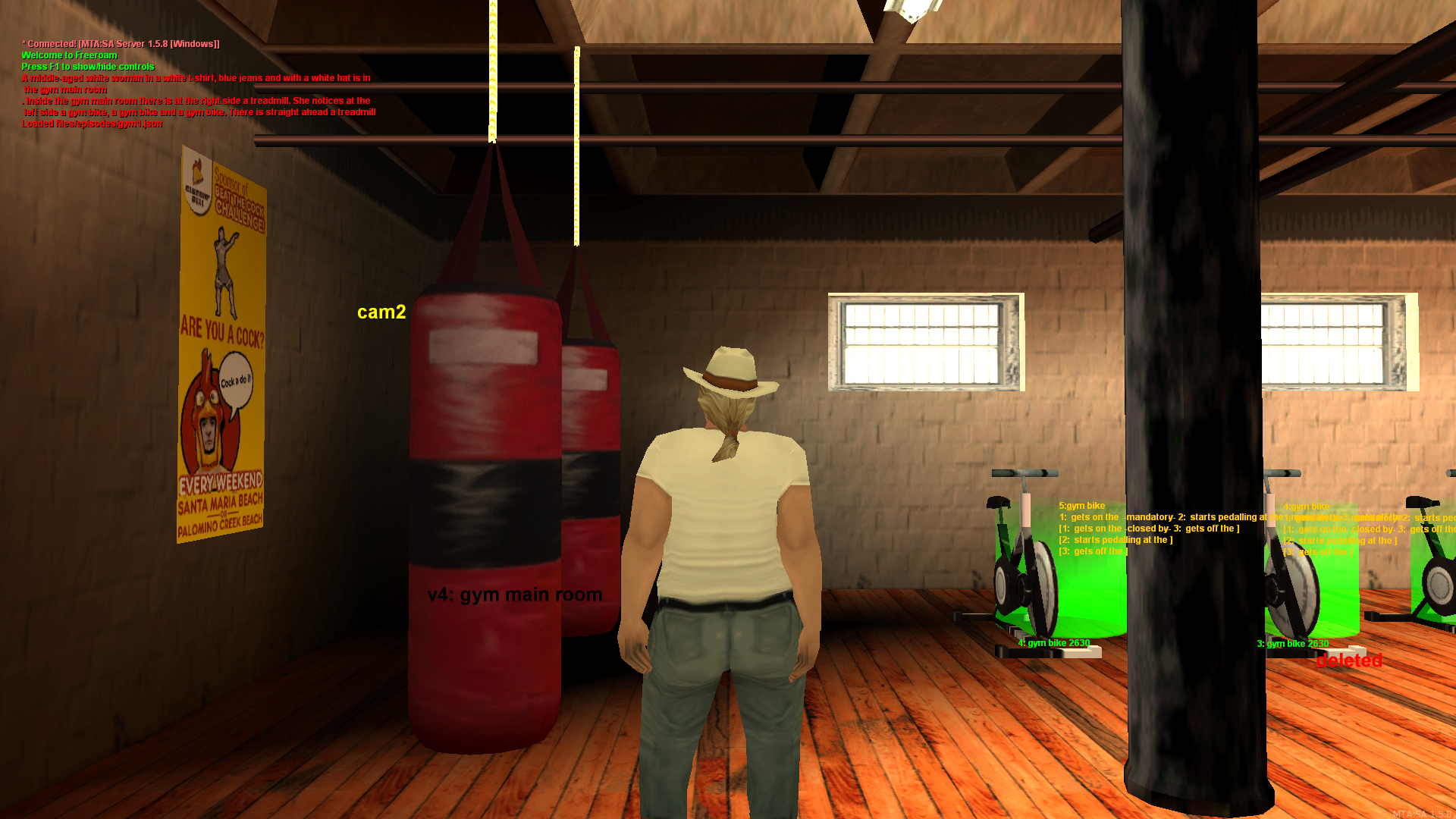}
\caption{\textbf{The World Editor in use.} Left: a living room episode showing POIs (green and red cylinders), region vertices (v3, v4 labeled ``livingroom''), a camera position (cam1), and action lists attached to each POI. Right: a gym episode showing POIs for gym bikes and punching bags, with their respective action chains listed. Green POIs have at least one action defined; red POIs have none.}
\label{fig:world_editor}
\end{figure}

\textbf{Templates.} We observed that in general, only specific sets of actions can be performed with specific objects --- a chair affords sitting, a bed affords sleeping, a phone affords calling. We formalized this by introducing Templates: reusable bundles that capture a POI together with all its associated actions, target objects, and spatial relationships. Standing at a POI and running the template save command serializes everything into a portable JSON definition. Templates make content reusable across episodes: once a ``desk with laptop'' template is defined (with actions for sitting, opening the laptop, typing, closing the laptop, and standing up), it can be placed in any episode that has a suitable desk.

\begin{figure}[htbp]
\centering
\includegraphics[width=0.49\textwidth]{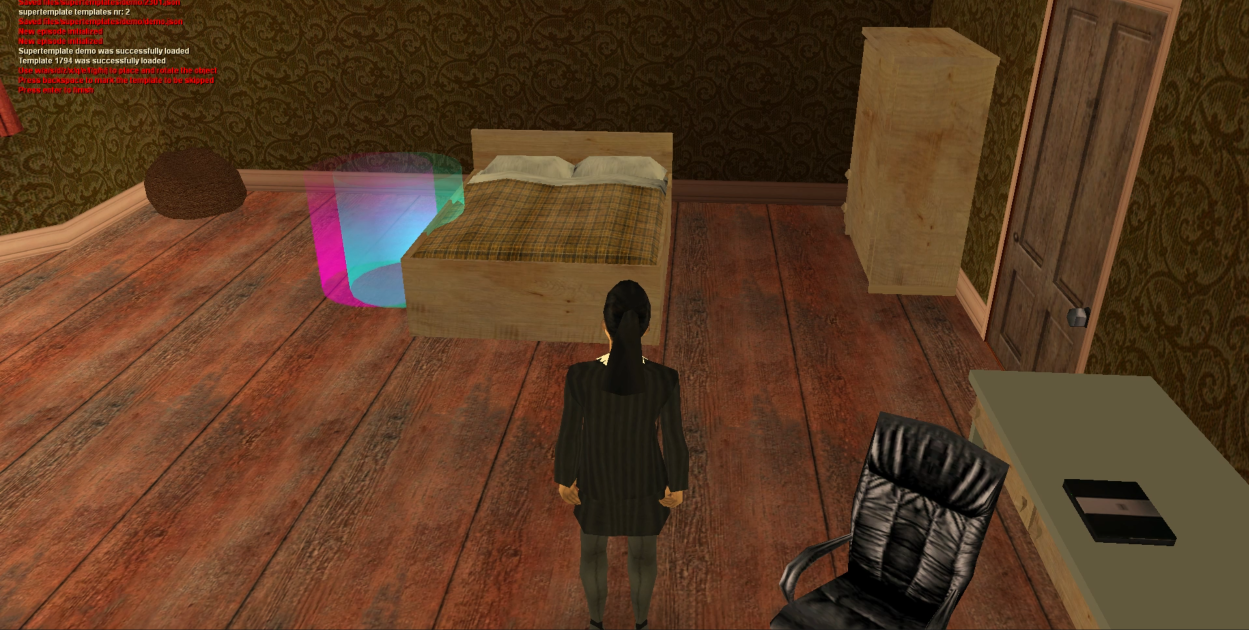}
\includegraphics[width=0.49\textwidth]{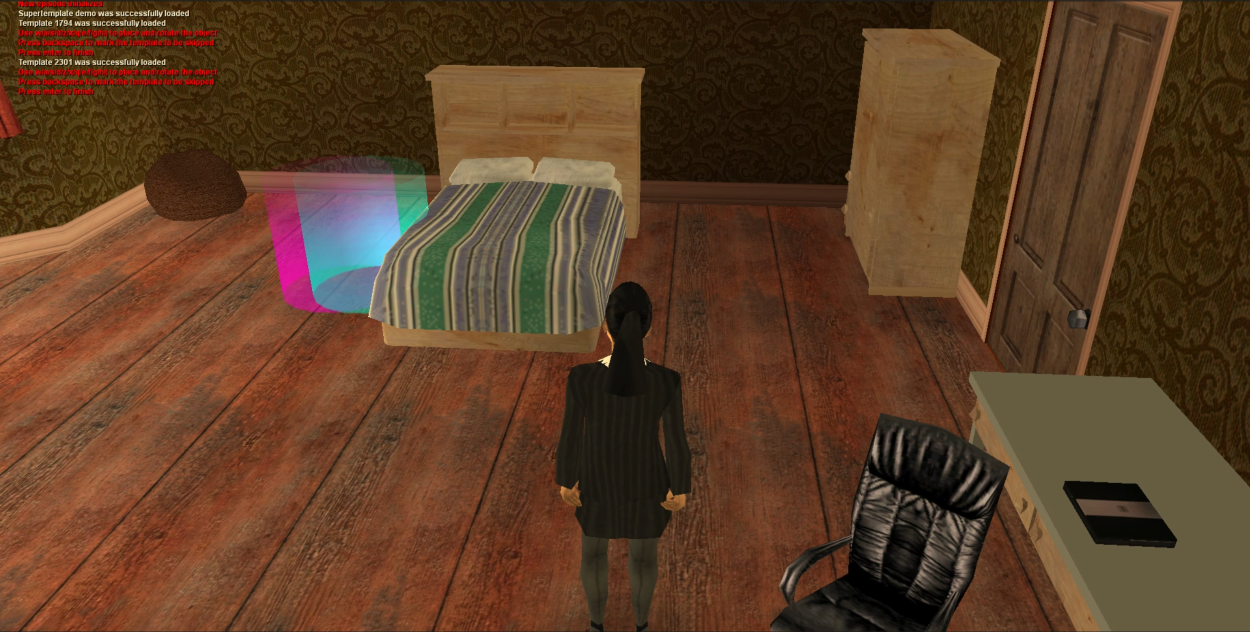}
\caption{\textbf{Supertemplate variability.} Two templates from the same Bed supertemplate, shown in the same bedroom. The POI position (colored cube on the left side of the bed) is identical, but the bed model differs --- producing visually distinct scenes from the same GEST specification. The room layout, furniture placement, and all other objects remain unchanged.}
\label{fig:supertemplate_beds}
\end{figure}

\textbf{Supertemplates.} To introduce visual variability, we defined Supertemplates: collections of Templates that contain objects from the same semantic category but with different visual appearances. For example, a ``Chair'' Supertemplate might group three different chair models (an office chair, a wooden chair, and an armchair), each with its own POI configuration and spatial offsets. When a GEST references a ``Chair,'' the system randomly selects one of the Supertemplate's variants at runtime. This means that no two executions of the same GEST need look identical --- different chair models, different sofa styles, different table configurations can appear each time, while the semantic content (``the actor sits on a chair'') remains the same. Figure~\ref{fig:supertemplate_beds} illustrates this with a Bed supertemplate: the same bedroom, the same POI position (shown as the colored cylinder), but two different bed models selected from the supertemplate --- producing visually distinct videos from the same GEST specification.

\begin{figure}[htbp]
\centering
\includegraphics[width=\textwidth]{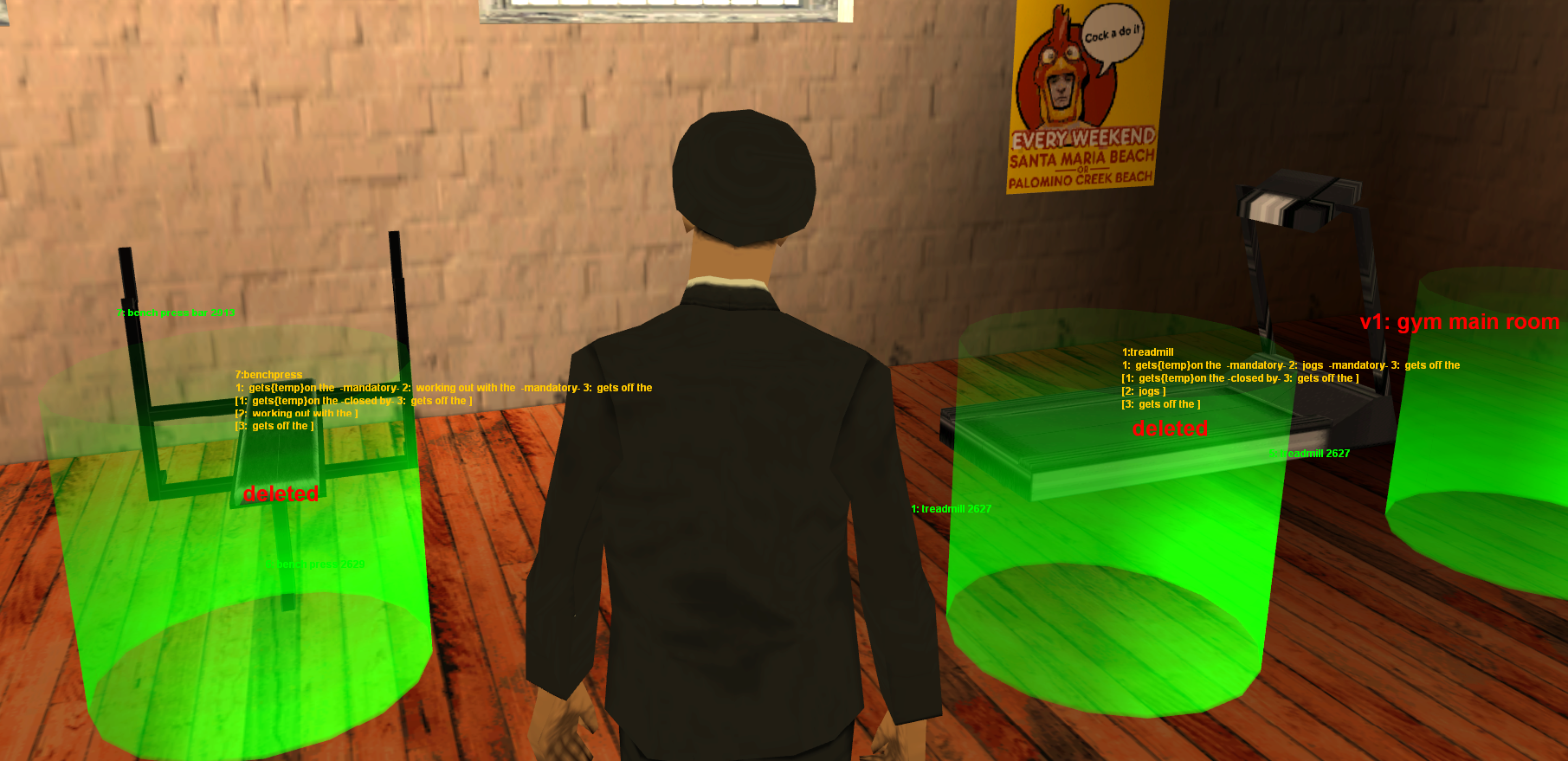}
\caption{\textbf{Templates in the World Editor.} A gym environment showing three POI templates (green cylinders): gym bikes with action chains listed (``gets on, starts pedaling, gets off'') and associated objects. Each template bundles a POI, its actions, and target objects into a reusable unit.}
\label{fig:world_editor_templates}
\end{figure}

\textbf{Cameras.} Static camera positions are placed within regions to cover key areas of the environment, similar to surveillance cameras. During simulation, the camera system (Section~\ref{sec:camera_system}) uses these positions for tracking actors and switching between viewpoints.

\textbf{Pathfinding graphs.} Since MTA does not expose the game's navigation mesh, we define a separate pathfinding graph for each episode: a set of nodes at walkable positions connected by edges along valid paths. The A* algorithm, provided by the \texttt{ml\_pathfind} module, uses this graph to compute routes for actors moving between POIs. The pathfinding graph is distinct from the GEST event graph --- it defines where actors \emph{can} walk, not what they \emph{should} do.

\textbf{Episodes and contexts.} An episode is the complete description of a simulation environment: a collection of regions, POIs (with their templates), cameras, pathfinding graphs, and a list of pre-existing game objects that need to be deleted or moved from the initial state of the game world. Episodes are stored as JSON files and loaded dynamically at runtime. For stories that span multiple contexts, episodes can be optionally linked together through connection points, forming a MetaEpisode (Section~\ref{sec:stage_parsing}). When the graph specifies unkinked episodes, the system still allows this and teleports actors (e.g. without routing them through connection POIs), mimicking different scenes in a movie that take place in different locations.

Table~\ref{tab:environment_types} summarizes the environment types currently configured in the system. Each environment type may have multiple variations (e.g., three different house interiors), each defined as a separate episode with its own layout, objects, and POI placement.

\begin{table}[tb]
\centering
\caption{Environment types configured in the GEST-Engine and their characteristics.}
\label{tab:environment_types}
\small
\begin{tabular}{@{}llcp{7.5cm}@{}}
\toprule
\textbf{Type} & \textbf{Setting} & \textbf{Var.} & \textbf{Rooms and Available Actions} \\
\midrule
House & Interior & 3 & Living room (sofas, music player, table), kitchen (counter, sink, table), bedroom (bed), bathroom (sink), hallway. Supports sitting, eating, drinking, dancing, sleeping, smoking, phone use, laptop use, and social interactions. \\
\midrule
Classroom & Interior & 1 & Desks with chairs, whiteboard area. Supports sitting, reading, typing, drinking, eating, smoking, phone use, and social interactions. \\
\midrule
Office & Interior & 2 & Desks, chairs. Supports smoking, phone use, and social interactions. \\
\midrule
Gym & Interior & 3 & Treadmills, stationary bikes, bench presses, punching bags, dumbbells, tai chi area. Supports all physical activity actions, social interactions, smoking, and phone use. \\
\midrule
Garden & Exterior & 1 & Porch, garden area, driveway, street. Supports tai chi, smoking, phone use, and social interactions. \\
\midrule
Common & Interior & 1 & Shared living room across interiors. Supports smoking, phone use, and social interactions. \\
\bottomrule
\end{tabular}
\end{table}

The World Editor proved essential to the project's scalability. Defining a new episode takes approximately one to two hours of interactive work within the game, depending on complexity, after which the environment is permanently available for simulation without any code changes. The template and supertemplate systems further accelerate this process: once a template for a common object (e.g., a dining table with food) is defined, it can be reused across any number of episodes. As of the current system, 11 episodes have been defined across 6 environment types, with 20 supertemplates containing a total of 124 templates --- including 20 bed variants, 19 sofa variants, 25 television variants, 8 armchair variants, and 8 chair variants, among others (see Appendix~\ref{app:supertemplates} for the full inventory) --- providing substantial visual variety for the experiments reported in a companion evaluation. The World Editor can be used at any time to expand the available environments to the rest of the game world.

\begin{figure}[H]
\centering
\includegraphics[width=\textwidth]{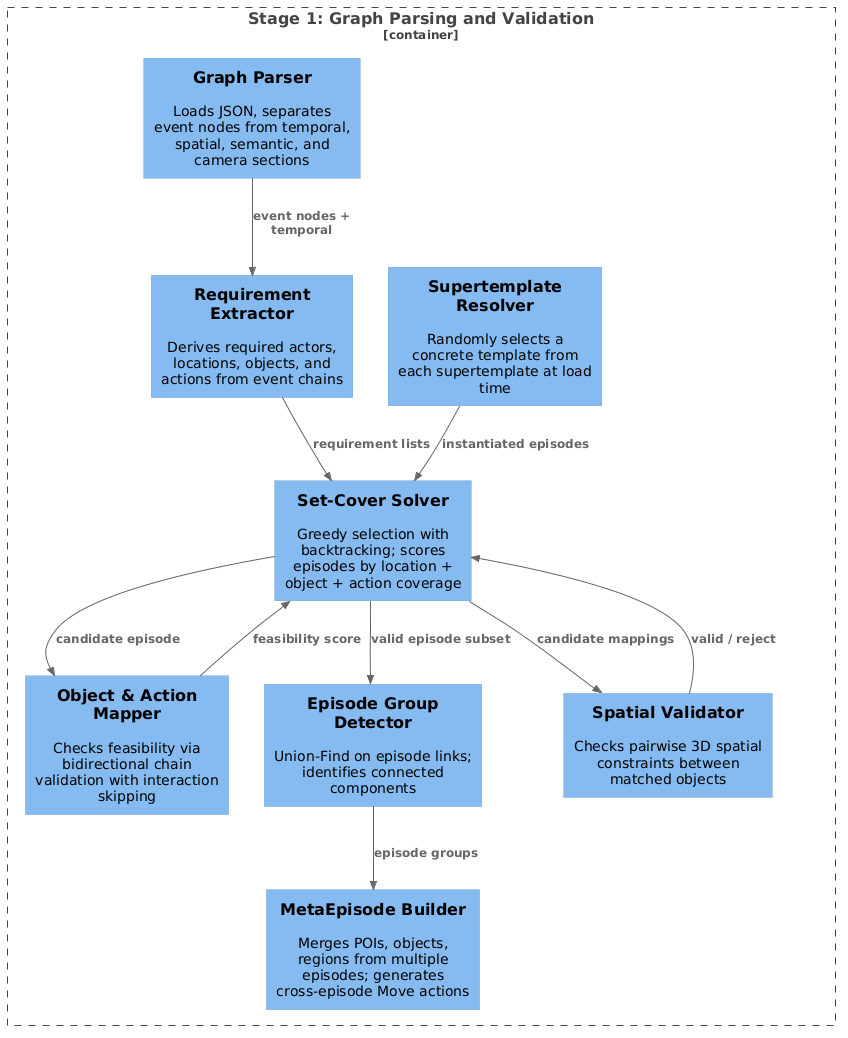}
\caption{\textbf{Stage~1 components.} The graph parser extracts requirements from the GEST specification, selects valid episodes via a greedy set-cover algorithm with backtracking, detects episode groups, and constructs a unified MetaEpisode. The Object \& Action Mapper validates feasibility by checking that at least one valid mapping exists for each required entity; in Stage~2 (Section~\ref{sec:stage_grounding}), the same component runs again in commit mode to produce the definitive bidirectional maps.}
\label{fig:stage1_components}
\end{figure}

\section{Stage 1: Graph Parsing and Validation}
\label{sec:stage_parsing}

The first stage of the pipeline determines whether a given GEST specification can be realized in the available simulation environments, and if so, prepares a unified execution environment for the subsequent stages. Given a GEST in the format described in Appendix~\ref{app:gest_json}, the graph parser must accomplish several tasks: parse the input JSON and separate its structural sections, extract all requirements --- which actors, locations, objects, and actions the story needs --- find a valid combination of episodes that can satisfy those requirements, detect which episodes are physically connected and which require context switches, and aggregate the selected episodes into a unified execution environment. Validation includes checking that at least one feasible object and action mapping exists for each required entity (Section~\ref{sec:set_cover}); the definitive mapping that produces the persistent bidirectional maps is performed in Stage~2 (Section~\ref{sec:stage_grounding}). Figure~\ref{fig:stage1_components} provides an overview of the components involved.

\subsection{The Input Format}
\label{sec:engine_input_format}

The input to the pipeline is a GEST serialized as a single self-contained JSON document in the current format described in Appendix~\ref{app:gest_json}, which also provides a complete example. While the GEST model has been introduced at the conceptual level --- events, edges, hierarchical structure --- here we describe the concrete fields that the graph parser operates on, as they are essential to understanding the algorithms that follow.

Although both the engine and the GEST format evolved over time (Appendix~\ref{app:gest_json}, Appendix~\ref{app:gest_json}), this report describes the current state of the system and refers to the latest version of the format simply as ``the GEST.'' A GEST file is a flat JSON object whose keys are event identifiers. Each event is a dictionary with the following fields:

\begin{itemize}
    \item \textbf{Action}: a string naming the action type (e.g., \texttt{"SitDown"}, \texttt{"Eat"}, \texttt{"Move"}, \texttt{"Talk"}). The special action \texttt{"Exists"} declares an entity without performing any physical action --- it serves as a declaration node for actors and objects.
    \item \textbf{Entities}: an ordered array of entity identifiers. For unary actions (e.g., \texttt{Move}), this contains only the actor. For binary actions (e.g., \texttt{Eat}), it contains the actor and the target object. For ternary actions (e.g., \texttt{Give}), it contains the giver, the receiver, and the object being transferred.
    \item \textbf{Location}: an array of location names. For most actions, this is a single-element array (e.g., \texttt{["bedroom"]}). For \texttt{Move} actions, it contains two elements: the source and destination locations (e.g., \texttt{["bedroom", "kitchen"]}).
    \item \textbf{Timeframe}: stores the start and end frame numbers for the event. This field is populated when the engine produces a proto-graph during artifact collection (Section~\ref{sec:temporal_alignment}); in input specifications, it is \texttt{null}.
    \item \textbf{Properties}: a dictionary of metadata. For actor \texttt{Exists} nodes, this includes \texttt{Gender} (1 for male, 2 for female) and \texttt{Name}. For object \texttt{Exists} nodes, this includes \texttt{Type} (the semantic category, e.g., \texttt{"Bed"}, \texttt{"Chair"}, \texttt{"Laptop"}).
\end{itemize}

A special top-level key, \texttt{"temporal"}, encodes the temporal structure of the graph. It contains:

\begin{itemize}
    \item \textbf{starting\_actions}: a dictionary mapping each actor identifier to the identifier of their first event (e.g., \texttt{\{"alice": "a1", "bob": "b1"\}}).
    \item \textbf{Per-event entries}: for each action event, a dictionary with a \texttt{next} field pointing to the subsequent event in that actor's chain (or \texttt{null} for the final event), and a \texttt{relations} field listing any temporal constraints attached to that event.
    \item \textbf{Temporal constraint definitions}: named entries such as \texttt{"tm0": \{"type": "starts\_with"\}} or \texttt{"m8\_before\_a1": \{"source": "m8", "type": "before", "target": "a1"\}}. Four constraint types, drawing on Allen's interval algebra \cite{allen1983maintaining}, are supported: \texttt{before}, \texttt{after}, \texttt{same\_time}, and \texttt{concurrent}. The \texttt{before} and \texttt{after} constraints are directional: each carries explicit \texttt{source} and \texttt{target} fields identifying the two events --- for instance, \texttt{\{"source": "a3", "type": "before", "target": "b1"\}} means that event \texttt{a3} must complete before event \texttt{b1} can begin. The \texttt{same\_time} constraint, by contrast, is symmetric: it is defined as a named entry (e.g., \texttt{"tm0": \{"type": "same\_time"\}}), and any two events that list this identifier in their \texttt{relations} arrays will begin simultaneously. \texttt{Concurrent} events execute in overlapping time windows without requiring synchronized start times.
\end{itemize}

Optional top-level keys include \texttt{"spatial"} (pairwise spatial constraints between objects, such as \texttt{left\_of} or \texttt{on\_top\_of}), \texttt{"semantic"} (semantic relationships between events), \texttt{"logical"} (causal or dependency constraints), and \texttt{"camera"} (spectator camera configuration with mode and specific commands).

Figure~\ref{fig:gest_json_example} shows a sample GEST specification with two actors and one shared object, simplified for illustrative purposes. Real graphs are substantially more complex: the most complex graphs in our corpus contain six actors across four locations with over 50 events and 30 temporal constraints (Appendix~\ref{app:gest_json} provides a full-scale example).

\begin{figure}[htbp]
\centering
\begin{minipage}[t]{0.48\textwidth}
\begin{verbatim}
{
 "Bed": {
   "Action": "Exists",
   "Entities": ["Bed"],
   "Location": ["bedroom"],
   "Properties": {"Type": "Bed"}
 },
 "alice": {
   "Action": "Exists",
   "Entities": ["alice"],
   "Location": ["bedroom"],
   "Properties": {
     "Gender": 2, "Name": "Alice"}
 },
 "bob": {..."Action": "Exists"...},
 "a1": {
   "Action": "GetOn",
   "Entities": ["alice", "Bed"],
   "Location": ["bedroom"],
   "Properties": {}
 },
 "a2": {
   "Action": "Sleep",
   "Entities": ["alice", "Bed"],
   "Location": ["bedroom"],
   "Properties": {}
 },
 "a3": {
   "Action": "GetOff",
   "Entities": ["alice", "Bed"],
   "Location": ["bedroom"],
   "Properties": {}
 },
 "b1": {
   "Action": "GetOn",
   "Entities": ["bob", "Bed"],
   "Location": ["bedroom"],
   "Properties": {}
 },
\end{verbatim}
\end{minipage}
\hfill
\begin{minipage}[t]{0.48\textwidth}
\begin{verbatim}
 "b2": {
   "Action": "Sleep",
   "Entities": ["bob", "Bed"],
   "Location": ["bedroom"],
   "Properties": {}
 },
 "b3": {
   "Action": "GetOff",
   "Entities": ["bob", "Bed"],
   "Location": ["bedroom"],
   "Properties": {}
 },
 "temporal": {
   "starting_actions": {
     "alice": "a1", "bob": "b1"},
   "a1": {"next": "a2",
     "relations": ["tm0"]},
   "tm0": {"type": "same_time"},
   "a2": {"next": "a3",
     "relations": null},
   "a3": {"next": null,
     "relations": ["a3_before_b3"]},
   "a3_before_b3": {"source": "a3",
     "type": "before", "target": "b3"},
   "b1": {"next": "b2",
     "relations": ["tm0"]},
   "b2": {"next": "b3",
     "relations": null},
   "b3": {"next": null,
     "relations": ["b3_after_a3"]},
   "b3_after_a3": {"source": "b3",
     "type": "after", "target": "a3"}
 }
}
\end{verbatim}
\end{minipage}
\caption{\textbf{Sample GEST specification.} Two actors share a Bed in a bedroom. Both chains follow \texttt{GetOn}$\to$\texttt{Sleep}$\to$\texttt{GetOff}. Three temporal constraint types are illustrated: \texttt{same\_time} (\texttt{tm0}) synchronizes both \texttt{GetOn} events; \texttt{before} (\texttt{a3\_before\_b3}) ensures Alice gets off before Bob; \texttt{after} (\texttt{b3\_after\_a3}) expresses the same ordering from Bob's perspective. \texttt{Timeframe} fields omitted for brevity; Bob's \texttt{Exists} node abbreviated.}
\label{fig:gest_json_example}
\end{figure}

\subsection{Parsing and Requirement Extraction}
\label{sec:requirement_extraction}

When the pipeline starts, the graph parser loads the specified JSON file, deserializes it, and separates the graph into its constituent sections. The temporal, spatial, semantic, logical, and camera sections are extracted and stored independently, leaving the main structure containing only event definitions. Each event is annotated with its own identifier, so that subsequent processing can reference events directly without maintaining separate key lookups.

The system then identifies which resources the graph requires. This extraction proceeds in four phases:

\textbf{Phase 1: Actor extraction.} The system scans all events for those with \texttt{Action == "Exists"} and a \texttt{Gender} property (or \texttt{Type == "Actor"}). Each such event declares an actor who must be present in the simulation. The actor's gender determines the pool of available skins from which a visual appearance will be randomly selected at runtime.

\textbf{Phase 2: Location extraction.} All non-empty \texttt{Location} fields across all events are collected and deduplicated. For \texttt{Move} events, both the source and destination locations are included. The resulting set represents all locations that must be available in the selected episodes --- for instance, \texttt{\{bedroom, bathroom, kitchen, livingroom, hallway\}} for a house-based story.

\textbf{Phase 3: Object extraction.} For each actor, the system traverses their temporal chain from the \texttt{starting\_actions} entry, following the \texttt{next} pointers through each subsequent event. At each event, if the \texttt{Entities} array contains a second element (or third, for ternary actions), and that element is not itself an actor (determined by the absence of a \texttt{Gender} property on its \texttt{Exists} node), it is recorded as a required object. Each object carries its type (from the \texttt{Properties.Type} field of its \texttt{Exists} node) and its declared location.

\textbf{Phase 4: Action extraction.} A second traversal of each actor's temporal chain collects all non-\texttt{Exists} actions, recording the action name, the location in which it occurs, and the target object (if any). \texttt{Move} actions receive special treatment: if a \texttt{Move} event references another entity (rather than a location pair), it is interpreted as a ``move toward entity'' rather than a ``move between rooms,'' and the target entity is recorded accordingly.

The output of this phase is a set of four requirement lists --- \texttt{requiredActors}, \texttt{requiredLocations}, \texttt{requiredObjects}, and \texttt{requiredActions} --- that together specify everything the simulation environment must provide.

\subsection{Episode Selection via Set-Cover with Backtracking}
\label{sec:set_cover}

Given the requirement lists, the system must find a set of episodes (simulation environments) that collectively satisfy all requirements. This is a variant of the weighted set-cover problem: each episode ``covers'' a subset of the required locations, objects, and actions, and we seek a minimal set of episodes whose union covers everything.

The algorithm proceeds as a greedy set-cover with hierarchical backtracking. At each step, the system evaluates all candidate episodes and selects the one with the highest coverage score. The coverage score for an episode is the sum of the number of required locations, objects, and actions that the episode can satisfy. Episode validation involves three checks:

\begin{enumerate}
    \item \textbf{Location matching}: each required location is matched against the episode's regions using case-insensitive substring matching (e.g., the required location \texttt{"bedroom"} matches a region named \texttt{"bedroom\_house1"}).
    \item \textbf{Object mapping}: each required object is matched against the episode's objects by type. Objects that belong to a predefined set of spawnable types (such as \texttt{Cigarette}, and \texttt{MobilePhone}) are automatically satisfied --- they do not need to exist as fixed objects in the episode because the engine will create them dynamically at runtime. For fixed objects, the system searches the episode's Points of Interest for actions that target objects of the matching type. For objects that can be carried and interacted with in multiple locations (e.g., \texttt{Food} or \texttt{Drinks}), the system is permissive, and a match in any location counts toward satisfying the requirement.
    \item \textbf{Action matching}: each required action is validated against the actions available at the episode's POIs, checking that the action name, location, and target object type all match.
\end{enumerate}

If the graph specifies spatial constraints between objects (e.g., ``the chair must be to the left of the desk''), these are validated against the actual 3D positions of the matched objects in the candidate episode. The spatial validator checks each pairwise relation using the objects' world coordinates and rotations. Spawnable objects pass spatial checks automatically, since they have no fixed position and can be placed anywhere. This is work in progress and still needs to be fully tested and implemented.

After selecting the highest-scoring episode, the algorithm explores two strategies for covering any remaining unsatisfied requirements:

\begin{enumerate}
    \item \textbf{Linked episodes}: the system follows the episode links defined in the selected episode's POIs to find connected episodes (e.g., a house interior linked to a garden through a door POI). This produces groups of episodes that share physical connection points, enabling actors to walk between them.
    \item \textbf{All remaining episodes}: if the linked-episode strategy fails (or no links exist), the system recursively explores all unselected episodes.
\end{enumerate}

The algorithm prefers linked episodes when available, as they produce spatially coherent stories where actors can physically move between environments through connection points. When both strategies fail for the current selection, the algorithm backtracks: it removes the current best episode from the candidate pool and tries the next-best episode. This backtracking continues until either a valid covering set is found or all candidates are exhausted.

Algorithm~\ref{alg:set_cover} formalizes this procedure.

\begin{algorithm}[tb]
\caption{Episode selection via greedy set-cover with backtracking.}
\label{alg:set_cover}
\KwInput{Candidate episodes $\mathcal{E}$, requirement sets $R_{\text{loc}}, R_{\text{obj}}, R_{\text{act}}$, current selection $S$}
\KwOutput{A valid episode subset $S^* \supseteq S$ covering all requirements, or $\emptyset$ if none exists}
\If{$R_{\text{loc}} = \emptyset$ \textbf{and} $R_{\text{obj}} = \emptyset$ \textbf{and} $R_{\text{act}} = \emptyset$}{
    \Return $S$ \tcp*{All requirements satisfied}
}
\If{$\mathcal{E} = \emptyset$}{
    \Return $\emptyset$ \tcp*{No candidates remain}
}
$U \leftarrow \mathcal{E}$ \tcp*{Unexplored candidates}
\While{$U \neq \emptyset$}{
    $e^* \leftarrow \arg\max_{e \in U} \textsc{Score}(e, R_{\text{loc}}, R_{\text{obj}}, R_{\text{act}})$\;
    \If{$\textsc{Score}(e^*) = 0$}{
        \Return $\emptyset$ \tcp*{No episode covers any requirement}
    }
    $R'_{\text{loc}}, R'_{\text{obj}}, R'_{\text{act}} \leftarrow$ remaining requirements after $e^*$\;
    \tcp{Strategy 1: prefer linked episodes}
    $S_1 \leftarrow \textsc{ExploreLinks}(e^*, \mathcal{E}, R'_{\text{loc}}, R'_{\text{obj}}, R'_{\text{act}}, S \cup \{e^*\})$\;
    \If{$S_1 \neq \emptyset$}{
        \Return $S_1$\;
    }
    \tcp{Strategy 2: try all remaining episodes}
    $S_2 \leftarrow \textsc{SetCover}(\mathcal{E} \setminus \{e^*\}, R'_{\text{loc}}, R'_{\text{obj}}, R'_{\text{act}}, S \cup \{e^*\})$\;
    \If{$S_2 \neq \emptyset$}{
        \Return $S_2$\;
    }
    $U \leftarrow U \setminus \{e^*\}$ \tcp*{Backtrack: try next-best}
}
\Return $\emptyset$\;
\end{algorithm}

The algorithm is sound but not necessarily optimal: it will find a valid set of episodes if one exists, but may not find the smallest such set. Given the small number of candidate episodes (11 in the current system), the greedy heuristic with the preference for linked episodes produces near-minimal coverings in practice.

Before an episode can be scored, it must be fully instantiated: each supertemplate referenced in the episode definition is resolved by randomly selecting one of its constituent templates (as described in Section~\ref{sec:world_editor}). The selected template's POI definition --- including its 3D position, rotation, associated actions, and target objects --- replaces the supertemplate reference. The episode is then partially initialized: world objects, regions, and collision geometry are created, but actors are not yet spawned. This partial initialization is memoized --- the resolved templates and world geometry are retained and reused when the episode is fully initialized for execution, avoiding redundant work. The combination of supertemplate resolution and the deferred object binding described in Stage~2 (Section~\ref{sec:stage_grounding}) creates a two-level randomization: the visual appearance of objects varies between simulation runs (different bed models, different chair styles, different sofa designs), and the spatial placement and POI assignment of actors varies within a fixed set of resolved templates. Together, these ensure that repeated executions of the same GEST produce visually distinct videos while preserving the semantic content specified by the graph --- a property that is essential for generating the large-scale, varied training data used in a companion evaluation.

The scoring of each candidate episode goes beyond simple location and object type matching. For each required object, the system checks whether there exists at least one POI whose action chain can support the complete sequence of actions the graph specifies for that object. This is done through bidirectional action chain validation (Algorithm~\ref{alg:chain_validation}), which traverses the POI's action chain both backward and forward from a candidate match, simultaneously following the graph's temporal chain. At each step, the \textsc{Match} predicate checks three conditions: the action names match (case-insensitive), the locations match (the POI's region contains the event's location as a substring), and the target object types match (both nil, or both present with the same semantic type). When a match succeeds, the corresponding bindings (event-to-action, object-to-simulator-object, event-to-POI) are optionally recorded into shared maps passed by reference.

The forward traversal implements \emph{interaction skipping}: when the traversal encounters an interaction action (\texttt{Talk}, \texttt{Hug}, \texttt{Kiss}, \texttt{Handshake}, \texttt{Give}) or a passive observation action (\texttt{LookAtObject}, \texttt{Wave}, \texttt{Stash}, \texttt{TakeOut}), it advances the graph event pointer without advancing the POI action pointer, since interaction actions cause actors to move to special interaction POIs and do not need to be part of the object manipulation chain.

\begin{algorithm}[tb]
\caption{Bidirectional action chain validation for a candidate POI.}
\label{alg:chain_validation}
\KwInput{Graph event $e_0$ matching POI action $a_0$; POI action chain; graph temporal chain; shared maps $M$ (optional)}
\KwOutput{\textbf{true} if the entire chain matches, \textbf{false} otherwise. On success, bindings recorded in $M$.}
\tcp{Backward matching}
$e \leftarrow e_0$; $a \leftarrow a_0$\;
\While{$\textsc{PreviousAction}(a) \neq \texttt{nil}$}{
    $e_{\text{prev}} \leftarrow \textsc{PreviousGraphEvent}(e)$\;
    \If{$e_{\text{prev}} = \texttt{nil}$ \textbf{or} $\textsc{IsLocationChanging}(e_{\text{prev}})$}{
        \textbf{break}\;
    }
    $a_{\text{prev}} \leftarrow \textsc{PreviousAction}(a)$\;
    \If{$\neg\textsc{Match}(a_{\text{prev}}, e_{\text{prev}}, M)$}{
        \Return \textbf{false}\;
    }
    $e \leftarrow e_{\text{prev}}$; $a \leftarrow a_{\text{prev}}$\;
}
\tcp{Forward matching}
$e \leftarrow e_0$; $a \leftarrow a_0$\;
\While{$\textsc{NextAction}(a) \neq \texttt{nil}$ \textbf{and} $\textsc{NextGraphEvent}(e) \neq \texttt{nil}$}{
    $e_{\text{next}} \leftarrow \textsc{NextGraphEvent}(e)$\;
    \If{$\textsc{IsInteractionOrSkippable}(e_{\text{next}})$}{
        $e \leftarrow e_{\text{next}}$ \tcp*{Skip without advancing POI chain}
        \textbf{continue}\;
    }
    \If{$\textsc{IsLocationChanging}(e_{\text{next}})$}{
        \textbf{break}\;
    }
    $a_{\text{next}} \leftarrow \textsc{NextAction}(a)$\;
    \If{$\neg\textsc{Match}(a_{\text{next}}, e_{\text{next}}, M)$}{
        \Return \textbf{false}\;
    }
    $e \leftarrow e_{\text{next}}$; $a \leftarrow a_{\text{next}}$\;
}
\Return \textbf{true}\;
\end{algorithm}

The validation processes each object independently, continuing regardless of whether individual objects fail to map; the total number of successfully mapped locations, objects, and actions determines the coverage score. The mappings produced during validation are transient --- they are used only for scoring and then discarded. In Stage~2 (Section~\ref{sec:stage_grounding}), the full mapping algorithm (Algorithm~\ref{alg:mapping}) calls Algorithm~\ref{alg:chain_validation} for every candidate POI of every required object, aggregates all valid chains with unique chain-IDs, and persists the resulting maps for the remainder of the pipeline.

\subsection{Episode Group Detection}
\label{sec:episode_groups}

Once a valid set of episodes has been selected, the system determines which episodes are physically connected to each other. This is important because actors can walk between linked episodes (through connection POIs such as doors), but movement between unlinked episodes requires a context switch --- a teleportation with screen fading, similar to a scene cut in a movie.

The system uses a Union-Find (disjoint-set) data structure with path compression and union-by-rank to identify connected components among the selected episodes. The algorithm iterates through all POIs in all episodes; whenever a POI has a link pointing to another selected episode, the two episodes are unioned. After processing all links, the connected components represent episode groups --- sets of episodes that actors can navigate between continuously.

The resulting data structure provides two mappings: from each group identifier to the set of episodes it contains, and from each episode name to its group identifier. These mappings are used by the MetaEpisode wrapper and by the cross-episode movement validation, which ensures that explicit Move actions in the graph do not cross group boundaries --- only the system's auto-generated cross-group movements are allowed.

The episode group structure also has implications for the text-to-GEST generation approaches described below --- procedural generation (Section~\ref{sec:procedural_gen}) and agentic generation (Section~\ref{sec:agentic}). When the procedural generator or the agentic system creates GESTs, it can use the episode group information to understand which locations are physically adjacent and which would require a context switch, enabling it to generate stories with realistic spatial progressions.

\subsection{The MetaEpisode Wrapper}
\label{sec:meta_episode}

When the selected set contains multiple episodes, they are aggregated into a MetaEpisode --- a unified execution environment that presents all constituent episodes as a single coherent world to the rest of the pipeline. The MetaEpisode merges the POIs, objects, regions, actors, and supertemplates from all episodes into unified lists, while maintaining back-references from each POI to its source episode.

The MetaEpisode construction proceeds as follows:

\begin{enumerate}
    \item \textbf{Episode initialization}: each constituent episode is initialized in sequence. The first episode receives the actor assignments; subsequent episodes are initialized without actors, since all actors are managed at the MetaEpisode level.
    \item \textbf{POI aggregation}: all POIs from all episodes are collected into a single list. Each POI is annotated with a back-reference to its source episode, which is used later by the location system to determine which 3D context to activate when an actor moves to that POI.
    \item \textbf{Actor distribution}: the unified list of actors is propagated to all constituent episodes, ensuring that any actor can be dispatched to any episode during execution.
    \item \textbf{Link resolution}: explicit episode links defined in POIs are resolved to concrete source-POI/target-episode pairs, establishing the physical connection points between episodes.
    \item \textbf{Cross-episode movement generation}: for each pair of distinct episodes within the same group, the system generates bidirectional Move actions between all pairs of POIs. These auto-generated movements are added to the POIs' action lists, enabling the orchestrator to route actors between episodes as needed. For episodes in different groups, cross-episode movement is handled through context switching (Section~\ref{sec:cross_episode}) rather than physical navigation.
\end{enumerate}

The MetaEpisode thus presents a complete, unified view of all available locations, objects, and actions to the temporal orchestrator (Stage~3), which can plan actor movements across the entire multi-episode world without needing to know which specific episode each POI belongs to. The episode-level context switching is handled transparently by the execution layer (Stage~4).

\section{Stage 2: Entity \& Action Grounding}
\label{sec:stage_grounding}

\begin{figure}[htbp]
\centering
\includegraphics[width=0.5\textwidth]{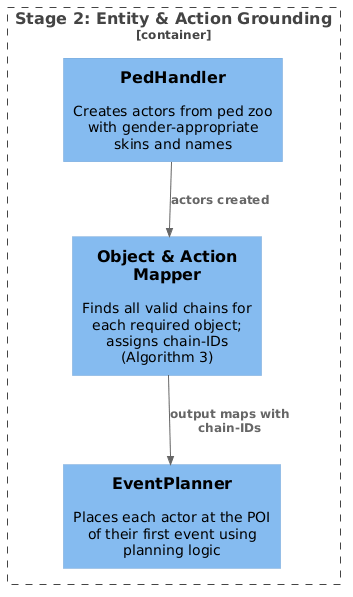}
\caption{\textbf{Stage~2 components.} The PedHandler creates actors with skins and names from the ped zoo. The Object \& Action Mapper finds all valid chains and assigns chain-IDs (Algorithm~\ref{alg:mapping}). The EventPlanner places each actor at the POI corresponding to their first event.}
\label{fig:stage2_components}
\end{figure}

Once validation confirms that the selected episodes can realize the story (Section~\ref{sec:set_cover}), Stage~2 grounds every abstract entity in the GEST to a concrete counterpart in the 3D simulation. This involves four kinds of grounding. \emph{Actor grounding} creates the actors (peds) with gender-appropriate visual appearances (skins) and names --- either specified in the GEST or randomly assigned from pools of approximately 1,000 male first names, 1,000 female first names, and 1,350 family names drawn from US census data. \emph{Action chain grounding} runs the mapping algorithm in commit mode --- unlike the validation phase, which only checks feasibility, this phase requires every object to map successfully and enumerates \emph{all} valid chains, producing one-to-many bidirectional maps with chain-ID assignments. \emph{Object grounding} binds each abstract graph object to one or more concrete 3D objects (or marks it as spawnable for on-demand creation). \emph{Location grounding} maps each event to specific POIs within specific regions, and places each actor at the POI corresponding to their first event. Figure~\ref{fig:stage2_components} shows the three components involved.

\subsection{Actor Grounding}
\label{sec:actor_grounding}

Each actor declared via an \texttt{Exists} node in the GEST must be realized as a ped (pedestrian character) in the 3D world. The GEST specifies each actor's gender and, optionally, a name and a skin identifier. If a skin identifier is provided, the system selects exactly that visual appearance; otherwise, it randomly selects from the 312 available skins, filtered by gender and excluding skins already assigned to other actors in the same story. The skin system provides a diverse range of visual appearances spanning different ages, races, and clothing styles, ensuring that actors in the same story are visually distinguishable.

Each actor is also assigned a name. If the GEST provides a name in the actor's properties, it is used directly. Otherwise, the system randomly selects a first name from a gender-appropriate pool (approximately 1,000 male and 1,000 female first names) and a family name from a shared pool of approximately 1,350 surnames, both derived from US census data. The assigned name is stored as actor metadata and used by the Logger (Section~\ref{sec:logger}) when generating natural language descriptions of the simulation.

Actor creation is managed through the \emph{ped zoo} --- a registry that tracks all created actors across the simulation lifetime. The ped zoo supports actor reuse across simulation runs: when a new actor is needed, the system first checks the zoo for an unassigned ped and recycles it, creating a new one only if none is available. Every actor is initialized with a default inventory containing a mobile phone and a cigarette, enabling spawnable actions (Section~\ref{sec:spawnable_objects}) without additional setup.

At this point, actors are created at temporary positions. Their definitive placement at the POI corresponding to their first event occurs after the mapping algorithm has determined where each actor's action chain begins (Section~\ref{sec:initial_placement}).

\subsection{Chains and Chain-IDs}
\label{sec:chains}

A \emph{chain} is the central concept of the grounding phase: it represents one complete, physically plausible action sequence at one specific POI, bound to one specific set of objects. The term comes from the fact that POI actions are linked through next-action pointers, forming a chain that the actor must follow in order.

Consider a concrete example. A classroom episode contains twelve desks, each defined through the same template: a desk object, a chair object, and a POI with the action chain \texttt{SitDown} $\to$ \texttt{OpenLaptop} $\to$ \texttt{TypeOnKeyboard} $\to$ \texttt{LookAtTheWatch} $\to$ \texttt{CloseLaptop} $\to$ \texttt{StandUp}. If a GEST specifies that Alice should sit down, open a laptop, type, close the laptop, and stand up, the mapping algorithm finds that all twelve desks can satisfy this sequence. Each desk produces a valid chain --- the same sequence of actions, but bound to a different physical chair, a different physical desk, and a different physical laptop at a different location in the classroom.

Each of these twelve chains receives a unique chain-ID during the mapping phase. The chain-ID is constructed from the POI description, the region name, the location identifier, and a global counter, producing identifiers such as \texttt{desk\_classroom\_loc5\_7}. The chain-ID serves two purposes. First, it ensures \emph{sequence consistency}: once the runtime selects a particular chain for an actor, all subsequent actions in that actor's sequence use the same physical objects and the same POI. Alice's \texttt{SitDown}, \texttt{OpenLaptop}, \texttt{TypeOnKeyboard}, \texttt{CloseLaptop}, and \texttt{StandUp} all execute at the same desk. Second, it enables \emph{deferred selection}: because all twelve chains are stored in the output maps, the runtime can choose among them based on availability and proximity, producing different videos from the same GEST.

A second example illustrates branching. A living room POI for a turntable has the action chain \texttt{TurnOn} $\to$ \texttt{Dance} $\to$ \texttt{TurnOff}. The \texttt{Dance} action executes at a nearby POI (not at the turntable itself), and the actor must return to the turntable POI for \texttt{TurnOff}. The chain captures this spatial progression: the chain-ID binds not just the object (turntable) but also the sequence of locations the actor visits during the chain.

\subsection{The Mapping Algorithm}
\label{sec:mapping_algorithm}

The mapping algorithm (Algorithm~\ref{alg:mapping}) iterates over all required objects and, for each one, finds \emph{every} valid chain in the selected episodes. Unlike the validation phase in Stage~1, which processes objects independently and tolerates failures, the mapping algorithm requires all objects to map successfully --- if any object fails, the entire grounding phase fails. For each required object, the algorithm proceeds through five stages:

\textbf{Stage A: Spawnable detection.} If the object's type belongs to the set of spawnable objects (\texttt{Cigarette}, \texttt{MobilePhone}, \texttt{Food}, \texttt{Drinks}, and a few others), it is immediately marked as spawnable and no further matching is needed. These objects will be created dynamically by the actor at the appropriate moment during execution.

\textbf{Stage B: Event collection.} The system identifies all events in the graph that use this object as a target --- i.e., events where the object appears as the second or third element in the \texttt{Entities} array. Events that are pure observation actions (\texttt{LookAt}, \texttt{LookAtObject}, \texttt{Wave}) or interaction actions (\texttt{Talk}, \texttt{Hug}, \texttt{Kiss}, \texttt{Handshake}) are excluded, since these do not require a physical object binding.

\textbf{Stage C: Chain matching.} For each POI in the episode that contains an action targeting an object of the matching type, the system calls Algorithm~\ref{alg:chain_validation} to attempt bidirectional chain matching. Each POI that produces a successful match yields a valid chain, with the corresponding event-to-action, object-to-simulator-object, and event-to-POI bindings recorded in the shared maps.

\textbf{Stage D: Chain-ID assignment.} Each valid chain receives a unique chain identifier (chain-ID), constructed from the POI description, region name, location identifier, and a global counter. The chain-ID groups all actions in a sequence so that they are bound to the same physical object and the same POI at runtime. All valid chains for each object are aggregated into the shared output maps, producing the one-to-many structure described in Section~\ref{sec:output_maps}.

\textbf{Stage E: PickUp/PutDown consistency.} When the graph contains \texttt{PickUp} and \texttt{PutDown} events for the same object, the algorithm enforces that \texttt{PutDown} maps to the same location where \texttt{PickUp} occurred. This prevents objects from being placed at arbitrary locations and ensures spatial consistency in the generated video.

\begin{algorithm}[tb]
\caption{Object and action mapping.}
\label{alg:mapping}
\KwInput{Required objects $\mathcal{O}$, episode $\epsilon$ with POIs, shared maps $M$}
\KwOutput{\textbf{true} if all objects mapped, \textbf{false} otherwise. Maps $M$ populated with all valid chains.}
\ForEach{required object $o \in \mathcal{O}$}{
    \If{$o.\text{type} \in \textsc{SpawnableTypes}$}{
        $M.\text{objects}[o] \leftarrow \{\text{spawnable}\}$\;
        \textbf{continue}\;
    }
    $E_o \leftarrow \textsc{CollectEvents}(o)$ \tcp*{Graph events using $o$ as target}
    $\text{matched} \leftarrow \textbf{false}$\;
    \ForEach{POI $p \in \epsilon$ with action targeting type $o.\text{type}$}{
        \If{$\textsc{ValidateChain}(p, E_o, M)$  \tcp*{Algorithm~\ref{alg:chain_validation}}}{
            $\text{chainId} \leftarrow \textsc{GenerateChainId}(p)$\;
            $\textsc{AggregateChain}(M, p, \text{chainId})$\;
            $\text{matched} \leftarrow \textbf{true}$\;
        }
    }
    \If{$\neg\text{matched}$}{
        \Return \textbf{false} \tcp*{Fail: no valid chain for $o$}
    }
}
\Return \textbf{true}\;
\end{algorithm}

\subsection{Output Maps}
\label{sec:output_maps}

The result of the grounding phase is a set of bidirectional maps:
\begin{itemize}
    \item \textbf{Object map}: maps each graph object identifier to a list of valid simulator object identifiers (with chain-IDs).
    \item \textbf{POI map}: maps each graph event identifier to a list of valid POI location identifiers (with chain-IDs).
    \item \textbf{Event-to-action map}: maps each graph event identifier to a list of valid simulator action identifiers (with chain-IDs).
    \item \textbf{Action-to-event map}: the reverse --- maps simulator action identifiers back to graph event identifiers.
\end{itemize}

These maps are one-to-many: a single graph object may map to multiple valid simulator objects, and a single graph event may be executable at multiple POIs. The final selection among valid alternatives is deferred to runtime (Stage~3 and Stage~4), where the location system and POI coordinator choose among candidates based on availability, spatial proximity, and conflict avoidance. This deferred binding is what enables the engine to produce visually different videos from the same GEST specification --- the same semantic content can be realized with different objects, at different POIs, with different actor skins, each time the graph is executed.

\subsection{Spawnable vs.\ Fixed Objects}
\label{sec:spawnable_objects}

The mapping algorithm distinguishes between two categories of objects based on how they are realized in the 3D world. \emph{Fixed objects} --- furniture, appliances, gym equipment, and other items that are part of the episode's environment --- exist as permanent entities with fixed positions, loaded from the episode's JSON definition during initialization. Each fixed object is bound to one or more POIs through the action definitions in its template: a bed is bound to the POI where the actor can sleep, a chair to the POI where the actor can sit. When the mapping algorithm matches a graph object to a fixed object, the match is anchored to a specific physical location in the episode.

\emph{Spawnable objects}, by contrast, do not exist in the episode until an actor needs them. The current system defines two spawnable types: \texttt{Cigarette} and \texttt{MobilePhone}. When the mapping algorithm encounters a spawnable object type, it marks the mapping as spawnable and skips further POI matching --- no physical object needs to exist in the episode at validation time. The object is created on demand during execution, when the actor performs a \texttt{TakeOut} action: the system instantiates the appropriate 3D model, assigns it an actor-specific identifier, and places it in the actor's inventory. Because each actor receives their own instance, spawnable objects never cause multi-actor conflicts. Additional object types (\texttt{Food}, \texttt{Drinks}, \texttt{Remote}) are defined as pick-up-able --- they exist as fixed objects in the episode but can be picked up, carried, and put down by actors, requiring the PickUp/PutDown constraint described in Stage~E of the mapping algorithm. A current limitation is that \texttt{PutDown} must return the object to the same location where it was picked up --- the system does not yet support placing objects at arbitrary locations or allowing a different actor to pick up an object that was put down elsewhere. Extending the engine to support arbitrary object placement and subsequent retrieval by any actor is left as future work (Section~\ref{sec:engineering}). Table~\ref{tab:object_categories} summarizes the three categories.

\begin{table}[htbp]
\centering
\caption{Object categories in the GEST-Engine and their grounding behavior.}
\label{tab:object_categories}
\small
\begin{tabular}{@{}llll@{}}
\toprule
\textbf{Category} & \textbf{Examples} & \textbf{Grounding} & \textbf{Conflict handling} \\
\midrule
Fixed & Bed, Chair, Desk, Sofa, & Bound to POI at & Chain-ID ensures \\
      & Laptop, TV, Sink, Treadmill & episode load time & different actors use \\
      &  &  & different instances \\
\midrule
Spawnable & Cigarette, MobilePhone & Created on demand & Each actor gets \\
          &  & during \texttt{TakeOut} & own instance \\
\midrule
Pick-up-able & Food, Drinks, Remote & Fixed at load time; & PutDown must return \\
             &  & can be picked up & to PickUp location \\
             &  & and carried &  \\
\bottomrule
\end{tabular}
\end{table}

\subsection{Initial Actor Placement}
\label{sec:initial_placement}

After the mapping algorithm has produced the bidirectional maps, each actor must be placed at a physically plausible starting location. The system uses the same planning logic that will later drive runtime action dispatch (Section~\ref{sec:event_planner}): for each actor, it looks up the POI mapped to that actor's first event and teleports the actor to that POI's 3D position and rotation. If the target POI is already occupied by a previously placed actor, the system falls back to a random unoccupied POI in the same region --- ensuring that all actors begin the simulation at locations consistent with the start of their action chains, without spatial conflicts.

\section{Stage 3: Temporal Orchestration}
\label{sec:stage_orchestration}

\begin{figure}[htbp]
\centering
\includegraphics[width=0.5\textwidth]{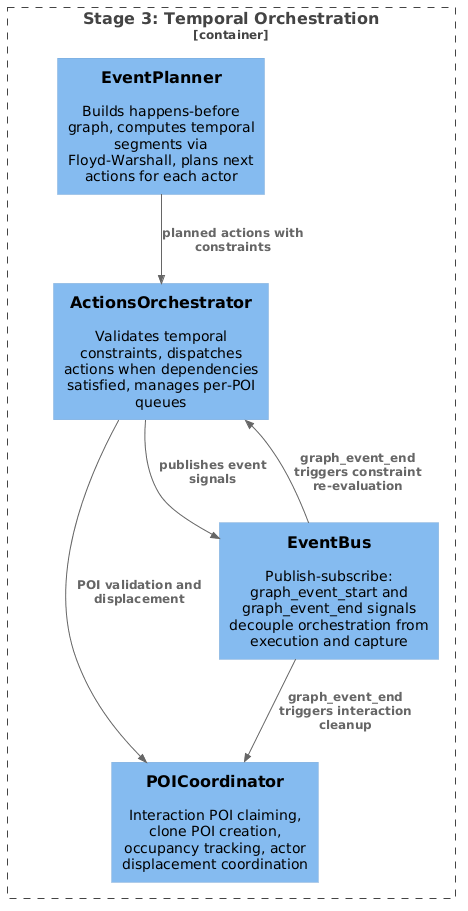}
\caption{\textbf{Stage~3 components.} The EventPlanner computes temporal segments and plans actions for each actor. The ActionsOrchestrator validates constraints and dispatches actions when dependencies are satisfied. The POICoordinator manages interaction claiming, clone POI creation, occupancy tracking, and actor displacement. The EventBus decouples orchestration from execution and artifact collection through publish-subscribe signals.}
\label{fig:stage3_components}
\end{figure}

With all entities grounded and actors placed at their starting positions, Stage~3 coordinates the temporal execution of the story. The challenge is to schedule events across multiple actors while respecting the temporal constraints specified in the GEST --- ensuring that actions happen in the correct order, that synchronized actions begin simultaneously, and that concurrent activities overlap as specified. Figure~\ref{fig:stage3_components} shows the three components involved.

\subsection{The EventPlanner}
\label{sec:event_planner}

The EventPlanner is responsible for determining \emph{what} each actor should do next and \emph{where} they should do it. Its central contribution is the computation of temporal segments --- groups of events that can execute concurrently within each segment, with segments themselves executing sequentially. This computation proceeds in three phases.

In the first phase, the EventPlanner constructs a \emph{happens-before graph} from the cross-actor temporal constraints in the GEST. For each \texttt{before} relation, an edge is added from the source event to the target event; for each \texttt{after} relation, the direction is inverted. For \texttt{same\_time} relations, the planner computes transitive edges: if event $A$ starts with event $B$, and $A$ chains to $A_2$ within the same actor, then $B$ must happen before $A_2$. The result is a directed graph where an edge from $e_i$ to $e_j$ means that $e_i$ must complete before $e_j$ can begin.

In the second phase, the happens-before graph is closed under transitivity using the Floyd-Warshall algorithm \cite{floyd1962algorithm,warshall1962theorem}. For every triple of events $(i, k, j)$, if $i$ happens-before $k$ and $k$ happens-before $j$, then $i$ happens-before $j$. This transitive closure reveals all indirect dependencies --- for instance, if Alice's eating must finish before Bob's phone call, and Bob's phone call must finish before Carol's exit, then Alice's eating implicitly constrains Carol's exit.

In the third phase, the planner partitions all events into temporal segments through a dependency-first iterative scheduling algorithm (Algorithm~\ref{alg:segments}). In each round, the planner identifies all events whose dependencies are satisfied: the event's same-actor predecessor must be complete, all cross-actor happens-before dependencies must be fulfilled, and any \texttt{same\_time} partner must also be ready. Ready events are grouped into concurrent segments: events that share a \texttt{same\_time} constraint are placed in the same segment, while events connected by happens-before edges are placed in different segments. The process repeats until all events are assigned to segments. If no events are ready in a round but unassigned events remain, the algorithm detects a deadlock --- a circular dependency in the temporal constraints --- and reports an error.

\begin{algorithm}[tb]
\caption{Temporal segment computation via dependency-first scheduling.}
\label{alg:segments}
\KwInput{GEST events $\mathcal{V}$, temporal constraints $\mathcal{C}$}
\KwOutput{Temporal segments $\{S_1, S_2, \ldots, S_k\}$ partitioning $\mathcal{V}$}
$H \leftarrow \textsc{BuildHappensBeforeGraph}(\mathcal{C})$\;
$H^* \leftarrow \textsc{FloydWarshallClosure}(H)$\;
$\text{satisfied} \leftarrow \emptyset$; $k \leftarrow 0$\;
\While{$|\text{satisfied}| < |\mathcal{V}|$}{
    $\text{ready} \leftarrow \{e \in \mathcal{V} \setminus \text{satisfied} \mid \textsc{IsReady}(e, H^*, \text{satisfied})\}$\;
    \If{$\text{ready} = \emptyset$}{
        \textbf{error} ``Deadlock: circular dependency detected''\;
    }
    $\text{groups} \leftarrow \textsc{GroupByConcurrency}(\text{ready}, H^*)$\;
    \ForEach{group $g \in \text{groups}$}{
        $k \leftarrow k + 1$; $S_k \leftarrow g$\;
        $\text{satisfied} \leftarrow \text{satisfied} \cup g$\;
    }
}
\Return $\{S_1, \ldots, S_k\}$\;
\end{algorithm}

The \textsc{IsReady} predicate in Algorithm~\ref{alg:segments} checks three conditions: (1) the event's same-actor predecessor is in the satisfied set, (2) all events that must happen before this event (per $H^*$) are in the satisfied set, and (3) if the event has a \texttt{same\_time} partner, that partner is also ready. The \textsc{GroupByConcurrency} procedure iteratively builds groups of compatible events: it starts with the first ready event, adds its \texttt{same\_time} partner if any, then adds further events that have no happens-before conflict with any event already in the group.

Once segments are computed, the EventPlanner plans actions for each actor on demand. When the ActionsOrchestrator requests the next action for an actor, the EventPlanner looks up the actor's next event, determines its temporal segment, and routes the event to the appropriate flow handler based on its type: fixed-chain actions are matched to POIs using the output maps from Stage~2, observation actions (such as \texttt{LookAt} or \texttt{Wave}) execute at the actor's current location, interaction actions are coordinated between two actors through a claiming mechanism, movement actions compute paths between regions, and spawnable actions are planned with a reference to the actor's inventory. For fixed-chain actions, the EventPlanner selects among valid POIs using a collision scoring function that penalizes POIs already assigned to other actors in the same temporal segment, minimizing spatial conflicts without requiring global coordination.

The EventPlanner operates entirely at the logical level --- it determines which actions to perform, at which locations, and with which objects, but does not interact with the game engine directly. Its output is a list of planned actions, each carrying a target POI, a target object descriptor, and the performing actor, annotated with the graph event identifier and temporal segment. This list is handed to the ActionsOrchestrator, which determines when to dispatch each action based on temporal constraints, and ultimately to Stage~4 (Section~\ref{sec:stage_execution}), where the planned actions are realized in the 3D world --- including the instantiation of spawnable objects when their corresponding actions are executed.

\subsection{The ActionsOrchestrator}
\label{sec:actions_orchestrator}

The ActionsOrchestrator is responsible for determining \emph{when} actions execute. It maintains a queue of event requests --- one per actor --- each annotated with the temporal constraints extracted from the GEST. When the EventPlanner produces a planned action for an actor, the ActionsOrchestrator stores it as a pending request and checks whether all of its constraints are satisfied.

Constraint validation proceeds as follows. For each pending request, the orchestrator examines the request's constraint list. A \texttt{before} constraint on event $e_i$ targeting $e_j$ is satisfied when $e_i$ has been marked as fulfilled. An \texttt{after} constraint on $e_j$ referencing $e_i$ is satisfied when $e_i$ is fulfilled. A \texttt{same\_time} constraint between $e_i$ and $e_j$ requires both to be planned and ready simultaneously --- the orchestrator holds both until their partner is also ready, then dispatches them together. A \texttt{concurrent} constraint is satisfied immediately, allowing the events to overlap without synchronization. When all constraints for a request are satisfied, the request is marked as valid and eligible for dispatch.

The orchestrator enforces segment-based sequential execution: events in the same temporal segment are dispatched concurrently (as their constraints allow), while the next segment cannot begin until all events in the current segment have completed. Completion is tracked through the EventBus: when an action finishes executing in Stage~4, a \texttt{graph\_event\_end} signal is published, and the orchestrator subscribes to these signals to update its fulfillment tracking. Each fulfilled event may unblock pending requests from other actors whose constraints depended on it.

To prevent physical collisions at POIs, the orchestrator maintains per-POI execution queues. When two actors are dispatched to the same POI in the same segment, their actions are queued rather than executed simultaneously --- the second actor waits until the first completes and vacates the POI.

When a required POI is occupied by another actor, the orchestrator attempts \emph{actor displacement}: moving the occupying actor to a different location to free the POI. Displacement is only permitted when the occupying actor is in a safe state --- idle, awaiting constraints, or finished with their story. Actors cannot be displaced if they are on furniture (e.g., sitting or sleeping), in the middle of an interaction, or mid-chain with pending actions at the same POI. The displacement logic includes cycle detection: if displacing actor $A$ would require displacing actor $B$, who in turn requires displacing actor $A$, the system detects the cycle and aborts. When displacement succeeds, the displaced actor is moved to the best available POI --- preferring empty transit POIs in the same region, then non-busy POIs in the same region, then expanding to other regions --- and its next action is re-planned from the new location.

\begin{figure}[htbp]
\centering
\includegraphics[width=0.7\textwidth]{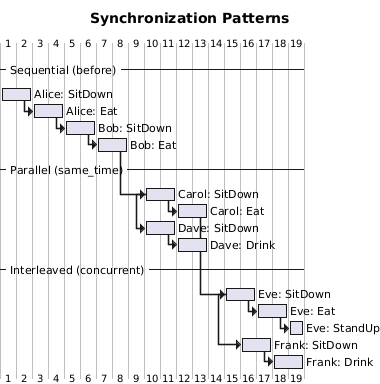}
\caption{\textbf{Synchronization patterns.} Three temporal coordination modes between actors. \emph{Sequential} (\texttt{before}/\texttt{after}): Bob's actions begin only after Alice's complete. \emph{Parallel} (\texttt{same\_time}): Carol and Dave begin simultaneously. \emph{Interleaved} (\texttt{concurrent}): Eve and Frank execute in overlapping time windows without synchronized start times.}
\label{fig:sync_patterns}
\end{figure}

\subsection{Synchronization Patterns}
\label{sec:sync_patterns}

The temporal constraint system produces three distinct synchronization patterns between actors, illustrated in Figure~\ref{fig:sync_patterns}.

\textbf{Sequential} (\texttt{before}/\texttt{after}). One actor's actions must complete before another actor's actions can begin. In Figure~\ref{fig:sync_patterns}, Alice's \texttt{SitDown} and \texttt{Eat} must finish before Bob can start his \texttt{SitDown}. This is the strictest form of coordination: the two actors' timelines do not overlap at all. Sequential constraints are the most common in practice, as they naturally express narrative causality --- one event triggering or enabling another.

\textbf{Parallel} (\texttt{same\_time}). Events from multiple actors begin at the same instant. The constraint is not limited to pairs: any number of events can share the same \texttt{same\_time} identifier. For instance, six actors sitting around a dinner table can all begin \texttt{SitDown} simultaneously by referencing the same constraint, and subsequently all begin \texttt{Eat} at the same time through a second shared constraint. The orchestrator holds all requests until every actor referencing the constraint is ready, then dispatches them together in the same tick. In Figure~\ref{fig:sync_patterns}, Carol and Dave illustrate the two-actor case; in practice, the same mechanism scales to the maximum number of actors per story (six in the GTASA corpus).

\textbf{Interleaved} (\texttt{concurrent}). Multiple actors' timelines overlap without requiring synchronized start times. In Figure~\ref{fig:sync_patterns}, Eve begins before Frank, and their activities run in parallel for part of the timeline. Like \texttt{same\_time}, this constraint supports any number of actors --- for example, six actors can independently begin activities in the same room within overlapping time windows. The \texttt{concurrent} constraint optionally accepts a \texttt{max\_delay} parameter that bounds the maximum time offset between the start of the events, providing partial control over their temporal proximity without requiring exact synchronization.

When no explicit cross-actor temporal relation is defined between events of different actors, those events are implicitly concurrent --- they execute independently, in whatever order the orchestrator dispatches them. Explicit constraints are only needed when the narrative requires a specific ordering or synchronization between actors.

Interaction actions (handshakes, hugs, kisses, conversations, object exchanges) require special coordination beyond temporal constraints. The first actor to reach an interaction event claims the interaction POI through the POICoordinator. When the second actor's matching event is dispatched, the system creates a clone POI at a slight offset from the primary POI, positions the second actor there, and both actors execute their interaction animations simultaneously. Once both actors complete the interaction (tracked via \texttt{graph\_event\_end} signals), the POICoordinator unregisters the interaction and releases both POIs.

\section{Stage 4: Execution and Capture}
\label{sec:stage_execution}

\begin{figure}[htbp]
\centering
\includegraphics[width=\textwidth]{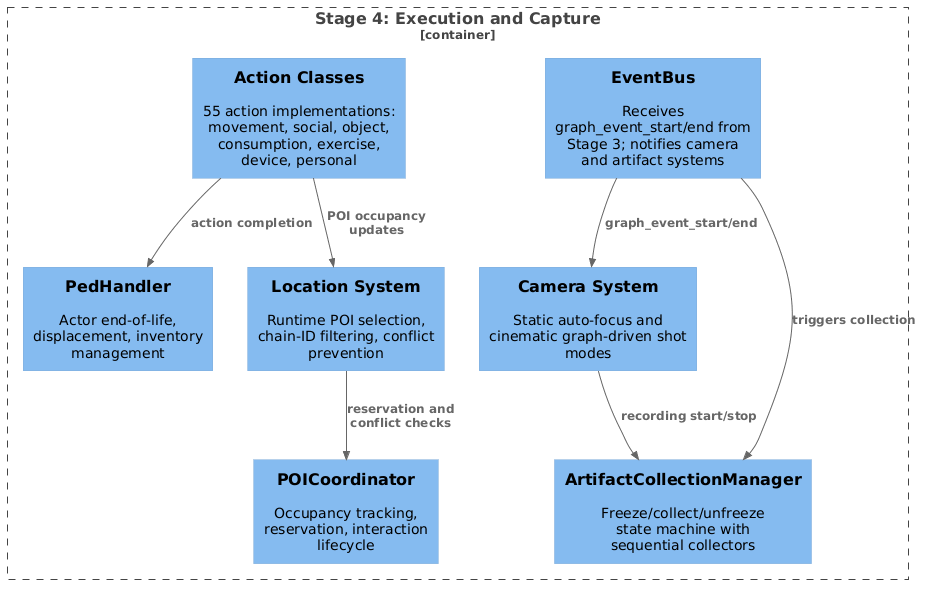}
\caption{\textbf{Stage~4 components.} Action classes execute animations and update actor state. The PedHandler manages actor end-of-life. The Location System and POICoordinator handle runtime POI selection and conflict prevention. The Camera System tracks actors across viewpoints. The ArtifactCollectionManager captures frame-aligned annotations via a freeze\slash collect\slash unfreeze state machine, triggered by EventBus signals from Stage~3.}
\label{fig:stage4_components}
\end{figure}

Stage~4 realizes the planned actions in the 3D world and simultaneously captures dense multi-modal annotations. When the ActionsOrchestrator dispatches an action, Stage~4 is responsible for executing the animation, managing the actor's physical state, selecting the correct camera viewpoint, and recording all artifacts --- RGB video, instance segmentation, spatial relations, event-frame mappings, and natural language descriptions --- in a single synchronized pass. Figure~\ref{fig:stage4_components} shows the components involved.

\subsection{Actor Control}
\label{sec:actor_control}

Action execution is implemented through a hierarchy of action classes. All 55 action classes inherit from a common base class that defines the execution protocol and the properties shared by all actions. Each action carries a reference to its \emph{Performer} (the actor executing it), a \emph{TargetItem} (the object or entity involved), a \emph{NextLocation} (the POI where the action occurs), a \emph{NextAction} pointer (the successor in an action chain), an optional \emph{ClosingAction} (to undo state, e.g., \texttt{StandUp} for \texttt{SitDown}), and a list of \emph{Prerequisites} (actions that must complete first). Each action receives a globally unique identifier via an incrementing counter.

When an action is dispatched, its \texttt{Apply} method is called. The base class implementation logs the action to the actor's history (enabling coreference tracking in the Logger, Section~\ref{sec:logger}), requests camera focus from the Camera System (unless the actor is marked as a background actor), registers the target location with the actor's state, and sets the current action name for tracking. Subclasses override \texttt{Apply} to trigger the appropriate game animation via the MTA animation system, specifying the animation library, animation name, duration, and loop behavior. Every action must signal completion by calling a global callback with a duration parameter; failure to call this callback causes the actor to deadlock, as the orchestrator will never dispatch the next event.

The base class also provides \texttt{pause} and \texttt{resume} hooks that subclasses override to handle context switches (Section~\ref{sec:cross_episode}). When an episode is paused, all actors in it have their current actions paused --- animations are suspended and timers are held. When the episode resumes, paused actions restart from their interrupted state. The base class provides empty implementations; complex actions like \texttt{Move} and \texttt{Wait} override these to preserve their internal state across context switches.

Actions maintain several state flags on the actor that are critical for physical plausibility. The \texttt{isOnFurniture} flag is set by \texttt{SitDown} and \texttt{GetOn}, and cleared by \texttt{StandUp} and \texttt{GetOff} --- while this flag is active, the actor cannot be displaced, preventing absurd situations like an actor being teleported away from a chair while sitting on it. The \texttt{isHoldingSpawnable} flag is set by \texttt{TakeOut} and cleared by \texttt{Stash}, tracking whether the actor is currently holding a spawnable object (phone, cigarette) that must be properly detached before any displacement.

Table~\ref{tab:action_taxonomy} organizes the 55 action classes into seven semantic categories. The action taxonomy covers the full range of activities available in the configured environments, from basic posture changes (sitting, standing, lying down) to complex multi-actor interactions (handshakes, conversations, object exchanges) and physical activities (gym equipment, dancing, tai chi).

\begin{table}[htbp]
\centering
\caption{Action taxonomy of the GEST-Engine, organized by semantic category.}
\label{tab:action_taxonomy}
\small
\begin{tabular}{@{}llp{8cm}@{}}
\toprule
\textbf{Category} & \textbf{Count} & \textbf{Actions} \\
\midrule
Movement & 5 & Move, GetOn, GetOff, SitDown, StandUp \\
Social & 10 & Talk, TalkPhone, AnswerPhone, HangUp, Handshake, Hug, Kiss, Laugh, Wave, LookAt \\
Object interaction & 10 & PickUp, PutDown, TakeOut, Give, Receive, Stash, TurnOn, TurnOff, PunchDesk, LayOnElbow \\
Consumption & 6 & Eat, Drink, Cook, Smoke, SmokeIn, SmokeOut \\
Physical activity & 8 & BarbellWorkOut, BenchpressWorkOut, DumbbellsWorkOut, JogTreadmill, PedalGymBike, TaiChi, Punch, Dance \\
Device & 3 & OpenLaptop, CloseLaptop, TypeOnKeyboard \\
Personal & 4 & WashHands, Sleep, LookAtTheWatch, Wait \\
\bottomrule
\end{tabular}
\end{table}

\textbf{The Move action.} \texttt{Move} is the most complex action class. It supports three locomotion modes --- walk, run, and skate --- selected based on the actor's skin model and the distance to the target. When dispatched, the Move action computes a path using the A* algorithm, provided by the \texttt{ml\_pathfind} module, over the pathfinding graph defined in the World Editor (Section~\ref{sec:world_editor}). The resulting path is a sequence of waypoints, each materialized as an invisible collision marker in the game world. The actor walks toward each marker in sequence; a polling loop with an adaptive timeout checks whether the actor has reached the marker, and advances the path when collision is detected. If the actor fails to reach a waypoint within a configurable timeout, the system teleports the actor to that waypoint as a fallback, preventing indefinite stalls from pathfinding failures or animation glitches.

For actor-targeted moves (where the destination is another actor rather than a fixed POI), the Move action periodically rechecks the target's position. If the target has moved more than 3 meters from the originally planned destination, the path is recalculated. This enables scenarios such as one actor walking toward another who is themselves moving.

For cross-episode movement (when the target POI is in a different episode), the Move action decomposes the journey into \emph{context segments} --- a sequence of intermediate POIs representing the path through linked episodes. At each segment boundary, the actor is teleported to the next episode's connection POI, their interior is switched, and any carried objects have their interiors updated to match. A global lock mechanism prevents concurrent pathfinding operations for the same actor, avoiding race conditions when multiple move requests are queued.

\textbf{The Wait action.} \texttt{Wait} implements a polling-based synchronization protocol for two-actor interactions. When two actors must perform an interaction (e.g., a handshake), both are first dispatched to a \texttt{Wait} action at the interaction POI. One actor is designated as \emph{active} and the other as \emph{passive}. Both enter a polling loop that periodically checks whether the other actor is waiting for the same interaction. Once both actors are co-located and waiting, they begin a mutual gaze behavior --- continuously rotating to face each other. The passive actor then signals readiness; when the active actor detects this, it triggers the completion callback, causing the orchestrator to dispatch the actual interaction action for both actors. This protocol ensures that both actors are physically present and facing each other before the interaction animation begins.

\textbf{Interaction actions.} Handshakes, hugs, kisses, conversations, and object exchanges inherit from a specialized base class that implements a four-step synchronization protocol: (1) both actors rotate to face each other, (2) a short rendering delay allows the game engine to update the visual models, (3) the inter-actor distance is verified and corrected along the facing direction to match the interaction type, and (4) both animations are triggered simultaneously via a callback. The performer's position is never modified during correction --- only the target actor is repositioned, preserving the performer's authority over the interaction location.

\textbf{Object transfer.} The \texttt{Give} and \texttt{Receive} actions implement a coordinated object transfer protocol. After the four-step synchronization, both actors play the transfer animation. The giver's callback detaches the object and removes it from their inventory; shortly after, the receiver's callback attaches the object and adds it to their inventory. The slight delay between the two callbacks ensures the object is fully detached before re-attachment. The object's chain-ID and origin location are transferred along with it, so that downstream actions (such as \texttt{PutDown}) know where the object belongs.

\textbf{Actor end-of-life.} When an actor completes all events in their temporal chain, the PedHandler marks them as finished. The story waits until all actors have completed before terminating. Finished actors are displaced to transit POIs to avoid blocking POIs needed by actors still executing. The PedHandler also manages actor cleanup at story end --- destroying ped instances, releasing skin assignments, and clearing inventory objects.

\subsection{Location System}
\label{sec:location_system}

The Location System manages the spatial semantics of the simulation at runtime. It comprises two components: POIs (Points of Interest) and Regions.

A \emph{POI} is the atomic unit of spatial interaction. Each POI stores a 3D position, rotation, interior identifier, a unique location identifier, and references to the actions available at that location (inherited from the template system described in Section~\ref{sec:world_editor}). At runtime, the POI maintains its occupancy state, tracks which actor is currently present, and provides the mechanism for physically placing actors --- setting the actor's position, rotation, and interior, marking the POI as busy, and updating the actor's location tracking data. When the EventPlanner selects a POI for an action, the Location System resolves the chain-ID to a concrete object instance: given an event's object identifier and the actor's current chain-ID, the POI looks up the corresponding entry in the output maps from Stage~2 and returns the specific simulator object bound to that chain. This resolution is what connects the abstract GEST event (``Alice sits on a chair'') to a particular physical chair in a particular location in the episode.

A \emph{Region} is a polygon defined in the World Editor that represents the spatial boundary of a room or area. Regions group POIs, objects, and cameras into coherent spatial units. The system uses point-in-polygon testing to determine which region an actor is in, enabling location-aware behaviors: when an actor enters a region for the first time, the Logger describes the room and its contents; region boundaries determine which objects are spatially associated with which areas; and in static camera mode (Section~\ref{sec:camera_system}), the Camera System selects the appropriate viewpoint from the region's predefined camera positions.

Runtime conflict prevention is handled through several complementary mechanisms. Each POI maintains an occupation flag: when an actor begins an action at a POI, the flag is set, preventing other actors from being dispatched there simultaneously. Actors also carry a reserved location identifier that signals their intent to move to a POI before they arrive, preventing race conditions where two actors are dispatched to the same POI in rapid succession. For pick-up-able objects, the system tracks which actor has picked up each object, preventing double pickups.

The POICoordinator --- introduced in Stage~3 for interaction coordination --- continues to operate at runtime. It tracks active interactions, manages clone POIs for two-actor interactions, and provides atomic group reservation for synchronized constraint groups (\texttt{same\_time} and \texttt{concurrent}). The atomic reservation is important: before dispatching a group of actors that must start simultaneously, the POICoordinator verifies that \emph{all} required POIs are available, reserving them as a single atomic operation. If any POI is occupied, the entire group is deferred rather than partially dispatched. When an interaction completes (both actors signal \texttt{graph\_event\_end}), the POICoordinator unregisters the interaction, cleans up clone POIs, and releases both the primary and clone locations.

\subsection{Camera System}
\label{sec:camera_system}

The Camera System provides automated cinematography during simulation. The GEST's optional \texttt{camera} section allows per-event camera commands: each event identifier can be associated with a recording directive (\texttt{start} or \texttt{stop}) and, optionally, a shot specification. Recording directives control the artifact collection system --- a \texttt{start} command publishes an \texttt{artifact\_start\_collection} signal to the EventBus, activating the freeze\slash collect\slash unfreeze cycle, while a \texttt{stop} command halts collection. This decoupling allows selective recording of specific portions of a story rather than capturing the entire simulation.

The Camera System operates in one of two modes.

In \emph{static mode} (the default, used for all videos in the GTASA corpus), the camera automatically switches focus between actors on a timed rotation. When an action is dispatched, it requests camera focus; the Camera System queues the request and assigns focus to actors in order, selecting the camera position defined for the actor's current region in the World Editor. The region's camera selection prefers viewpoints where the focused actor is within the field of view. This mode produces surveillance-camera-style footage that covers all actors without manual direction and is the mode used in all experiments reported in a companion evaluation.

\emph{Cinematic mode} is a work-in-progress extension that enables graph-driven camera control. In this mode, the per-event camera commands in the GEST's \texttt{camera} section specify not only recording directives but also shot types for each event: \texttt{follow} (continuous tracking behind an actor), \texttt{close up} (one-time positioning near an actor), \texttt{over shoulder} (behind one actor looking at another), \texttt{two shot} (framing both actors), \texttt{show} (centering on an object), and \texttt{static} (using the region's predefined camera). The infrastructure for parsing and dispatching these commands is implemented, but the mode has not yet been fully tested across all action types and context switch scenarios. Completing and validating cinematic mode --- enabling the procedural generator and the agentic system (Section~\ref{sec:agentic}) to produce GESTs with camera direction --- is left as future work.

Both modes handle context switches transparently: when the focused actor moves to a different episode, the Camera System coordinates the pause-fade-switch-resume sequence described in Section~\ref{sec:cross_episode}, ensuring that the spectator always sees the active episode.

\subsection{Cross-Episode Logic}
\label{sec:cross_episode}

When a story spans multiple episodes (e.g., a house interior and a garden), actors may need to move between 3D contexts. This poses unique challenges rooted in the architecture of the GTA San Andreas engine. The game uses a streaming mechanism that loads and renders 3D geometry based on the player's current interior --- only entities in the active interior are rendered and animated. If actors in one interior continue performing animations while the rendering context switches to another interior, the results are undefined: animations may reset, positions may desynchronize, and objects may disappear or appear in incorrect states.

From a cinematographic perspective, we also do not want to miss what actors are doing in the source context. If Alice is eating in the kitchen while Bob moves to the garden, we want to freeze Alice's action at the exact frame of the context switch, show Bob arriving in the garden, and later resume Alice from exactly where she was paused --- analogous to the time-freeze technique used in film editing when cutting between simultaneous scenes. This is why the system pauses and resumes actions rather than letting them run unsupervised.

To control the rendering context, the system uses a \emph{spectator}: the actual MTA player entity (the one that would normally be controlled by a human player through the client) is repurposed as an invisible camera carrier. The spectator's visual model is made fully transparent, its collisions are disabled, and it is moved between interiors to trigger the engine's streaming mechanism. When the spectator enters a new interior, the engine begins streaming that interior's geometry, making it visible for screen capture. All actors in the simulation are server-side peds, not player-controlled --- the spectator exists solely to direct the rendering.

The context switch proceeds through five phases:

\begin{enumerate}
    \item \textbf{Pause request}: all actors in the current episode receive a pause request. Each actor's current action is responsible for reaching a safe pause point --- the Move action stops at the next waypoint, the Wait action suspends its polling loop, and furniture-bound actions (sitting, sleeping) remain in their current state. The system polls periodically until all actions confirm they have paused, with a timeout to prevent indefinite waits if an action fails to respond.

    \item \textbf{Fade out}: once all actions are confirmed paused, the screen fades to black for all spectators. This hides the interior transition from the viewer.

    \item \textbf{Teleport and interior switch}: the moving actor is teleported to the target POI in the new episode. Their interior identifier is updated to match the target episode, and any objects the actor is carrying --- both picked-up fixed objects and spawnable objects attached to the actor's hand --- have their interiors updated as well. For linked episodes (connected through door POIs in the World Editor), the actor walks to the connection POI before the context switch begins, producing a spatial progression that mirrors a physical doorway. For unlinked episodes (different episode groups), the teleportation is direct, analogous to a scene cut in a movie.

    \item \textbf{Camera reassignment}: the Camera System switches the spectator to the new episode's interior. In static mode, this involves selecting the appropriate region camera in the target episode and assigning focus to the arriving actor. In cinematic mode, the camera command associated with the next event determines the shot. The Camera System tracks which episode is currently focused and maintains this state across multiple consecutive context switches.

    \item \textbf{Resume and fade in}: the target episode becomes active, and all actors in it (including those who were already there from a previous context switch) resume their paused actions. The screen fades back from black. If the arriving actor has a pending action in the new context, it is re-planned from the new location to account for any changes in POI availability since the original planning.
\end{enumerate}

When multiple actors need to switch contexts in rapid succession --- for instance, two actors both moving from a house to a garden in the same temporal segment --- the context switches are queued. Each switch completes its full pause-fade-teleport-resume-fade cycle before the next begins, preventing visual artifacts from overlapping transitions. The queue ensures that the spectator sees a coherent sequence of scene changes rather than a jumbled mix of interiors.

\subsection{Multi-Modal Artifact Collection}
\label{sec:artifact_collection}

The artifact collection system captures dense, frame-aligned annotations during simulation at zero marginal cost. The system is architected around the adapter pattern: the ArtifactCollectionManager, the collection orchestration logic, and the collector interfaces are fully game-agnostic, while MTA-specific functionality is isolated in dedicated adapter layers. The simulation freeze adapter pauses all game timers and animations (preserving their callbacks and remaining durations for later restoration); the screenshot adapter interfaces with the native C++ capture module; and the render mode controller toggles between normal rendering and the segmentation shader. This separation means that porting the artifact collection system to a different 3D platform would require implementing only the adapter layer, without modifying the collection orchestration or the individual collectors.

\begin{figure}[p]
\centering
\includegraphics[width=\textwidth]{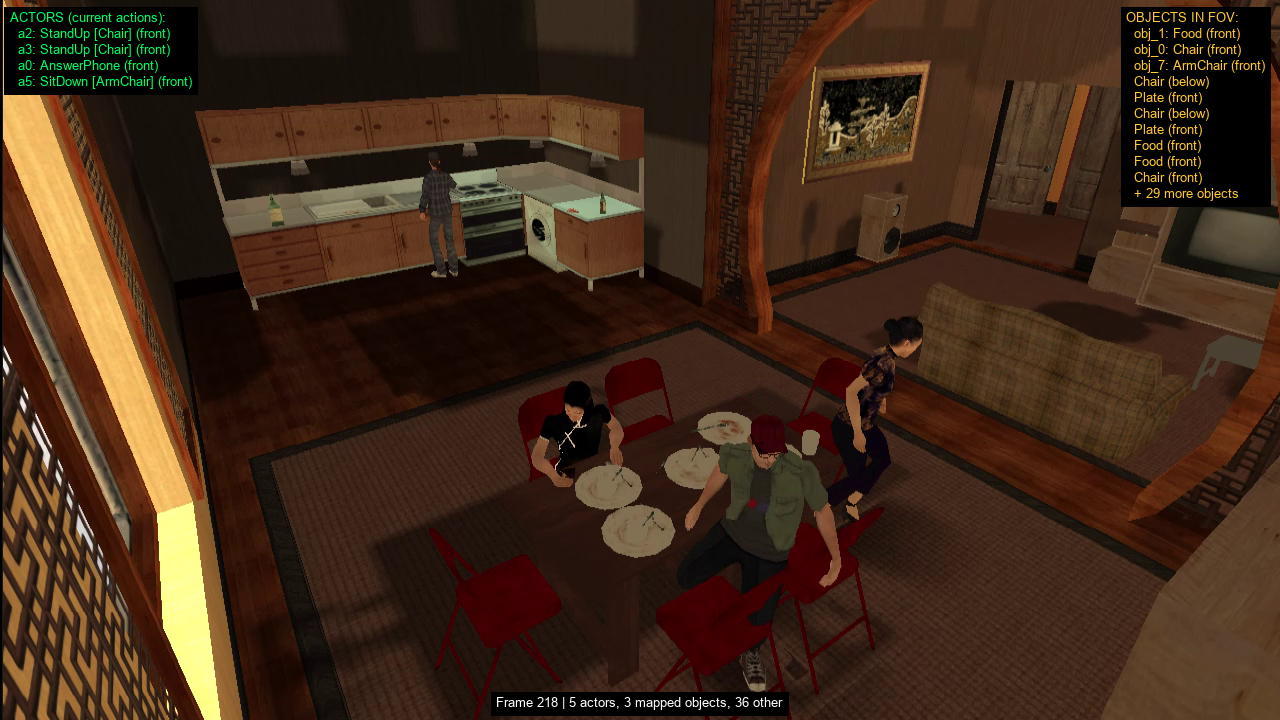}\\[6pt]
\includegraphics[width=\textwidth]{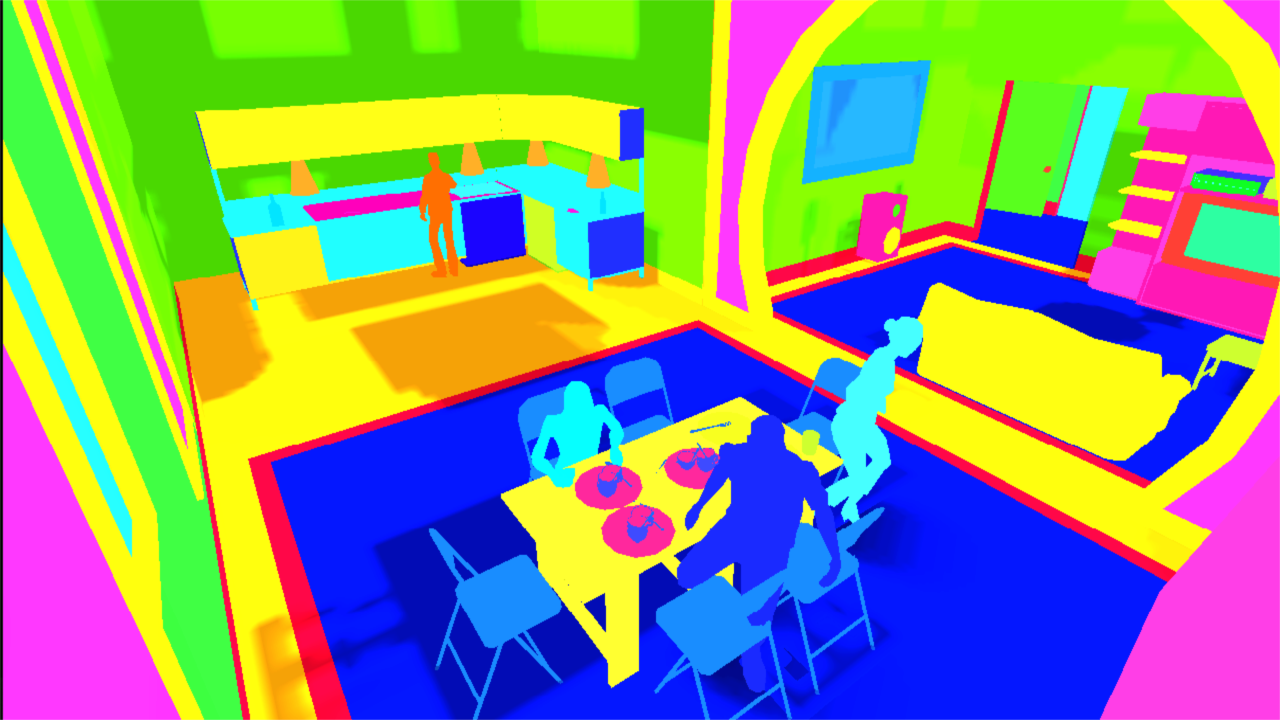}\\[6pt]
\includegraphics[width=\textwidth]{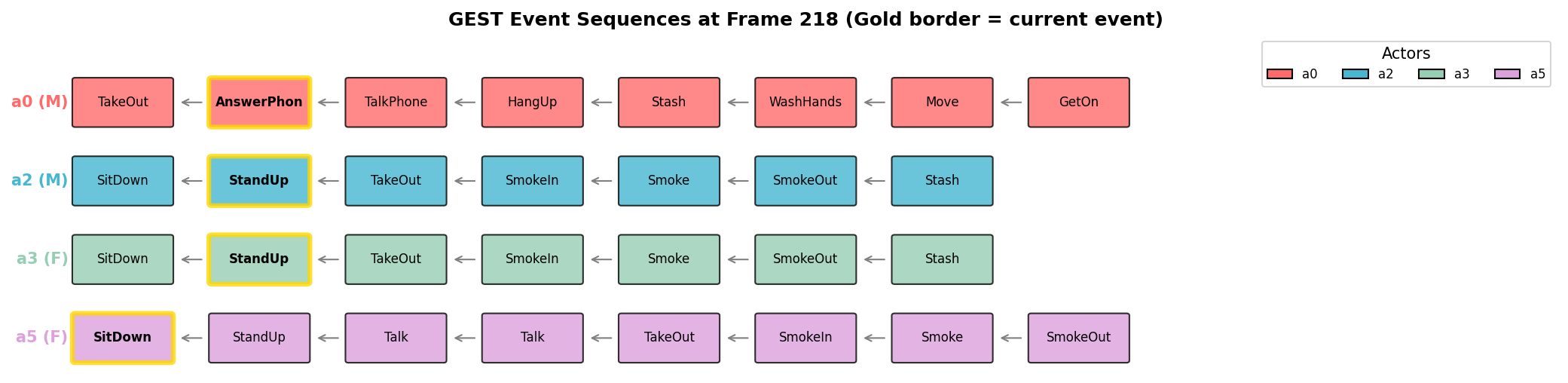}
\caption{\textbf{Multi-modal outputs.} Top: RGB frame with active events and camera-relative direction annotations. Middle: instance segmentation via HLSL texture hashing --- each unique texture maps to a distinct color. Bottom: GEST event sequences across four actors; gold border indicates the currently active event.}
\label{fig:multimodal_outputs}
\end{figure}

The ArtifactCollectionManager coordinates the capture process through a freeze\slash collect\slash unfreeze state machine: at each capture point, the simulation is frozen via the freeze adapter, each registered collector captures its modality in sequence, and the simulation is unfrozen. This ensures frame-consistent multi-modal capture --- all modalities see exactly the same simulation state. The manager excludes its own timeout timer from the freeze to detect collector failures. Beyond the RGB, instance-segmentation, spatial-relation, and temporal tracks described below, the system captures dense per-pixel depth (Section~\ref{sec:depth}) and per-actor skeletal pose (Section~\ref{sec:pose}), and projects every tracked entity onto the image plane as 2D bounding boxes and screen-space coordinates (Section~\ref{sec:screenspace}). New collectors can be registered at initialization time, and the system is designed for extension to further modalities such as optical flow and surface normals. Figure~\ref{fig:multimodal_outputs} shows three of the primary artifact types captured by the system.

\subsubsection{RGB Capture}
\label{sec:rgb_capture}

The earliest version of the capture system used MTA's built-in \texttt{takeScreenShot} API, which captures pixels on the client and transmits them to the server over the network. This approach suffered from significant latency and frame drops at higher resolutions, because the full JPEG pixel data had to travel from client to server for each frame --- a round-trip that introduced unpredictable delays and made frame-level synchronization with the other collectors impossible.

To address these limitations, we developed a native C++ capture module (approximately 8,500 lines) built on the Windows Desktop Duplication API (DXGI). This API, used by professional screen-sharing tools such as AnyDesk and TeamViewer as well as game streaming services, acquires frames directly from the GPU's front buffer with minimal latency. All operations --- frame acquisition, cropping, resizing, and video encoding --- are performed on the dedicated GPU, avoiding CPU-bound bottlenecks. The switch to Desktop Duplication eliminated the network round-trip entirely and enabled deterministic frame-level alignment with the other artifact collectors.

The capture pipeline begins with window detection: the module locates the MTA game window by enumerating all windows and matching against the title string, caching the window handle for subsequent frames. The window's screen rectangle is tracked on every frame to handle resizing or repositioning. Because MTA renders in a windowed application, the raw screen capture includes the window's title bar and a watermark footer. These are cropped at the GPU level using Direct3D's \texttt{CopySubresourceRegion} with a bounding box that excludes a configurable number of pixels from the top and bottom.

The only synchronous operation in the pipeline is the frame capture itself --- acquiring the GPU texture from the front buffer while the simulation is frozen. Everything else is asynchronous: captured frames are cloned as GPU textures and pushed to a thread-safe task queue, where a dedicated worker thread handles resizing, encoding, and disk I/O in the background. The callback to the Lua layer is invoked immediately after the GPU texture is cloned, allowing the simulation to unfreeze without waiting for encoding or file writes to complete. This decouples the capture rate from the encoding rate, preventing frame drops during computationally expensive operations. The module supports three image formats: JPEG with configurable quality for RGB frames, standard 32-bit PNG, and indexed 8-bit PNG with at most 256 palette colors (used for segmentation masks, where nearest-neighbor interpolation preserves exact color values during resizing). When captured frame dimensions differ from the target output, the module resizes via a GPU-accelerated video processor when available, falling back to CPU-based bilinear interpolation. For video output, frames are encoded to H.264 via Windows Media Foundation directly on the GPU.

A modality manager coordinates multiple capture modes within the same session. Each modality --- RGB, segmentation, depth --- is registered with its own video encoder, target dimensions, and output path. During the freeze\slash collect\slash unfreeze cycle, the ArtifactCollectionManager triggers each modality in sequence: the render mode is switched (e.g., normal rendering for RGB, segmentation shader for instance segmentation), the frame is captured, and the mode is restored before the next modality.

The result is that for each frame identifier, the system produces a synchronized set of artifacts on disk: an RGB image (JPEG), a segmentation mask (indexed PNG), and a spatial relations graph (JSON) --- all captured from exactly the same frozen simulation state and tagged with the same frame number. Combined with the event-frame temporal alignment (Section~\ref{sec:temporal_alignment}), this per-frame synchronization enables the spatiotemporal probing experiments reported in a companion evaluation, where the exact 3D positions and pairwise relations of every entity at every frame serve as ground truth for evaluating what video encoders actually learn about spatial structure.

\subsubsection{Instance Segmentation}
\label{sec:instance_seg}

Instance segmentation is produced by an HLSL pixel shader that replaces the game's standard rendering with a flat-color mode. The shader operates at the texture level: each unique texture applied to a surface receives a distinct color via FNV-1a hashing of the texture identifier. A global texture-to-color mapping is maintained across frames and written incrementally to a JSON file, enabling post-hoc identification of which color corresponds to which game texture. The segmentation collector activates this shader mode before capture and deactivates it afterward, waiting two render frames for the mode switch to propagate through the rendering pipeline.

Because the segmentation operates at the texture level rather than the object level, each material within a composite object is segmented individually. An armchair, for instance, may consist of wooden legs and handles rendered with a wood texture and a textile seat and backrest rendered with a fabric texture --- these appear as two distinct segments in the output, not one unified object mask. Similarly, a kitchen cabinet made of glass and wood panels produces separate segments for each material. This texture-level granularity is finer than traditional instance segmentation, which typically produces one mask per object. The texture identifiers can be mapped to semantic categories through the game's object and texture catalogues, enabling post-hoc aggregation of textures into object-level or category-level masks when needed.

During segmentation capture, the spectator (Section~\ref{sec:cross_episode}) must be hidden from the frame. In normal RGB rendering, the spectator is transparent and does not appear in captures. However, when the segmentation shader is active, it renders all textures with flat colors regardless of transparency --- making the spectator visible. To prevent this, the system repositions the spectator behind the camera before each segmentation capture: it computes the camera's view direction, inverts it, and places the spectator several units behind the camera, outside the view frustum.

A further challenge arises from the MTA rendering pipeline's treatment of lighting and shadows. In GTA San Andreas, dynamic lights and shadows are implemented as semi-transparent textures that overlay the underlying surface textures. When the segmentation shader is active, these light and shadow textures receive their own hash-based colors, which overlap and partially obscure the object textures beneath them. This can alter the segmentation output in unpredictable ways, particularly in outdoor scenes where dynamic lighting is prevalent. Mitigating this artifact --- either by filtering light textures from the shader or by disabling dynamic lighting during segmentation capture --- remains an open engineering challenge.

\subsubsection{Unified In-Client Capture and Dense Depth}
\label{sec:depth}

\begin{figure}[htbp]
\centering
\includegraphics[width=\textwidth]{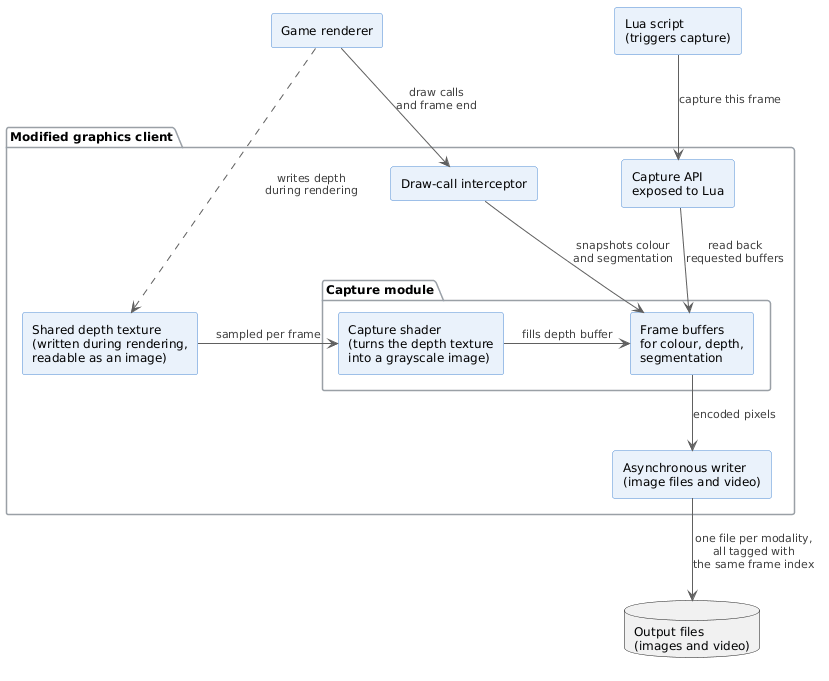}
\caption{\textbf{Unified multi-modal capture path inside the graphics client.} During normal rendering, the game writes depth into a shared texture that the capture module later reads as an image, and a draw-call interceptor snapshots the color and segmentation buffers directly. One scripting-side call triggers readback of the requested buffers; an asynchronous writer produces the output files, all tagged with the same frame index.}
\label{fig:capture_path}
\end{figure}

The RGB and segmentation collectors described above operate at the external capture boundary: within a single freeze window the client is switched between its normal render path and the segmentation shader, and each resulting frame is grabbed from the screen. This route has two costs --- each modality requires its own render-mode swap and its own re-rendered frame, and depth cannot be recovered at all, since a screen capture only sees the final color image while the engine's depth buffer stays internal to the GPU. To lift both limitations, the capture boundary can be moved inside the graphics client itself. A dedicated capture module, instantiated at client startup, owns its own color, depth, and segmentation buffers and hooks into the graphics pipeline at three points: device initialization, each game draw call, and frame end. All requested modalities are then produced as side effects of the engine rendering a single frame --- no render-mode swaps, no repeated draws of the scene --- and the whole bundle is delivered to the scripting layer through one blocking call. Figure~\ref{fig:capture_path} shows the components and the flow of data from the game renderer to disk.

Color and depth are acquired through two small extensions to the rendering path. Color is snapshot at frame end: because the game draws into a back buffer that is reused for the next frame, the module copies the back buffer into its own color render target before its contents are discarded, so the most recent fully-rendered color frame is always resident when the scripting side asks for it. Depth is acquired through an \texttt{INTZ}-format depth texture --- a depth-stencil surface that the rasterizer can write to as it renders the scene and that a shader can later read back as an image. At startup the module installs this texture as the device's depth buffer (restoring the original at shutdown); the engine's normal rendering then writes depth into it as a by-product of drawing the scene, and on capture a small fullscreen-quad shader copies it into the module's depth render target. Segmentation fits the same construction by design: a per-draw-call hook can emit each rendered primitive a second time into a dedicated segmentation target with a constant-color shader keyed on the primary texture of that draw. A single scripting entry point reads back the requested render targets and dispatches asynchronous PNG or JPEG writes, stamping every file with the same integer frame index so that all modalities from one frame share the same name; a companion entry point pipes any modality through a hardware H.264 encoder, yielding a video stream synchronized with the per-frame annotation tracks.

\begin{figure}[htbp]
\centering
\begin{minipage}[t]{0.49\textwidth}
\centering
\includegraphics[width=\textwidth]{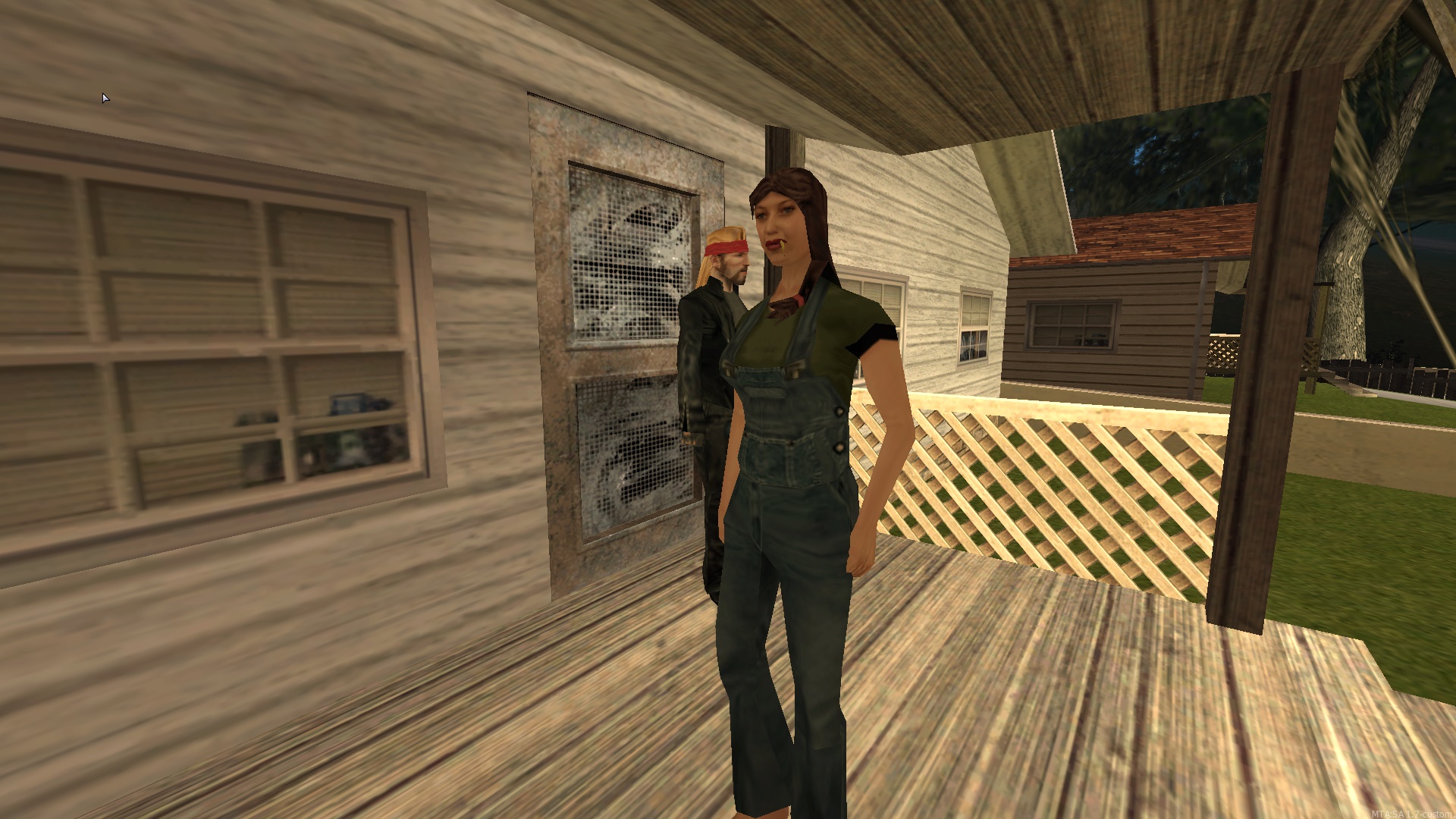}\\[1pt]
\includegraphics[width=\textwidth]{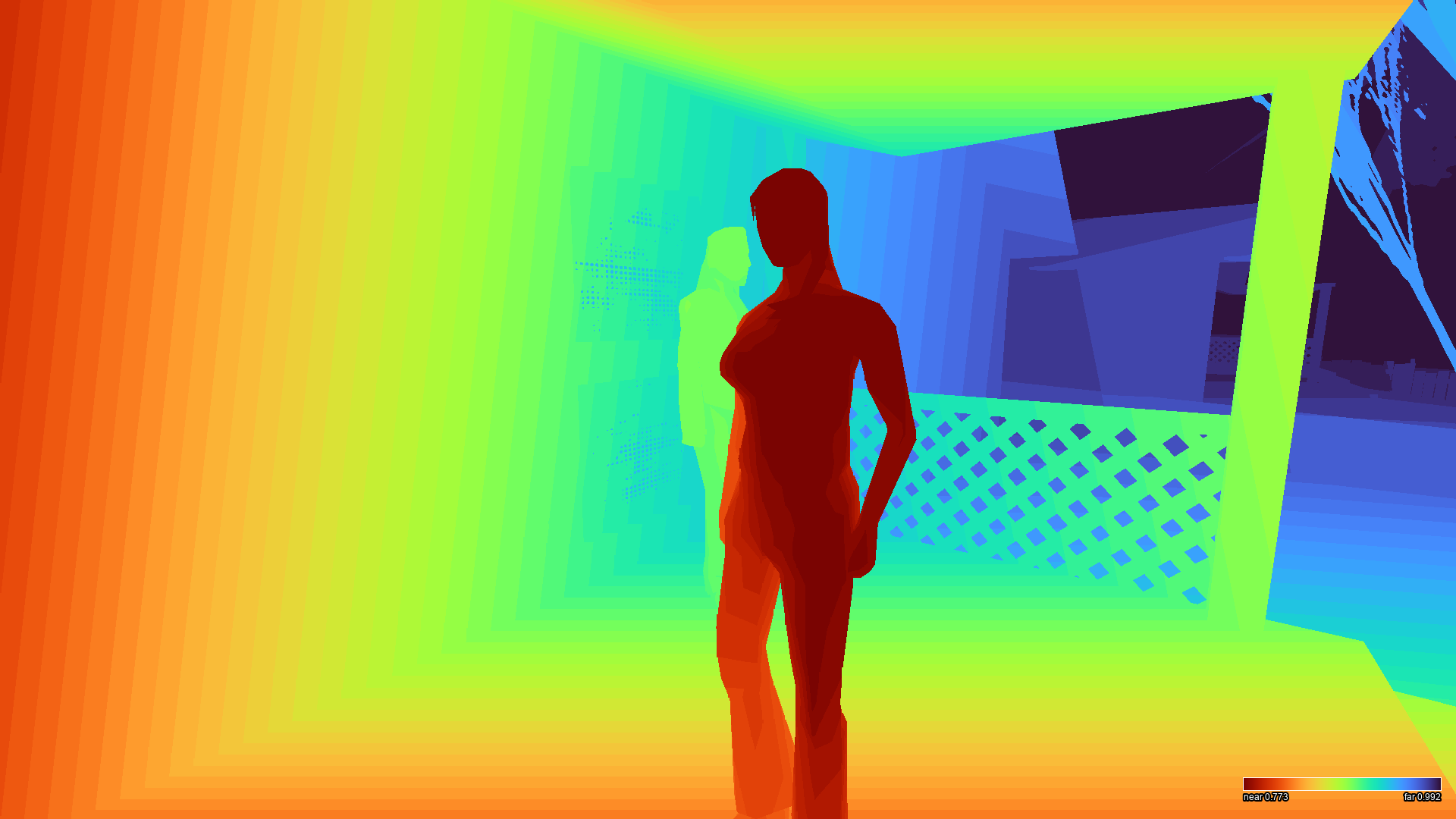}\\[2pt]
{\footnotesize (a) Porch (exterior).}
\end{minipage}\hfill
\begin{minipage}[t]{0.49\textwidth}
\centering
\includegraphics[width=\textwidth]{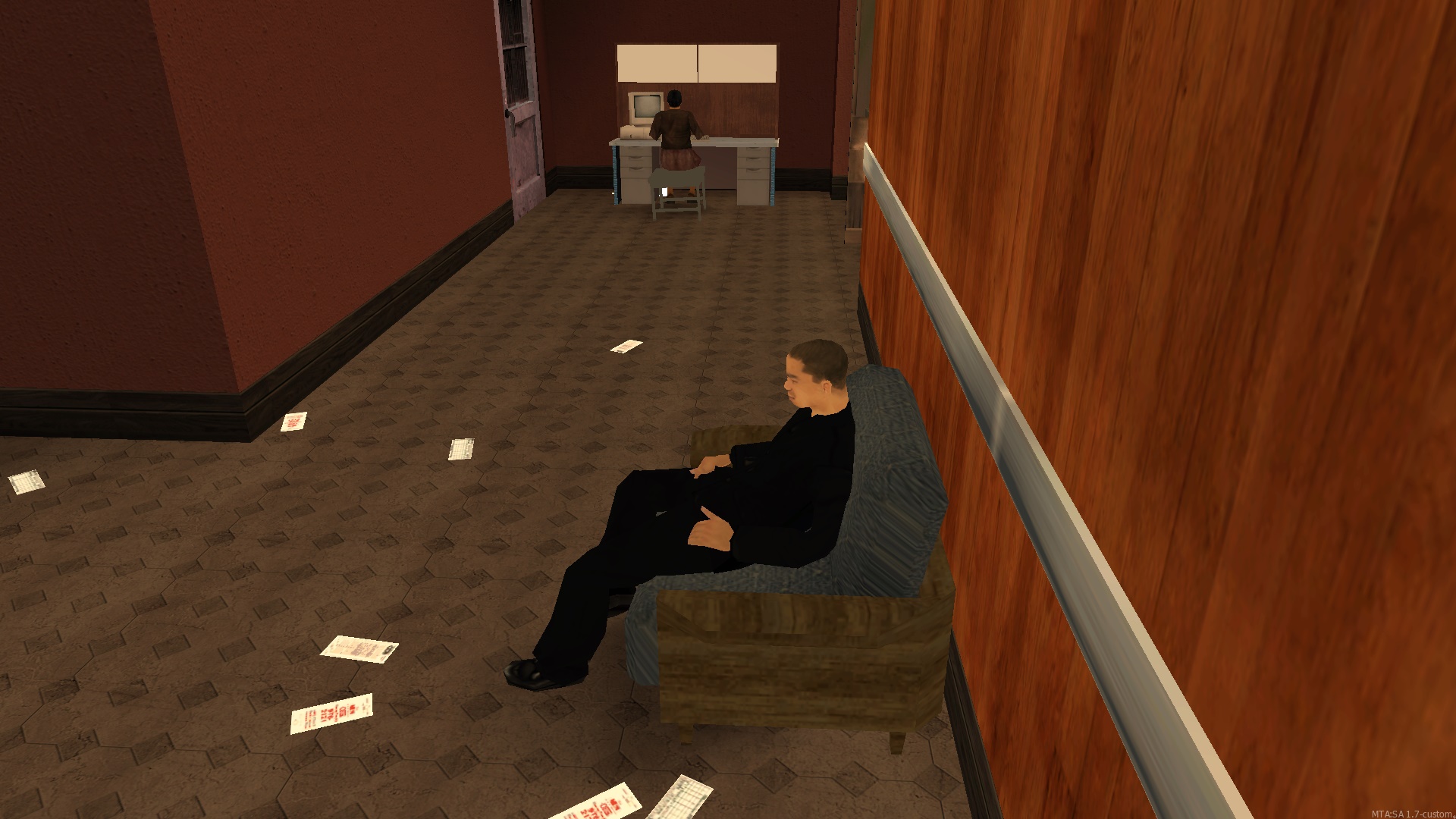}\\[1pt]
\includegraphics[width=\textwidth]{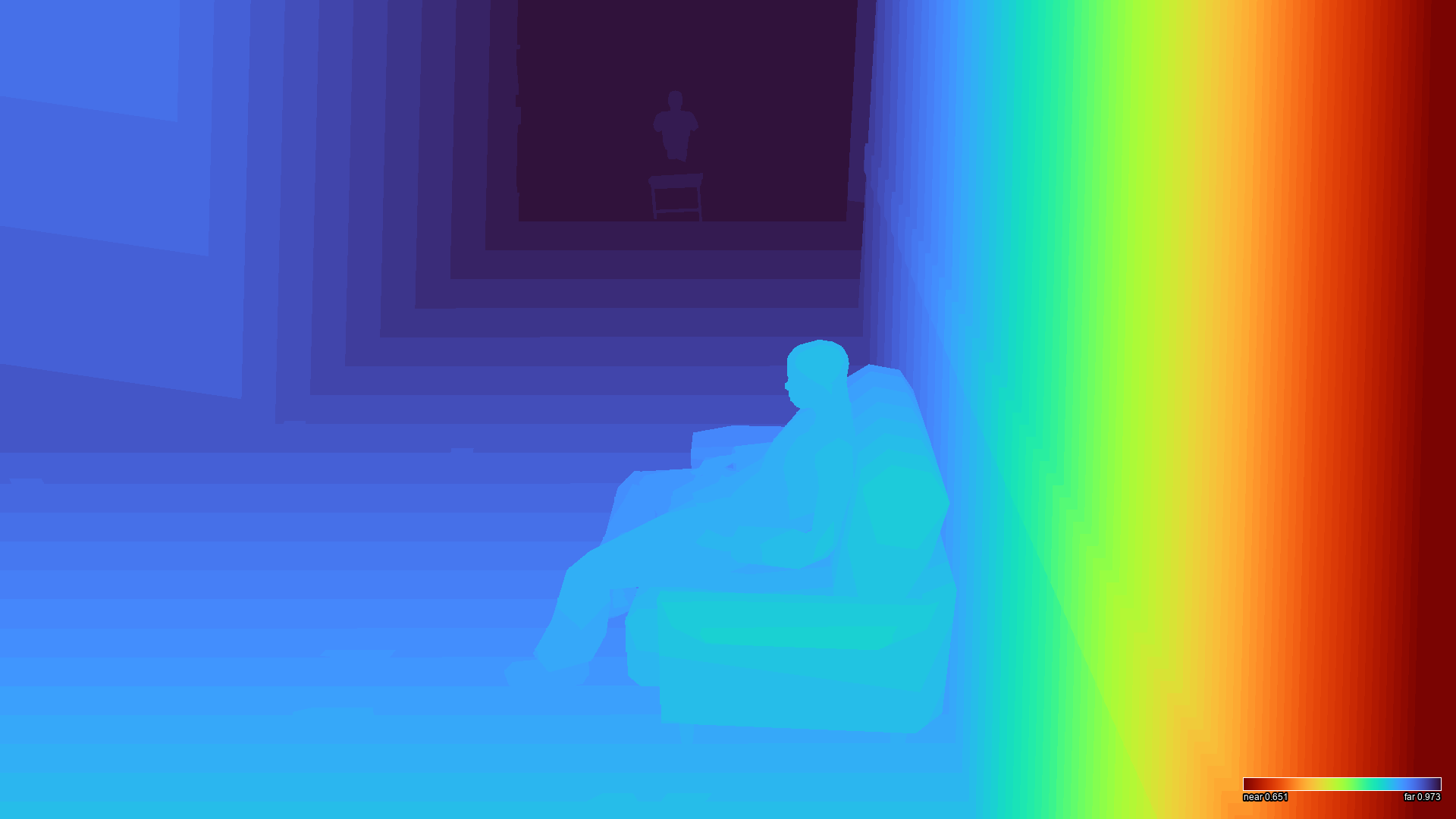}\\[2pt]
{\footnotesize (b) Study (interior).}
\end{minipage}

\vspace{6pt}

\begin{minipage}[t]{0.49\textwidth}
\centering
\includegraphics[width=\textwidth]{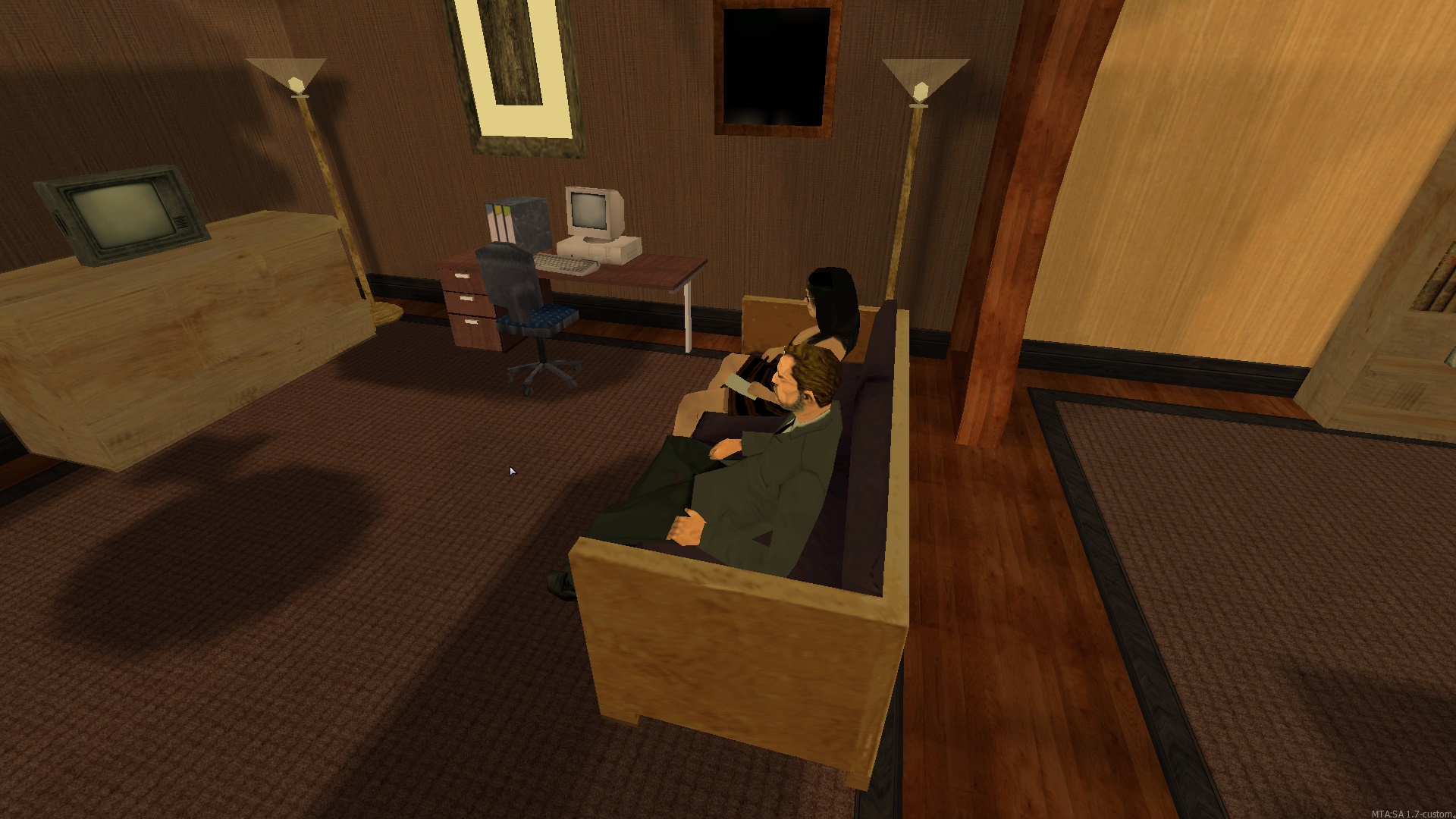}\\[1pt]
\includegraphics[width=\textwidth]{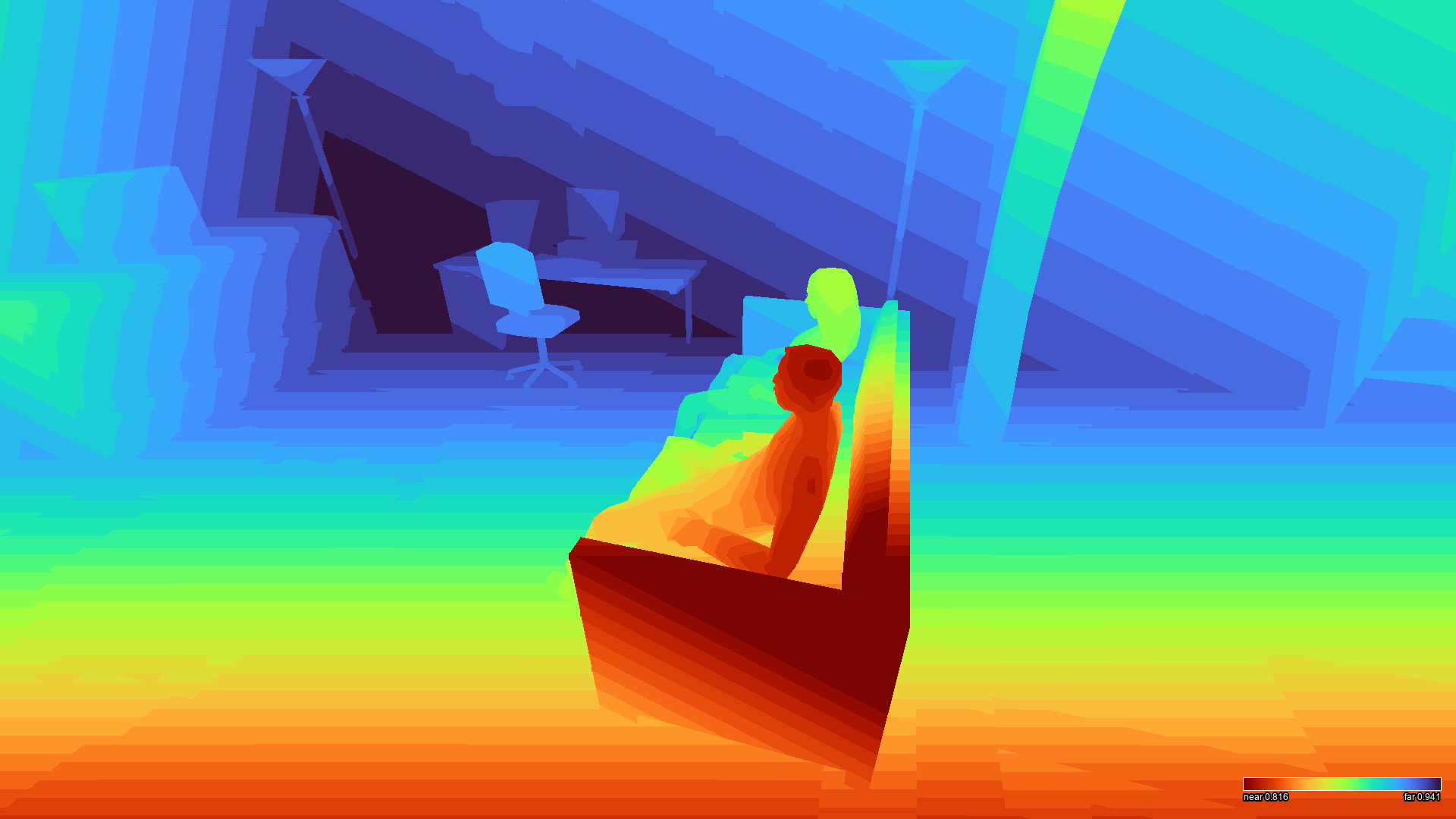}\\[2pt]
{\footnotesize (c) Living room.}
\end{minipage}\hfill
\begin{minipage}[t]{0.49\textwidth}
\centering
\includegraphics[width=\textwidth]{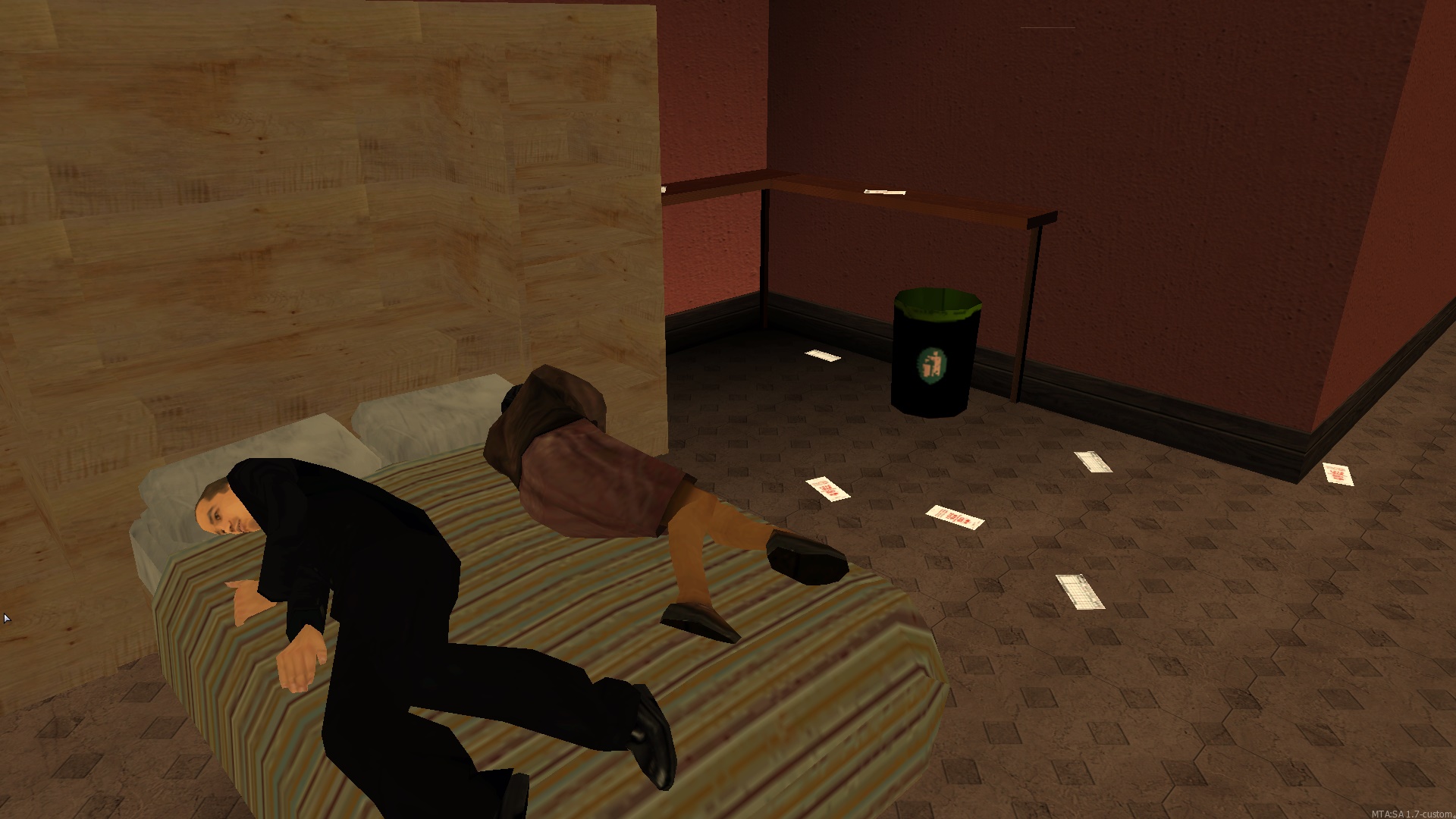}\\[1pt]
\includegraphics[width=\textwidth]{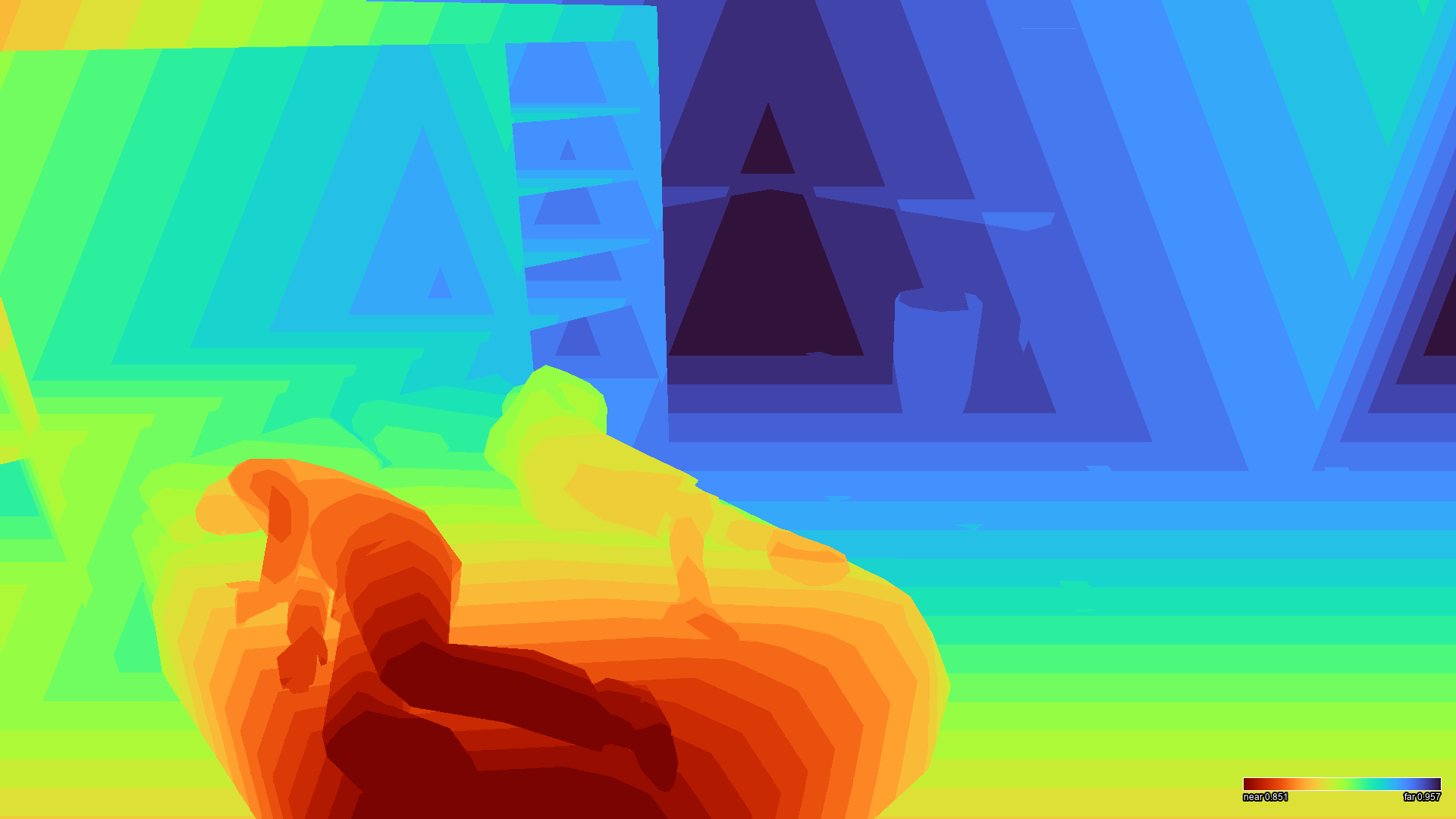}\\[2pt]
{\footnotesize (d) Bedroom.}
\end{minipage}
\caption{\textbf{RGB and dense depth from the same rendered frames}, across a porch exterior, an indoor study, a living room, and a bedroom --- depth is produced consistently alongside RGB regardless of scene type. Depth is shown with an inverted percentile-normalized colormap (near\,$=$\,red, far\,$=$\,blue); the on-disk files store the raw non-linear post-projection depth described in the text.}
\label{fig:depth_gallery}
\end{figure}

For every requested frame the depth channel is written to disk as a grayscale image at the same resolution as the RGB output, and the two are pixel-aligned: any color-to-depth comparison is a straightforward index lookup, requiring neither reprojection nor interpolation. The stored value at each pixel is the post-projection depth as written by the rasterizer, in the $[0,1]$ range the graphics API uses natively; linearization to metric distance is a scalar transformation that depends only on the near and far planes of the projection matrix, both constants for a given run and logged once at capture start. A downstream consumer therefore recovers metric distance with a single formula, and may work either in the non-linear form --- which devotes more dynamic range to the region near the camera, where the scene is most detailed --- or in the linear form convenient for depth regression. The non-linear encoding concentrates resolution near the camera: depth is reliable at the scales the scripted multi-actor scenarios typically cover --- a few tens of meters around the camera, inside a single room, courtyard, or stretch of street --- and loses precision beyond the far plane. Dense synthetic depth is not new in itself --- CARLA~\cite{dosovitskiy2017carla} exposes it alongside simulated LiDAR, and VirtualHome~\cite{puig2018virtualhome} emits it with segmentation and optical flow --- but here it is captured at the same frame boundary as the spatial-relation graph, the event-to-frame mapping, and the per-actor pose described below, across multi-actor scenes whose structure is driven by an explicit narrative specification. Figure~\ref{fig:depth_gallery} shows RGB and depth pairs across a range of environments.

\subsubsection{Per-Frame Spatial Relation Graphs}
\label{sec:spatial_relations}

At each captured frame, the spatial relations collector queries the 3D position and rotation of every entity (actors and objects) along with the camera state (position, look-at target, field of view, roll). For each entity, it computes camera-relative spatial data: Euclidean distance, horizontal and vertical angles, a coarse direction bucket (front, back, left, right, above, below), and an in-field-of-view flag. It then computes pairwise relations between \emph{all} entity pairs --- not just entity-to-camera --- producing a complete per-frame spatial relation graph. Each entity retains its story-level identifier from the input GEST, maintaining the link between the symbolic specification and the visual observation across the entire video. Content-hash deduplication avoids redundant writes for static scenes where no entity has moved. queries the 3D position and rotation of every entity (actors and objects) along with the camera state (position, look-at target, field of view, roll). For each entity, it computes camera-relative spatial data: Euclidean distance, horizontal and vertical angles, a coarse direction bucket (front, back, left, right, above, below), and an in-field-of-view flag. It then computes pairwise relations between \emph{all} entity pairs --- not just entity-to-camera --- producing a complete per-frame spatial relation graph. Each entity retains its story-level identifier from the input GEST, maintaining the link between the symbolic specification and the visual observation across the entire video. Content-hash deduplication avoids redundant writes for static scenes where no entity has moved.

Figure~\ref{fig:spatial_sample} shows an excerpt from a spatial relations file, illustrating the structure of the per-entity and pairwise data.

\begin{figure}[tb]
\centering
\begin{small}
\begin{verbatim}
{ "camera": {"fov": 70,
    "position": {"x": 2317.13, "y": -1017.36, "z": 1052.31},
    "lookAt": {"x": 2313.20, "y": -1008.20, "z": 1049.21}},
  "entities": [
    { "storyActorId": "a1", "elementType": "ped",
      "currentActionName": "TakeOut", "currentEventId": "a1_3",
      "position": {"x": 2312.92, "y": -1009.71, "z": 1050.21},
      "spatial": {"distance": 8.99, "direction": "front",
        "angleHorizontal": 5.64, "angleVertical": -13.51,
        "inFOV": true}},
    { "storyObjectId": "10_house9", "elementType": "object",
      "objectType": "Sink",
      "position": {"x": 2312.67, "y": -1010.22, "z": 1049.08},
      "spatial": {"distance": 9.02, "direction": "front",
        "angleHorizontal": 8.79, "angleVertical": -20.98,
        "inFOV": true}}],
  "objectRelations": [
    { "fromId": "10_house9", "toId": "16_house9",
      "distance": 3.01, "direction": "left",
      "angleHorizontal": 77.84, "angleVertical": 2.46}]}
\end{verbatim}
\end{small}
\caption{\textbf{Spatial relations excerpt} (abridged from a frame with 32 entities and 496 pairwise relations). Each entity carries its story-level identifier, 3D position, and camera-relative spatial data. Pairwise relations include Euclidean distance, direction, and angular offsets between every entity pair.}
\label{fig:spatial_sample}
\end{figure}

\subsubsection{Skeletal Pose Extraction}
\label{sec:pose}

Unlike the depth channel, the pose track is produced not by the rendering pipeline but by the engine's animation system. At every captured frame, for every active actor, the server queries a per-bone position function exposed by the host engine; the returned values are the authoritative 3D world-space positions of every skeleton joint at that instant --- identical to the pose the renderer is about to use for skinning. A companion world-to-screen projection, again native to the host engine, gives the 2D image-plane coordinate of each joint under the same camera parameters that produce the RGB frame. The pose track is therefore ground truth in the strict sense: no estimator sits in the loop, and the joint positions come from the same animation state that produces the pixels.

\begin{figure}[htbp]
\centering
\includegraphics[width=0.9\textwidth]{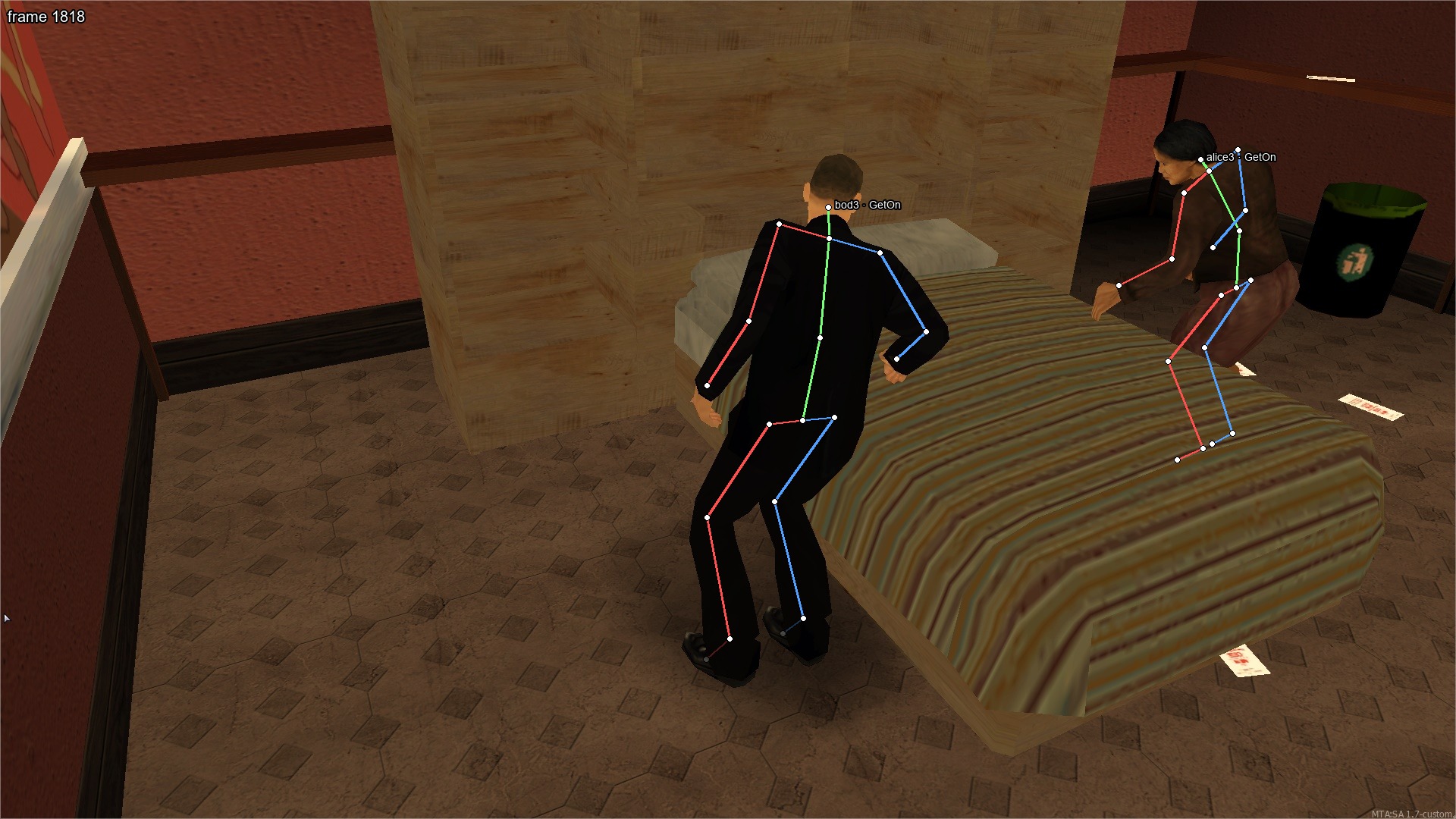}
\caption{\textbf{Pose overlay} for a two-actor scene in which both actors are mounting a bed. Bones are color-coded by side (red for left-body chains, blue for right-body chains, green for the center line); the per-actor label combines the story-level actor identifier from the input GEST with the name of the active action at this frame. The skeletons come from the engine's animation state rather than from image estimation and therefore carry no jitter and exact visibility labels.}
\label{fig:pose}
\end{figure}

The exported skeleton has eighteen joints --- head, neck, spine, and pelvis along the center line, and symmetric shoulder/elbow/hand and hip/knee/ankle/foot chains --- in the same ballpark as the seventeen COCO keypoints~\cite{lin2014microsoft} and short of the twenty-four body joints of the SMPL body model~\cite{SMPL:2015} used in synthetic pose datasets such as SURREAL~\cite{varol2017learning} and BEDLAM~\cite{black2023bedlam}. Each bone in the per-frame record carries its name, a 3D world position, a 2D screen projection, and a visibility flag distinguishing joints inside the output image from those that project outside it or behind the camera. The record is stamped with the same frame index as the RGB and depth tracks, and per-actor identity is preserved across frames through the story-level identifier the input GEST assigns to that actor. Figure~\ref{fig:pose} overlays the resulting skeletons for a two-actor interaction. Because joint positions are read from the animation state rather than inferred from image content, the track has two properties estimator-based data cannot match: there is no temporal jitter, since a joint moves exactly as the animation curve moves it; and occlusion is known exactly, since the visibility flag is derived geometrically from the screen projection rather than from a confidence score. This makes the track usable both as supervision for pose estimators and as reference ground truth for evaluating pose tracking on coordinated multi-actor scenes.

\subsubsection{Screen-Space Annotation Alignment}
\label{sec:screenspace}

\begin{figure}[htbp]
\centering
\includegraphics[width=0.9\textwidth]{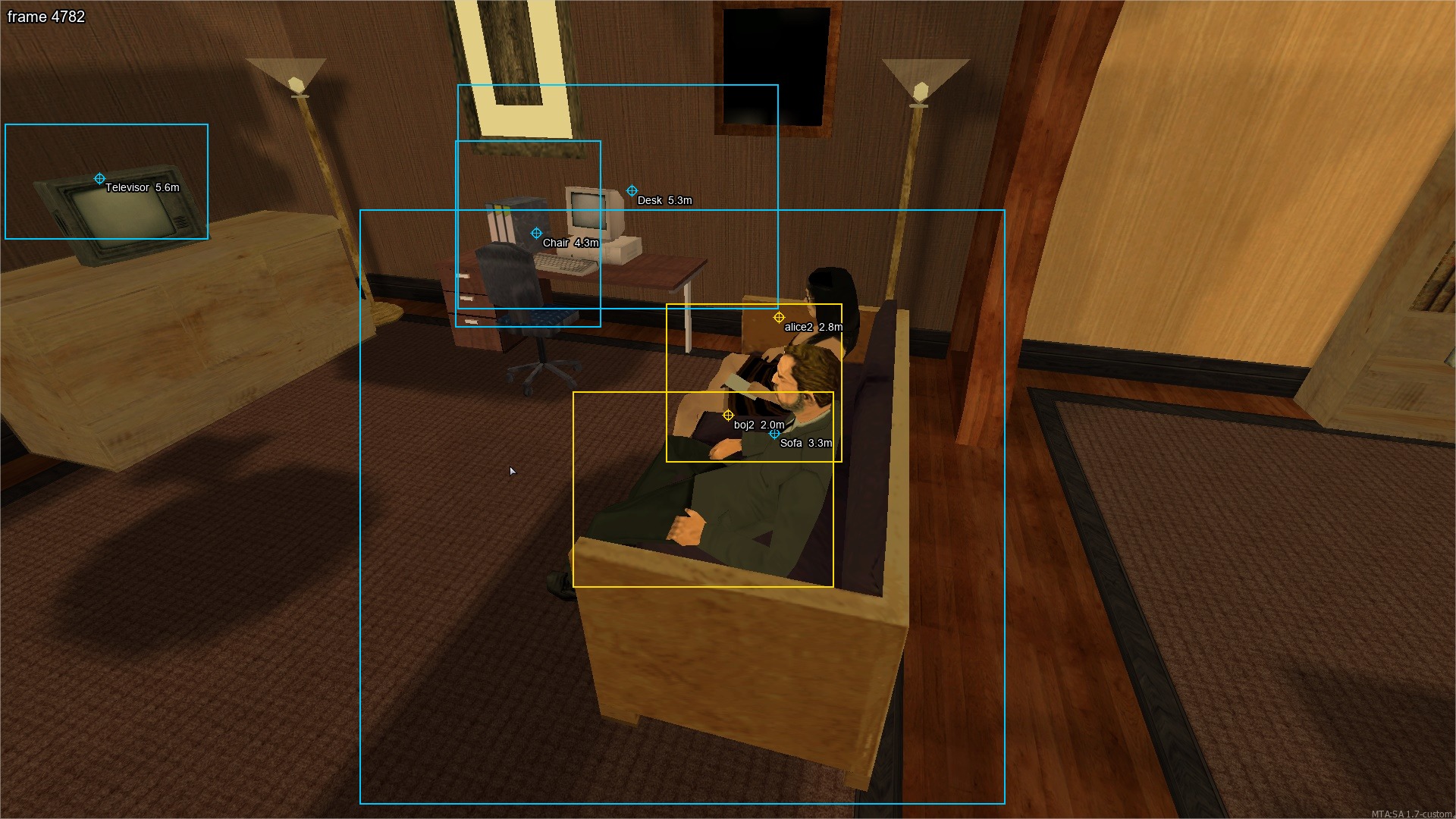}
\caption{\textbf{Screen-space alignment overlay} on a living-room scene. Per-actor bounding boxes (yellow) come from the pose-bone envelope of Section~\ref{sec:pose}; per-object bounding boxes (cyan) come from the engine's collision geometry. Each entity additionally carries a center marker, a camera-relative distance label, and its story-level identifier; entities whose projections fall outside the viewport are rendered in grey.}
\label{fig:screenspace}
\end{figure}

The per-frame spatial relation graph of Section~\ref{sec:spatial_relations} expresses entity positions and pairwise relations in the engine's 3D coordinate frame. That description is natural for a consumer reasoning in 3D but awkward for a 2D vision model that expects pixel-level labels, so the world-space log is complemented with a parallel image-plane description that every tracked entity carries alongside its 3D coordinates. Point-like quantities --- entity centers, bone joints, and any camera-to-entity vector --- are projected through the host engine's world-to-screen function, which returns both the pixel coordinate and an in-viewport flag. Extent-like quantities --- 2D bounding boxes around actors and objects --- are obtained by projecting the eight corners of a 3D bounding box and taking the 2D envelope of the projections. The source of the 3D box differs by entity type: for actors it is the head-to-foot envelope of the pose bones (Section~\ref{sec:pose}), which follows the silhouette of the currently animated body; for rigid objects (chairs, desks, phones, beds) it is the axis-aligned box exposed by the engine's collision-geometry API. Together with the pose joints and the depth map, this exposes the full 2D annotation vocabulary --- bounding boxes, keypoints, and per-pixel depth --- that standard dataset formats such as COCO~\cite{lin2014microsoft} expect. Figure~\ref{fig:screenspace} shows the resulting overlay on a living-room scene. The object-level boxes inherit the approximation of the collision mesh, a conservative axis-aligned proxy that carries a small margin around the visible silhouette: this is sufficient for the entity-level, position-level, and coarse-detection tasks the engine targets, while pixel-precise outlines would require a post-processing refinement against the rendered silhouette.

\subsubsection{Event-Frame Temporal Alignment}
\label{sec:temporal_alignment}

An EventBus-driven collector subscribes to \texttt{graph\_event\_start} and \texttt{graph\_event\_end} signals emitted by the orchestration engine, recording the exact frame at which each GEST event begins and ends:
\[
\{e_i \mapsto [\text{startFrame}_i, \text{endFrame}_i]\}
\]
This produces a complete temporal bridge between the input GEST and the output video: every frame can be mapped back to which graph events are active, and every event can be located to its precise frame range. The full temporal constraint graph is preserved alongside the mappings, enabling downstream evaluation that tests genuine temporal reasoning rather than single-frame recognition. The mapping is written to disk incrementally to prevent data loss in the event of a crash.

\subsubsection{Logger System}
\label{sec:logger}

The Logger generates a procedural natural language description of the simulation as it unfolds. Unlike the proto-language used in the text generation pipeline (Section~\ref{sec:text_gen}), which is derived from the GEST and refined by an LLM, the Logger's output is assembled entirely at runtime from the descriptions attached to each component in the system: every action class carries a description template (e.g., ``takes out,'' ``sits down on''), every object instance carries a description (e.g., ``a red apple,'' ``a glass of beer''), and every actor skin carries a detailed appearance description (e.g., ``a middle-aged white man with blonde hair in a black police uniform'').

When an actor enters a region for the first time, the Logger describes the room relative to the camera viewpoint, listing the objects and entities present and their spatial arrangement (left, right, in front, straight ahead). For example: ``Inside a kitchen there is 1 individual: a middle-aged white man [...] named Actor\_a1. Inside the kitchen in the right side there are 5 chairs, 8 red apples, a glass of beer, a table [...]. In front there is a gas cooker.'' Object counts are computed dynamically --- duplicate objects of the same type are grouped (e.g., ``5 chairs'' rather than listing each individually).

For each action, the Logger assembles a sentence from the actor's name (or pronoun if previously introduced), the action's description template, and the target object's description. Phrase links (``Then,'' ``Afterwards,'' ``When [actor] finishes'') connect sentences into a flowing narrative. Long-duration actions use present continuous tense with a completion clause. The result is a complete textual description of the video, generated at zero cost alongside the other artifacts. An example of Logger output for a two-actor story in a house is shown below:

\begin{quote}
\small
\textit{``Inside a kitchen there is 1 individual: a middle-aged white man with blonde hair in a black police uniform named Actor\_a1. Inside the kitchen in the right side there are 5 chairs, 8 red apples, a glass of beer, a table, a bottle of beer and 4 plates. In front there is a gas cooker. [...] Actor\_a1 takes out a phone from his pocket. Actor\_a0 sits down on the chair. Afterwards she stands up from the chair. Actor\_a1 answers the phone and talks. Actor\_a0 picks up a red apple from kitchen counter. [...] When Actor\_a1 finishes he hangs up. Then he stashes the phone in his pocket and is waiting for something.''}
\end{quote}

\section{Simulator Capability Extraction}
\label{sec:capability_extraction}

For any system that generates GESTs --- whether a procedural random generator (Section~\ref{sec:procedural_gen}) or an LLM-based agentic system (Section~\ref{sec:agentic}) --- the generator must know what the simulation engine can actually do: which episodes exist, which actions are available at which POIs, which objects can be manipulated, which actor skins are available, and how actions chain together into valid sequences. Hardcoding this knowledge would be fragile: every time a new episode is defined in the World Editor, or a new action is implemented, or a new supertemplate is added, the generator's knowledge would need to be manually updated. Instead, the engine extracts its own capabilities dynamically at runtime.

The capability extractor scans the engine's live data structures --- the same episode definitions, POI templates, action classes, and object catalogs that the pipeline uses during simulation --- and serializes them into a single JSON file. The extraction process loads each episode from disk, initializes it to populate regions and resolve supertemplates (using the same partial initialization described in Section~\ref{sec:set_cover}), and then iterates through the full hierarchy: episodes contain regions, regions contain POIs and objects, and POIs contain action chains with next-action links. For each action encountered, the extractor records its name, the entity roles it requires (actor, second actor for interactions, target object), whether it requires an object, and which object types are compatible. Actions are automatically categorized by inferring the category from the action name (e.g., \texttt{SitDown} and \texttt{StandUp} are classified as positional, \texttt{Eat} and \texttt{Drink} as consumption, \texttt{Handshake} and \texttt{Hug} as social).

Table~\ref{tab:capability_registry} summarizes the contents of the capability registry.

\begin{table}[htbp]
\centering
\caption{Contents of the simulator capability registry.}
\label{tab:capability_registry}
\small
\begin{tabular}{@{}lp{9cm}@{}}
\toprule
\textbf{Category} & \textbf{Contents} \\
\midrule
Episodes & Full episode data: regions, POIs with action lists, objects, episode links for cross-episode navigation \\
Episode catalog & Lightweight index: episode names, linked episodes, region names \\
Action catalog & All actions with entity roles, required object types, category, and description \\
Object types & Object type catalog with spawnable/fixed classification, compatible actions, and instance counts per episode \\
Action chains & Valid action sequences (e.g., PickUp $\to$ Use $\to$ PutDown), ordering constraints, and interaction synchronization rules \\
Spawnable objects & Object types that can be created on demand \\
Interactions & Multi-actor action types requiring synchronized execution \\
Spatial relations & Supported pairwise relations: near, behind, left, right, on, in\_front \\
Temporal relations & Supported constraint types: after, before, same\_time, concurrent, next \\
Player skins & 312 character skins organized by gender, each with a visual description \\
Camera actions & Cinematic camera system documentation: shot types, framing, recording control \\
\bottomrule
\end{tabular}
\end{table}

The action chain section of the registry deserves particular mention. It documents the valid action sequences that the engine supports --- for instance, that a sitting sequence must follow the pattern \texttt{SitDown} $\to$ (seated actions) $\to$ \texttt{StandUp}, that pick-up-able objects require \texttt{PickUp} before use and \texttt{PutDown} after, and that interactions between two actors must be synchronized with a \texttt{same\_time} temporal constraint. These rules were originally written as natural language instructions for an early LLM-based generation workflow; in the current system, the same constraints are enforced programmatically by the procedural generator, but the natural language documentation remains in the registry as a reference for LLM-based approaches.

The capability registry is the foundation on which all GEST generation approaches build. The procedural generator (Section~\ref{sec:procedural_gen}) reads it to determine which actions are available at which POIs, ensuring that every generated graph is executable by construction. The agentic text-to-GEST system (Section~\ref{sec:agentic}) provides it as context to the LLM, enabling the model to generate GESTs that respect the engine's actual capabilities rather than hallucinating actions or objects that do not exist. Because the registry is generated from the same data structures that the engine uses at runtime, it is guaranteed to be consistent with the engine's actual behavior --- adding a new episode or action automatically makes it available to all downstream generators without any manual synchronization.

\section{Procedural GEST Generation}
\label{sec:procedural_gen}

The procedural generator creates random but valid GESTs from the capability registry described in Section~\ref{sec:capability_extraction}. Its purpose is twofold. First, it is the tool used to generate the GTASA corpus used in a companion evaluation: because it only produces specifications that respect the engine's capabilities, every generated graph is executable by construction, enabling unattended batch production of hundreds of videos without manual intervention. Second, it serves as the state-management backend for the agentic text-to-GEST system (Section~\ref{sec:agentic}), where an LLM directs the generation process through function calls while the generator maintains the evolving graph state and enforces validity constraints programmatically.

\subsection{Episode Selection}
\label{sec:procedural_episode_selection}

The generator begins by selecting which simulation environments the story will take place in. Episode selection follows a two-stage process designed to ensure balanced coverage across environment types. In the first stage, an episode \emph{type} is chosen with equal probability from the available categories (classroom, gym, garden, house). This prevents bias from unequal episode counts --- for instance, the house category contains more individual episodes than the classroom category, but each type is equally likely to be selected. In the second stage, a specific episode within the chosen type is selected uniformly at random. If the selected episode has linked episodes (e.g., a house interior linked to a garden through a door), the linked episodes are included in the group, enabling multi-context stories.

For house-type episodes, the generator optionally adds a gym episode to the group with 50\% probability, enabling stories that span both a residential interior and a gym environment. The episode group defines the full set of regions available for the story.

\subsection{Actor and Region Setup}
\label{sec:procedural_actors}

From the selected episode group, the generator identifies all regions that contain actionable POIs (regions with at least one POI that has defined actions). It then selects a subset of these regions to form the story's spatial progression --- the sequence of locations that actors will visit over the course of the narrative. The number of regions is configurable, defaulting to a random selection of one to four regions.

For the first region, the generator creates a random number of actors, each with a randomly assigned gender. The number of actors in a region is bounded by the region's \emph{capacity} --- the number of POIs with defined actions, reflecting how many actors the region can meaningfully support simultaneously. In the GTASA corpus, the maximum actors per region was set to five, which given the multi-region stories resulted in stories with up to six actors. Each actor receives an \texttt{Exists} node in the GEST and is initialized with spawnable object references (mobile phone and cigarette) in their inventory.

\subsection{Action Chain Generation}
\label{sec:procedural_chains}

The core of the generation algorithm is the round-based action chain generation. For each region in the spatial sequence, the generator iterates through a configurable number of rounds (default three). In each round, every actor present in the region receives one action chain.

An action chain is generated by selecting an unused POI in the region and following its next-action sequence as defined in the capability registry. The generator picks the first available action at the POI, creates a corresponding event in the GEST, and then follows the \texttt{possible\_next\_actions} links to generate subsequent events until the sequence terminates (i.e., an action with no next-actions is reached). For example, at a desk POI with the chain \texttt{SitDown} $\to$ \texttt{OpenLaptop} $\to$ \texttt{TypeOnKeyboard} $\to$ \texttt{CloseLaptop} $\to$ \texttt{StandUp}, the generator creates five events in sequence, each linked temporally to its predecessor within the same actor's chain.

If no unused POI is available for an actor in a given round, the generator falls back to a spawnable chain: the actor takes out a phone or cigarette and performs the associated actions (e.g., \texttt{TakeOut} $\to$ \texttt{TalkPhone} $\to$ \texttt{HangUp} $\to$ \texttt{Stash}). This ensures that every actor has something to do in every round, even when all POIs in the region are occupied.

All events are initially created in temporary buffers. Only when an entire chain is successfully generated --- all actions have valid POI bindings and no conflicts exist --- are the events committed to the global GEST state. If chain generation fails at any point, the temporary events are silently discarded and the generator moves on to the next actor or the fallback chain. This transactional pattern is what guarantees executability by construction: no partially valid chains can enter the output.

\subsection{Interactions and Object Exchange}
\label{sec:procedural_interactions}

When multiple actors share a region, the generator creates coordinated interactions between them. Interaction types include handshakes, hugs, kisses, conversations, and object exchanges (\texttt{Give}/\texttt{Receive}). Both actors must be co-located for the interaction to occur, and the events are linked with a \texttt{same\_time} temporal constraint to ensure synchronized execution.

For object exchanges, the generator creates a \texttt{Give} event for one actor and a corresponding \texttt{INV-Give} (inverse give, equivalent to \texttt{Receive}) event for the other, both referencing the same object. The receiver then generates a follow-up chain using the received object (e.g., eating food that was given to them), enabling multi-actor narrative threads where objects flow between characters.

\subsection{Temporal Relation Injection}
\label{sec:procedural_temporal}

Temporal relations between actors are injected at several levels. Within each round, actors' chains are implicitly concurrent --- they execute in overlapping time windows without explicit ordering. Between rounds, \texttt{before} constraints ensure that all actors complete round $n$ before any actor begins round $n+1$ in the same region.

When actors move between regions, the temporal ordering becomes more involved. All non-moving actors must complete their current actions before any moving actor begins their \texttt{Move} event --- preventing a situation where a moving actor departs while an interaction partner is still mid-action. Cross-mover constraints ensure that each mover's pre-move action completes before other movers begin moving, establishing a clean sequential departure. These constraints are the result of iterative debugging of race conditions encountered during large-scale corpus generation, and they ensure that the temporal orchestrator (Stage~3) can always find a valid execution order.

\subsection{Region Transitions}
\label{sec:procedural_regions}

After all rounds in a region are complete, the generator selects a random subset of actors to move to the next region in the spatial sequence. At least one actor moves, and up to all actors may move. Each moving actor receives a \texttt{Move} event with the target region as the destination. Actors who do not move remain in the current region and are available for future interactions with newly arriving actors.

When actors arrive in a new region that has not been visited before, the generator optionally creates additional actors (zero to two) in that region, simulating the presence of characters who were already there. These new actors receive their own \texttt{Exists} nodes and are initialized with spawnable objects, and they participate in subsequent rounds of chain generation alongside the arriving actors.

\subsection{Complexity Parameters}
\label{sec:procedural_params}

Table~\ref{tab:procedural_params} summarizes the configurable parameters that control the complexity of the generated GESTs. These parameters were tuned to produce the range of story complexities found in the GTASA corpus, which contains stories with 2--6 actors, 7--65 events, and 1--4 regions.

\begin{table}[htbp]
\centering
\caption{Configurable parameters of the procedural GEST generator.}
\label{tab:procedural_params}
\small
\begin{tabular}{@{}llp{7cm}@{}}
\toprule
\textbf{Parameter} & \textbf{Default} & \textbf{Description} \\
\midrule
chains-per-actor & 3 & Number of action chain rounds per actor per region \\
max-actors-per-region & 5 & Maximum actors in a single region \\
max-regions & -- & Maximum regions to visit (default: random 1--4) \\
episode-type & random & Force a specific environment type (classroom, gym, garden, house) \\
seed & -- & Random seed for reproducible generation \\
\bottomrule
\end{tabular}
\end{table}

\subsection{Dual Role as State Backend}
\label{sec:procedural_backend}

Beyond standalone generation, the procedural generator serves as the state-management backend for the agentic text-to-GEST system of Section~\ref{sec:agentic}. The generator's internal state --- actors, events, temporal constraints, spatial relations, object tracking, and POI capacity --- is exposed through a delegation pattern that allows an external agent (in this case, an LLM with function-calling capabilities) to invoke the same methods the random generator uses internally. The agent calls functions like ``create actor,'' ``add action event,'' or ``add temporal constraint,'' and the generator maintains the evolving graph state, enforces validity constraints, and tracks which POIs and objects are available. This ensures that LLM-generated GESTs benefit from the same executability guarantees as procedurally generated ones, because the state backend programmatically prevents invalid operations regardless of whether the caller is a random generator or an intelligent agent.

\section{Agentic Text-to-GEST Generation}
\label{sec:agentic}

Procedural generation (Section~\ref{sec:procedural_gen}) produces event-dense, structurally valid stories, but has no authored narrative intent: actors execute random sequences of available actions, and story-level structure --- if present --- is latent in the action sequence rather than explicitly authored. The agentic system described here closes this gap by introducing an LLM \emph{Director} that plans stories with explicit intent, motivation, and causal structure, while preserving the executability-by-construction guarantee of the underlying state backend. It also inverts the dominant paradigm for multi-agent video generation, in which LLM agents~\cite{hu2024storyagent,wang2026mavis,wang2026dreamrunner} ultimately drive a neural video generator and therefore inherit its lack of semantic guarantees --- objects appearing and disappearing, actors morphing between frames, temporal ordering violated~\cite{brooks2024sora,wan2025wan}. Here the agents instead construct a formal GEST specification that is executed deterministically by the engine described in the preceding sections, yielding multi-actor narrative video with dense annotations as a by-product. This turns video generation from a programming task, where story structure is fixed in the procedural generator's code, into a specification task, where natural language shapes the story without any code changes.

\subsection{The Staged Approach and Its Failure}
\label{sec:staged}

A GEST must be both \emph{narratively coherent} (telling a meaningful story) and \emph{simulator-valid} (respecting the engine's action chains, object lifecycles, POI capacities, and temporal-constraint rules). LLMs excel at the former but cannot maintain the precise state tracking the latter requires across the 50--200 events of a multi-actor story. Our first design was a staged refinement pipeline chaining six specialized stages --- \emph{Concept} (an abstract hierarchical GEST with parent and leaf scenes and actor archetypes), \emph{Casting} (skin assignment from the engine's catalog), \emph{Episode Placement} (mapping scenes to valid episodes), \emph{Setup} (off-camera preparation and backstage positioning), \emph{Screenplay} (translating abstract narrative into concrete action sequences), and \emph{Scene Detail} (expanding each leaf scene into a complete executable GEST) --- each a single prompt-and-parse step with structured-output validation. The pipeline produced narratively coherent graphs, but across 50 generation attempts \emph{zero} produced an executable specification. The failure modes compounded across stages: hallucinated actions (e.g., \texttt{WalkTo} instead of \texttt{Move}), invalid action chains (\texttt{TypeOnKeyboard} without a preceding \texttt{SitDown} and \texttt{OpenLaptop}), object-lifecycle violations (putting down objects never picked up, two actors holding the same object), temporal cycles the Floyd--Warshall solver cannot resolve, and context overflow from the large capability registry. The fundamental problem is that each stage makes locally reasonable decisions while the accumulated state --- which actors are where, what they hold, which POIs are occupied, which constraints are in effect --- drifts from validity as the graph grows, and post-hoc correction requires backtracking through dependent state at superlinear cost. The lesson motivated the architecture that follows: \emph{the LLM should handle narrative decisions (what should happen), while a programmatic backend enforces state and constraints (what is allowed to happen)}.

\subsection{Architecture Overview}
\label{sec:agentic_arch}

\begin{figure}[htbp]
\centering
\includegraphics[width=0.9\textwidth]{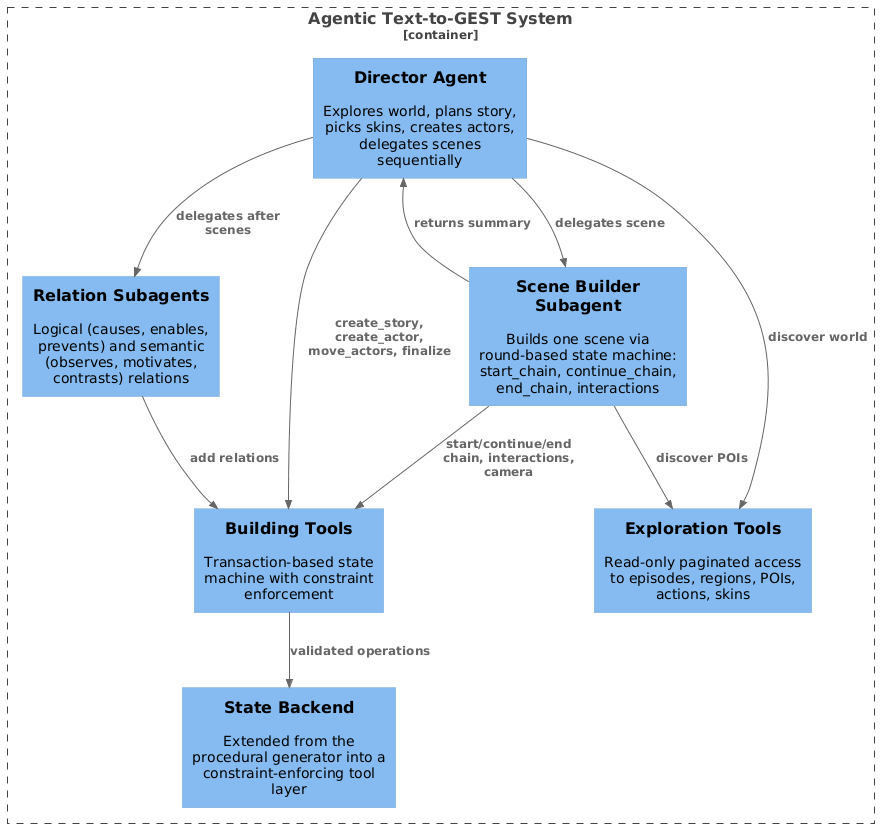}
\caption{\textbf{Agentic system architecture.} The Director Agent explores the simulation world and plans the story, delegating individual scenes to the Scene Builder Subagent. Both agents interact through Exploration Tools (read-only) and Building Tools (constraint-enforcing), which operate on the State Backend --- the procedural generator of Section~\ref{sec:procedural_gen} repurposed as a tool layer.}
\label{fig:agentic_architecture}
\end{figure}

The system uses a hierarchical two-agent architecture. A \emph{Director Agent} handles high-level story planning --- exploring the simulation world, selecting episodes and regions, picking actor skins, creating actors, and determining the sequence of scenes. For each scene it delegates detailed event construction to a \emph{Scene Builder Subagent}, which builds action chains, interactions, and temporal constraints through tool calls. After all scenes, the Director finalizes the GEST by linking scene boundaries with temporal constraints. Both agents operate exclusively through function calling: every action --- creating an actor, starting a chain, adding a constraint --- is a tool call validated by the state backend (Section~\ref{sec:agentic_constraints}), and the agents never manipulate the GEST directly. This operationalizes the separation of concerns above: because the same procedural generator that guarantees executability for random generation (Section~\ref{sec:procedural_gen}) validates and applies every tool call, every GEST the system produces is executable by construction, regardless of what the LLM attempts. Figure~\ref{fig:agentic_architecture} shows the architecture.

\subsection{The Director Agent}
\label{sec:director}

The Director receives an optional seed prompt and a generation configuration (target scenes, actors), and proceeds in four phases. \textbf{Exploration}: using read-only, paginated tools it discovers the world --- browsing episodes, examining regions and their objects, inspecting POI action chains, and browsing skins by gender (each with a visual description, e.g., ``a middle-aged white man with blonde hair in a black police uniform''). Pagination is essential, as the capability registry is too large for a single prompt, so the agent must actively navigate the space. \textbf{Casting}: it creates the story structure via \texttt{create\_story} and \texttt{create\_actor} calls, specifying each actor's name, gender, skin, and starting region, before any scene is delegated. \textbf{Scene building}: scenes are processed sequentially --- for each, the Director calls \texttt{start\_scene} with the episode, region, and participating actors, delegates to the Scene Builder with a natural-language task (e.g., ``build a scene where James and Sarah have lunch in the kitchen''), and afterward handles follow-ups and \texttt{move\_actors} transitions. \textbf{Finalization}: it calls \texttt{finalize\_gest} to link scene boundaries with cross-scene temporal constraints, producing the complete GEST.

\subsection{The Scene Builder Subagent}
\label{sec:scene_builder}

The Scene Builder receives an isolated context --- scene identifier, episode, region, actors, and a narrative description --- and does not see the full story, reducing context requirements and preventing accidental modification of other scenes. Events are constructed through a round-based state machine: \texttt{start\_round} begins a temporal unit and returns the current state of all actors (posture, held objects, location); for each actor, \texttt{start\_chain(actor, POI)} returns the valid next actions at the selected POI, one or more \texttt{continue\_chain(actor, action)} calls each advance the chain by one action and return the next valid actions, and \texttt{end\_chain(actor)} commits it; \texttt{do\_interaction(a1, a2, type)} handles synchronized two-actor actions; \texttt{start\_recording} and \texttt{stop\_recording} control the camera on committed events; \texttt{add\_temporal\_dependency} and \texttt{add\_starts\_with} express cross-actor ordering; and \texttt{end\_round} closes the unit. Multiple actors can hold active chains simultaneously and the subagent can interleave calls across them. The Scene Builder cannot create actors, move actors between regions, or finalize the GEST --- those responsibilities belong exclusively to the Director --- enforcing a clean split between story-level and scene-level concerns.

\subsection{Relation Subagents}
\label{sec:relation_agents}

After each scene is built and after the final GEST is assembled, the Director optionally delegates to two subagents that enrich the graph with higher-level annotations that procedural generation never produces. A \emph{Logical Relations Agent} adds causal and dependency edges between events (causes, enables, prevents, requires), capturing why events occur in relation to each other. A \emph{Semantic Relations Agent} adds narrative-coherence edges with free-text types (observes, interrupts, motivates, sets\_context\_for, contrasts\_with), capturing how events relate at the story level rather than the temporal level. Both receive the list of committed event identifiers and operate through the same validated tool interface, so only valid event references are used. These relations are optional --- the GEST is executable without them --- but they populate the logical and semantic edge types defined in the GEST formalism, exercising the full expressive capacity of the representation in a single end-to-end pipeline.

\subsection{Tool-Based Constraint Enforcement}
\label{sec:agentic_constraints}

The building tools implement a transaction-based state machine with four states (\texttt{IDLE}, \texttt{STORY\_CREATED}, \texttt{IN\_SCENE}, \texttt{IN\_ROUND}); each tool call verifies the correct state before executing. Actor state tracking maintains posture (standing, sitting, lying), held objects (with type and origin region), and current location --- actors must be standing to start chains or interactions, and held objects can only be put down in their origin region. Chains are transactional: \texttt{continue\_chain} creates events in temporary buffers and only \texttt{end\_chain} commits them, so failed chains are silently discarded. Exclusive-use objects (chairs, beds, gym equipment) are tracked per region to prevent two actors from claiming the same object. Before adding a \texttt{before} relation the system walks the full dependency graph and rejects any edge that would create a cycle, returning an explanatory error. Interaction rules require both actors to be standing, co-located, and not holding spawnables, and forbid consecutive interactions without an intervening chain; the \texttt{Give} action requires an explicit receiver identifier and automatically creates the synchronized \texttt{INV-Give} event. Spawnable phone and cigarette actions are atomic \texttt{TakeOut}$\to$\texttt{Use}$\to$\texttt{Stash} sequences that lock the actor until completion. Complementing these building tools, read-only \emph{exploration tools} provide paginated access to the same capability registry (Section~\ref{sec:capability_extraction}) --- browsing episodes, regions, and POIs, discovering valid first and next actions at a POI, listing skins, spawnable types, and interaction types, and retrieving the constraint documentation. This explore-before-plan pattern is what prevents hallucination: the agent must verify through a tool call that a POI affords an action before including it in a plan, directly addressing the primary failure mode of the staged approach.

The agentic system operates either from a text prompt (seeded mode) or fully autonomously, where the Director explores the world, conceives a narrative, and emits both the executable GEST and a natural-language story description. A companion evaluation compares autonomous agentic narratives against procedural baselines and against neural video generators; the generation architecture is the focus here.

\section{Text Generation Pipeline}
\label{sec:text_gen}

Each video produced by the GEST-Engine is accompanied by a natural language description of the depicted events. The text generation approach builds on the proto-language pipeline originally developed by Masala et al.\ \cite{masala2025vision} for converting GESTs extracted from real videos into textual descriptions. That pipeline operates in two stages: a deterministic conversion from the GEST to a proto-language (a concise, machine-generated textual representation), followed by LLM-based refinement into fluent natural language.

We adapted this pipeline for our setting, where a fundamental difference exists: we have access to the complete ground-truth GEST produced by the engine, rather than a noisy extraction from video. This access enabled three enhancements. First, we incorporate \emph{location context}: the episode setting is prepended to the proto-language (e.g., ``In the garden the first man\ldots''), move actions include their source and destination regions (e.g., ``moves from garden to driveway''), and when an actor enters a new region after a move, the proto-language inserts the new location (e.g., ``in the driveway the first man takeouts a mobilephone''). Second, we incorporate \emph{actor gender}: because the GEST specifies each actor's gender via the \texttt{Exists} node properties, the proto-language uses gender-appropriate references (``the first woman,'' ``the man''), and the LLM refinement preserves these distinctions with correct pronouns. Third, we incorporate the \emph{event-frame temporal alignment} captured during simulation (Section~\ref{sec:temporal_alignment}), embedding exact frame ranges into each event so that downstream processing can reference the precise temporal extent of each action.

\subsection{Proto-Graph and Proto-Language}
\label{sec:proto_language}

Before the proto-language can be generated, the output GEST must be transformed into a \emph{proto-graph} --- an intermediate format compatible with the text generation system. This transformation applies three changes. Entity identifiers are normalized to a canonical format: actor IDs such as \texttt{a0}, \texttt{a1} become \texttt{actor0}, \texttt{actor1}; spawnable object IDs become sequential identifiers with type annotations (e.g., \texttt{id:0.0-class:mobilephone}). The \texttt{Timeframe} field of each event, which is \texttt{null} in the input GEST, is populated with the exact start and end frame numbers from the event-frame mapping captured during simulation. And the temporal section is preserved with the transformed identifiers. Appendix~\ref{app:text_gen_example} provides a concrete example of this transformation.

The proto-graph is then converted to prose by iterating through each actor's event chain and assembling atomic sentences from the action's verb form, the involved entities, and the target objects. Temporal connectors link the sentences: ``After that'' for sequential actions, ``at the same time also'' for \texttt{same\_time} constraints, and ``meanwhile'' for concurrent actions across actors. The result is a robotic but complete description that preserves the full information content of the GEST. For example, a four-actor sequence in a kitchen might produce:

\begin{quote}
\small\ttfamily
In the kitchen the first woman sitdowns a chair. Meanwhile, the man sitdowns a chair. Meanwhile, the second woman takeouts a cigarette. After that, the third woman washhandss a sink. After that, the first woman pickups a food. [\ldots] After that, the first woman moves from kitchen to bedroom at the same time also the man moves from kitchen to bedroom.
\end{quote}

This proto-language intentionally uses ungrammatical verb forms (\texttt{sitdowns}, \texttt{washhandss}, \texttt{pickups}) because it is a mechanical concatenation of action names with entity descriptions --- the grammatical refinement is delegated to the next stage.

\subsection{LLM Refinement}
\label{sec:llm_refinement}

The proto-language is wrapped in a prompt that instructs the LLM to rewrite the paragraph into natural, concise, and accurate text while preserving all actions and their temporal relations. The prompt includes the location context, a one-shot example showing the expected transformation from proto-language to refined text, and instructions for resolving ambiguous objects by selecting the most plausible option from the context.

We conducted a qualitative comparison across several OpenAI models, including GPT-4o, GPT-4.1, GPT-5.2, and their mini variants. GPT-4o (specifically the \texttt{gpt-4o-2024-08-06} checkpoint) was selected because it produced the most descriptive and faithful refinements. The GPT-5 variants, while more capable in general, tended to over-summarize --- condensing multiple actions into comma-separated lists rather than preserving the sequential narrative structure that matches the video content. The mini variants lacked sufficient detail in their descriptions.

The refined output for the kitchen example above becomes:

\begin{quote}
\small
\textit{``In the kitchen, the first woman sits down on a chair, as does the man. Meanwhile, the second woman takes out a cigarette, and the third woman washes her hands at the sink. The first woman and the man both pick up food. [\ldots] Then, the first woman and the man move from the kitchen to the bedroom.''}
\end{quote}

\noindent A complete example showing the full chain from GEST to proto-graph to proto-language to LLM-refined output is provided in Appendix~\ref{app:text_gen_example}.

\subsection{Integration in the Production Pipeline}
\label{sec:text_integration}

The text generation pipeline is integrated into the batch production system as a post-simulation step. After a simulation completes successfully and all artifacts have been collected, the batch controller first exports the proto-graph (combining the output GEST with the event-frame mapping from the capture system), then invokes the text generation pipeline. The pipeline can operate in two modes: \emph{prompt-only} mode generates and saves the GPT prompt without calling the API, enabling deferred or batched LLM processing; \emph{full} mode generates the prompt and immediately calls the OpenAI API to produce the refined description.

The resulting textual description directory for each simulation contains three artifacts: the engine-generated text produced by the Logger during simulation (Section~\ref{sec:logger}), the GPT prompt (the proto-language with instructions), and the LLM-refined natural language description. The refined descriptions serve a dual purpose in a companion evaluation: they are used as training data for video captioning models, and as text prompts for neural video generators (VEO~3.1, WAN~2.2) to produce comparison videos for the human evaluation study.

\section{Production Orchestration}
\label{sec:production}

\begin{figure}[htbp]
\centering
\includegraphics[width=\textwidth]{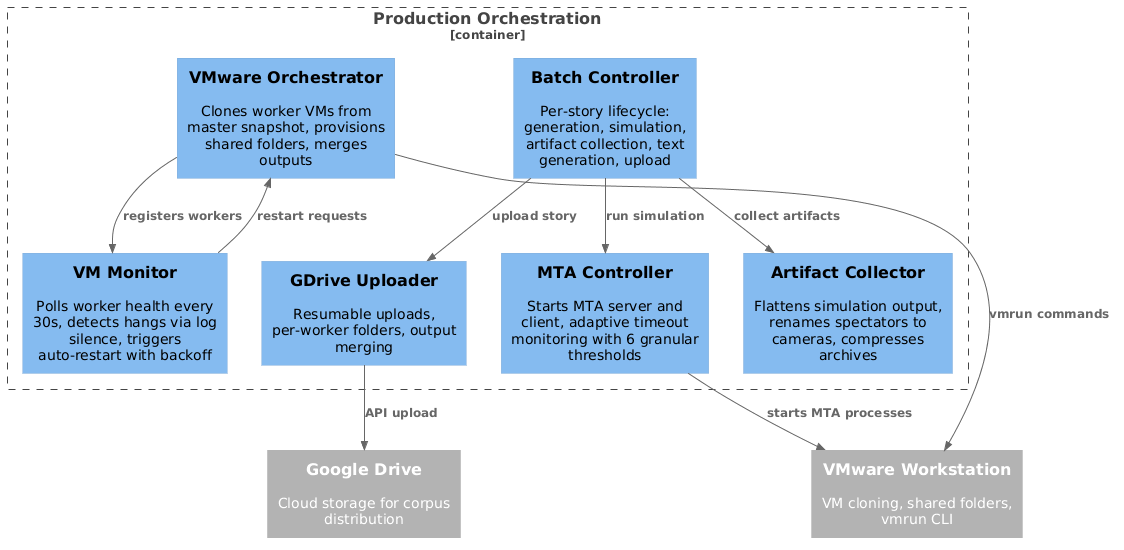}
\caption{\textbf{Production orchestration components.} The VMware Orchestrator provisions worker VMs from a master snapshot and monitors their health. Each VM runs a Batch Controller that manages the per-story lifecycle through the MTA Controller, Artifact Collector, and GDrive Uploader. External systems include VMware Workstation for VM management and Google Drive for corpus distribution.}
\label{fig:production_components}
\end{figure}

Generating a corpus of hundreds of videos requires unattended batch execution across many hours, with robust handling of the inevitable crashes, freezes, and stalls that occur when running a 20-year-old game engine at scale. The production orchestration system addresses this through a layered architecture of components that manage the full lifecycle of each story --- from GEST generation through simulation, artifact collection, text generation, and cloud upload --- with comprehensive monitoring and automatic recovery at every stage. Figure~\ref{fig:production_components} shows the components involved, and Figure~\ref{fig:story_lifecycle} (Appendix~\ref{app:story_lifecycle}) illustrates the per-story lifecycle in detail.

\subsection{VM Infrastructure}
\label{sec:vm_infrastructure}

The production system runs on VMware Workstation Pro, which enables multiple independent MTA instances to execute in parallel on a single physical machine. A \emph{master VM image} was prepared with the complete simulation environment pre-installed: GTA San Andreas, Multi Theft Auto, the GEST-Engine source code, Python 3.12, and all required dependencies. The Windows Task Scheduler was configured to automatically launch the simulation auto-runner on boot, so that each VM begins processing immediately when started. A command-line deployment tool allows the latest source code changes to be pushed to all VMs without recreating the master image.

From this master image, a \emph{snapshot} is taken that captures the complete system state. The VMware Orchestrator uses this snapshot to create \emph{linked clones} --- lightweight VM instances that share the master image's disk and only store their own changes. Linked cloning is fast (seconds rather than minutes) and storage-efficient, enabling the creation of many concurrent workers without duplicating the full disk image for each. Each worker VM receives its job configuration through VMware's shared folder mechanism: the host writes a job file specifying how many stories to generate, which generator to use, and where to upload results, and the VM's auto-runner reads this file on boot.

The auto-runner inside each VM performs several setup steps before starting the batch: it waits for the shared folders to mount, sets the display resolution to the target capture dimensions, and applies firewall rules that block MTA's external network access --- preventing the client from attempting to contact the MTA master server, which would cause the client to freeze during startup. Once configured, the auto-runner invokes the batch controller with the parameters from the job file and begins streaming progress back to the host through the shared folder.

\subsection{The MTA Controller}
\label{sec:mta_controller}

The MTA Controller manages the lifecycle of a single simulation. It starts the MTA server process, waits for it to initialize, then starts the MTA client process and waits for it to connect. Because the Desktop Duplication API (Section~\ref{sec:rgb_capture}) requires the game window to be visible and unobstructed, the controller sets the MTA window to always-on-top after the client starts.

The controller then enters an adaptive timeout monitoring loop that polls every two seconds. The monitoring checks multiple health signals simultaneously:

\begin{itemize}
    \item \textbf{Process health}: whether the server and client processes are still running, and whether a crash dialog has appeared.
    \item \textbf{Window responsiveness}: whether the MTA window responds to messages, detecting freezes where the process is alive but unresponsive.
    \item \textbf{Log-based progress}: the controller tails the server and client logs incrementally, looking for specific patterns that indicate simulation progress (action completions, event dispatches) or completion.
\end{itemize}

The monitoring uses a hierarchy of six timeout thresholds, each targeting a different failure mode:

\begin{enumerate}
    \item \textbf{Server ready} (180\,s): the MTA server must report that it is accepting connections.
    \item \textbf{Client connect} (180\,s): the MTA client must connect to the server. Failure here typically indicates a UI dialog blocking the client.
    \item \textbf{First action} (600\,s): the simulation must begin executing actions. A long delay suggests the graph parsing or grounding stages have encountered an unusual input.
    \item \textbf{Action progress} (600\,s, adaptive): after the first action, subsequent actions must appear in the log within this window. This timeout resets on every detected action, making it adaptive to stories of varying length.
    \item \textbf{Log silence} (60\,s): if neither log produces any output for this duration, the simulation has entered a complete deadlock.
    \item \textbf{Absolute maximum} (3,600\,s): a hard upper bound that terminates the simulation regardless of progress.
\end{enumerate}

When any timeout is triggered, the controller kills all MTA processes and reports the failure to the batch controller, which decides whether to retry.

\subsection{The Batch Controller}
\label{sec:batch_controller}

The Batch Controller manages the per-story lifecycle (Figure~\ref{fig:story_lifecycle}, Appendix~\ref{app:story_lifecycle}). For each story, it invokes the GEST generator (Section~\ref{sec:procedural_gen}), passes the resulting GEST to the MTA Controller for simulation, collects artifacts after completion, exports the proto-graph, generates the textual description (Section~\ref{sec:text_gen}), and optionally uploads the results to Google Drive.

Both generation and simulation failures trigger retries with exponential backoff (base delay multiplied by $2^{\text{attempt}}$, capped at a configurable maximum). The retry manager classifies errors as retriable (timeouts, validation failures, API errors) or fatal (configuration errors, missing dependencies), and only retries retriable errors. The batch controller persists its state to a JSON file after every significant operation --- story completion, phase transition, simulation result --- enabling the batch to resume from a checkpoint if the controller itself crashes or the host machine reboots.

After a successful simulation, the artifact collector flattens the MTA output directory structure (which uses internal GUIDs and spectator identifiers) into a clean per-camera folder layout, compresses spatial relation files and frame archives for efficient storage, and validates that no error markers are present. The batch controller then invokes the proto-graph exporter and text generation pipeline (Sections~\ref{sec:proto_language}--\ref{sec:text_integration}) before uploading the story to Google Drive.

\subsection{Distributed Monitoring and Recovery}
\label{sec:vm_monitoring}

When running multiple worker VMs in parallel, the VM Monitor on the host machine polls each worker's health every 30 seconds. It checks three signals: whether the VM process is still running (via VMware's command-line interface), whether the worker has written a completion marker file, and whether the worker's logs have been updated recently. If no log activity is detected for a configurable threshold (default one hour), the worker is marked as hung.

Hung or crashed workers are automatically restarted: the orchestrator stops the VM, starts it again from the same snapshot, and instructs the auto-runner to resume the batch from its last checkpoint (using the persisted state file). Restart attempts use exponential backoff and are limited to a configurable maximum. If a worker fails permanently (exceeding the restart limit), the orchestrator can spawn a replacement worker to handle the remaining stories, ensuring that the target corpus size is reached despite individual worker failures.

After all workers complete, the orchestrator merges the outputs from each worker's isolated directory into a single consolidated corpus, renumbers the stories sequentially, and generates a batch report summarizing success rates, retry statistics, and per-story metadata.

\subsection{Cloud Upload and Distribution}
\label{sec:cloud_upload}

The production system integrates with Google Drive for corpus distribution. Each worker VM uploads its completed stories to a dedicated subfolder in a shared Google Drive directory, using the Google Drive API with OAuth authentication. Uploads use resumable transfer for large files (videos, compressed archives), with automatic retry and exponential backoff on network errors. Each story is uploaded as a complete directory tree containing the GEST, video, artifact archives (compressed spatial relations, RGB frames, and segmentation masks), event-frame mappings, and textual descriptions.

Before upload, the artifact collector compresses bulky per-frame data: spatial relation JSON files, RGB screenshots, and segmentation PNGs are packed into ZIP archives within each camera folder, significantly reducing the number of individual files and the total upload time. Error simulations --- those containing error markers or timeout indicators --- are filtered out and excluded from upload, ensuring that only successful simulations reach the shared corpus.

After all workers finish uploading, the orchestrator merges the per-worker Google Drive folders into a single flat structure: all story folders are moved to the root of the parent folder, empty worker and batch subfolders are cleaned up, and the aggregated batch report and statistics are uploaded alongside the corpus. The result is a single shareable Google Drive folder containing the complete corpus ready for downstream use.

\section{Engineering Decisions and Lessons Learned}
\label{sec:engineering}

This section reflects on the key engineering decisions that shaped the system and the lessons learned over six years of continuous development.

\subsection{The Choice of GTA San Andreas and MTA}

Building a virtual world from scratch would have required years of asset creation, animation development, and physics implementation --- an infeasible effort for a single researcher. By building on GTA San Andreas via Multi Theft Auto, we inherited a vast library of pre-existing assets (733 objects, 2,500+ animations, 312 actor skins, 70+ contexts) and a mature, open-source scripting API maintained by a community for over two decades. The trade-off was accepting the constraints of a platform we did not control: no access to the game's navigation mesh (requiring our own pathfinding graphs), no programmatic camera API beyond basic position setting (requiring the spectator workaround), and rendering artifacts such as the light and shadow texture overlap that complicates instance segmentation (Section~\ref{sec:instance_seg}).

The platform's low hardware requirements proved essential for production-scale generation. Because GTA San Andreas runs on virtually any modern hardware without a dedicated GPU, we could run 25 parallel instances on a single workstation --- a density that would be impossible with a modern engine requiring GPU-accelerated rendering for each instance.

\subsection{Progressive Adapter Pattern}
\label{sec:future_gtav}

A deliberate architectural decision throughout the system has been the progressive separation of game-agnostic logic from engine-specific bindings. The artifact collection subsystem implements this cleanly: the collection manager, orchestration logic, and collector interfaces are fully game-agnostic, while MTA-specific implementations are isolated in dedicated adapter layers. The native C++ screenshot module similarly separates a reusable core from the MTA-specific Desktop Duplication backend. The remaining parts of the system --- particularly the action classes, location system, and actor handlers --- still contain MTA-specific code interleaved with the orchestration logic.

Completing the adapter pattern across all components would enable porting the system to a different 3D platform. The most natural target is GTA V via FiveM, which uses the same Lua scripting language as MTA and provides a substantially more modern rendering engine with higher-resolution textures, more realistic lighting, and a larger open world. Based on our experience, we estimate that a full port to FiveM would require approximately one month of full-time development, primarily implementing the adapter layer for the new platform's animation, object, and camera APIs without modifying the orchestration, planning, or collection logic. Beyond FiveM, the architecture supports any 3D platform with programmatic actor control, including Unity and Unreal Engine.

\appendix

\section{GEST JSON Schema Evolution}
\label{app:gest_json}

This appendix provides concrete JSON examples for each generation of the GEST schema. The examples are excerpted from actual graphs used in the project, illustrating how the technical representation evolved from a flat linked-list format to a structured specification with explicit temporal constraints.

\subsection*{Generation 1: Linked-List Format (2021)}
\label{app:gest_gen1}

The first-generation format used separate \texttt{Actor} and \texttt{Target} fields, UUID-based identifiers, and explicit \texttt{Next} pointers forming a singly-linked list per actor. The following excerpt shows a two-actor story (Leonel Carrillo and Kaiser Medrano) with actions in the living room and kitchen. Actor and object declarations use \texttt{Action: "Exists"}; action nodes reference actors and objects by UUID.

\begin{small}
\begin{verbatim}
[{
  "8448ec26-...": {
    "Actor": {"id": "8448ec26-...", "Gender": 1,
              "Name": "Leonel CARRILLO"},
    "Action": "Exists",
    "id": "8448ec26-..."
  },
  "bcfa74c6-...": {
    "Action": "Exists",
    "id": "bcfa74c6-...",
    "Target": {"id": "bcfa74c6-...", "Name": "cigarette"}
  },
  "64d4408f-...": {
    "Actor": {"id": "8448ec26-..."},
    "Action": "Smoke",
    "Next": "ee73fce4-...",
    "id": "64d4408f-...",
    "Target": {"id": "bcfa74c6-..."},
    "Location": "livingroom"
  },
  "ee73fce4-...": {
    "Actor": {"id": "8448ec26-..."},
    "Action": "SmokeOut",
    "Next": "d2f99c23-...",
    "id": "ee73fce4-...",
    "Target": {"id": "bcfa74c6-..."},
    "Location": "livingroom"
  },
  "d2f99c23-...": {
    "Actor": {"id": "8448ec26-..."},
    "Action": "Move",
    "Next": "b9025335-...",
    "id": "d2f99c23-...",
    "Target": {"id": "5ec4117a-..."},
    "Location": "kitchen"
  }
}]
\end{verbatim}
\end{small}

Key characteristics: UUIDs as node identifiers, \texttt{Actor.id} and \texttt{Target.id} as separate reference fields, sequencing via \texttt{Next} pointers on each node, no separate temporal section, the entire graph wrapped in a JSON array.

\subsection*{Generation 2: Unified Entity Format with Temporal Section (2021)}
\label{app:gest_gen2}

The second-generation format merged \texttt{Actor} and \texttt{Target} into an \texttt{Entities} array, moved attributes into a \texttt{Properties} dictionary, and introduced a top-level \texttt{temporal} section with per-actor action chains. The following example is a bAbI-derived graph \cite{weston2015towards} with four actors (Mary, John, Daniel, Sandra) performing movement actions.

\begin{small}
\begin{verbatim}
{
  "actor0": {
    "Action": "Exists",
    "Entities": ["actor0"],
    "Location": null,
    "Timeframe": null,
    "Properties": {"Gender": 2, "Name": "Mary"}
  },
  "actor1": {
    "Action": "Exists",
    "Entities": ["actor1"],
    "Location": null,
    "Timeframe": null,
    "Properties": {"Gender": 1, "Name": "John"}
  },
  "action0": {
    "Action": "Move",
    "Entities": ["actor0"],
    "Location": ["", "bathroom"],
    "Timeframe": null,
    "Properties": {}
  },
  "action1": {
    "Action": "Move",
    "Entities": ["actor1"],
    "Location": ["", "hallway"],
    "Timeframe": null,
    "Properties": {}
  },
  "action4": {
    "Action": "Move",
    "Entities": ["actor1"],
    "Location": ["hallway", "office"],
    "Timeframe": null,
    "Properties": {}
  },
  "temporal": {
    "starting_actions": {
      "actor0": "action0",
      "actor1": "action1",
      "actor2": "action2",
      "actor3": "action3"
    },
    "action0": {"relations": null, "next": "action6"},
    "action6": {"relations": null, "next": null},
    "action1": {"relations": null, "next": "action4"},
    "action4": {"relations": null, "next": "action8"},
    "action8": {"relations": null, "next": "action9"},
    "action9": {"relations": null, "next": null}
  }
}
\end{verbatim}
\end{small}

Key changes from Generation 1: human-readable identifiers, unified \texttt{Entities} array, \texttt{Properties} dictionary, \texttt{Location} as an array (source and destination for movement), and crucially, a separate \texttt{temporal} section with \texttt{starting\_actions} and per-action \texttt{next}/\texttt{relations} entries. The \texttt{timeline} field provides a human-readable summary of each actor's action sequence.

\subsection*{Generation 3: Temporal Constraint Graphs (2025--present)}
\label{app:gest_gen3}

The third-generation format retains the same event node structure but extends the temporal section with explicit inter-actor constraints (\texttt{before}, \texttt{after}, \texttt{starts\_with}) and adds reserved sections for \texttt{spatial}, \texttt{semantic}, and \texttt{camera} annotations. The following excerpt is from a complex six-actor, three-episode story (\texttt{c10\_sync.json}) coordinating actors across a bedroom, a kitchen, a porch, and a garden. We show event nodes for four actors alongside the temporal section that coordinates them.

\begin{small}
\begin{verbatim}
{
  "alice3": {
    "Action": "Exists", "Entities": ["alice3"],
    "Location": ["bedroom"],
    "Properties": {"Gender": 2, "Name": "Mary"}
  },
  "alice": {
    "Action": "Exists", "Entities": ["alice"],
    "Location": ["kitchen"],
    "Properties": {"Gender": 2, "Name": "Mary"}
  },
  "bob": {
    "Action": "Exists", "Entities": ["bob"],
    "Location": ["kitchen"],
    "Properties": {"Gender": 1, "Name": "Denzel"}
  },
  "m1": {
    "Action": "SitDown",
    "Entities": ["alice3", "officeChair"],
    "Location": ["bedroom"], "Properties": {}
  },
  "m10": {
    "Action": "Sleep",
    "Entities": ["alice3", "Bed"],
    "Location": ["bedroom"], "Properties": {}
  },
  "a1": {
    "Action": "SitDown",
    "Entities": ["alice", "chair2"],
    "Location": ["kitchen"], "Properties": {}
  },
  "b1": {
    "Action": "SitDown",
    "Entities": ["bob", "chair1"],
    "Location": ["kitchen"], "Properties": {}
  },
  "a6": {
    "Action": "Hug",
    "Entities": ["alice", "bob"],
    "Location": ["porch"], "Properties": {}
  },
  "b7": {
    "Action": "Move",
    "Entities": ["bob"],
    "Location": ["porch", "street"], "Properties": {}
  },

  "temporal": {
    "starting_actions": {
      "alice": "a1", "bob": "b1",
      "alice3": "m1", "bod3": "d1"
    },
    "m1": {
      "relations": ["m1_starts_with_d1"],
      "next": "m2"
    },
    "m1_starts_with_d1": {
      "type": "starts_with"
    },
    "m10": {
      "relations": null, "next": null
    },
    "m10_before_a1": {
      "source": "m10", "type": "before", "target": "a1"
    },
    "a1": {
      "relations": ["a1_starts_with_b1",
                     "a1_after_m10", "a1_after_d8"],
      "next": "a2"
    },
    "a1_starts_with_b1": {
      "type": "starts_with"
    },
    "b1": {
      "relations": ["a1_starts_with_b1",
                     "b1_after_m10", "b1_after_d8"],
      "next": "b2"
    },
    "b1_after_m10": {
      "source": "b1", "type": "after", "target": "m10"
    },
    "a6": {
      "relations": ["tm0", "a6_before_b7"],
      "next": "a6.1"
    },
    "a6_before_b7": {
      "source": "a6", "type": "before", "target": "b7"
    },
    "b7": {
      "relations": ["b7_before_j1", "b7_before_n1"],
      "next": null
    },
    "tm0": {
      "type": "starts_with"
    }
  },
  "spatial": {},
  "semantic": {},
  "camera": {}
}
\end{verbatim}
\end{small}

This excerpt illustrates the key additions in Generation 3. The story begins with two actors (alice3 and bod3) in the bedroom, whose first actions start simultaneously via \texttt{m1\_starts\_with\_d1}. After alice3 finishes sleeping (\texttt{m10}), two new actors (alice and bob) begin their sequences in the kitchen --- the \texttt{before} constraint \texttt{m10\_before\_a1} and the \texttt{after} constraint \texttt{b1\_after\_m10} ensure that the kitchen scene cannot start until the bedroom scene is complete. Alice and bob also start simultaneously via \texttt{a1\_starts\_with\_b1}. Later, alice hugs bob on the porch (\texttt{a6}), and the \texttt{before} constraint \texttt{a6\_before\_b7} ensures the hug completes before bob moves to the street. Multiple actors reference shared constraint identifiers (e.g., both \texttt{a1} and \texttt{b1} list \texttt{a1\_starts\_with\_b1} in their \texttt{relations}). Reserved top-level sections for \texttt{spatial}, \texttt{semantic}, and \texttt{camera} provide extension points. This format enables the Floyd-Warshall-based temporal orchestration described in Section~\ref{sec:stage_orchestration}.

\section{Supertemplate Inventory}
\label{app:supertemplates}

Table~\ref{tab:supertemplates} lists all supertemplates configured in the GEST-Engine, along with the number of visual template variants available for each object category. Each template represents a distinct 3D object model with its own POI configuration, spatial offsets, and action chains. When a GEST graph references an object category (e.g., ``Bed''), the system randomly selects one of the available templates at runtime, producing visual variability across different executions of the same graph.

\begin{table}[htbp]
\centering
\caption{Complete supertemplate inventory. Each supertemplate groups multiple visual variants of the same object category. The total of 124 templates across 20 supertemplates provides substantial visual variability.}
\label{tab:supertemplates}
\begin{tabular}{@{}llr@{}}
\toprule
\textbf{Supertemplate} & \textbf{Category} & \textbf{Templates} \\
\midrule
TV & Entertainment & 25 \\
Bed & Furniture & 20 \\
Sofa & Furniture & 19 \\
Armchair & Furniture & 8 \\
Chairs & Furniture & 8 \\
Office Desk & Furniture & 6 \\
Table & Furniture & 6 \\
Sinks & Appliance & 6 \\
Music Player & Entertainment & 4 \\
Chairs and Table & Furniture & 3 \\
Drinks & Consumable & 2 \\
Food & Consumable & 2 \\
Gym Bike & Exercise & 2 \\
Treadmill & Exercise & 2 \\
Dumbbells & Exercise & 1 \\
Interactions & Social & 2 \\
Smoke & Personal & 2 \\
Talk Phone & Personal & 2 \\
Tai Chi & Exercise & 2 \\
\midrule
\textbf{Total} & & \textbf{124} \\
\bottomrule
\end{tabular}
\end{table}


\section{Text Generation Pipeline --- Complete Example}
\label{app:text_gen_example}

This appendix provides a complete example of the text generation pipeline described in Section~\ref{sec:text_gen}, showing the full chain from the output GEST, through the proto-graph transformation, to the proto-language, and finally to the LLM-refined description. The example is a five-actor story set in a garden and driveway.

\subsection*{Step 1: Output GEST (Excerpt)}

After simulation, the engine produces an output GEST with 5 actors, 5 spawnable objects, and 38 action events across two regions. The \texttt{Timeframe} fields are \texttt{null} at this stage --- they will be populated in the next step. Below is an excerpt:

\begin{small}
\begin{verbatim}
"a0": {"Action": "Exists", "Entities": ["a0"],
       "Location": ["garden"], "Timeframe": null,
       "Properties": {"Name": "Actor_a0", "Gender": 2}},
"spawn_phone_a0": {"Action": "Exists",
       "Entities": ["spawn_phone_a0"],
       "Location": null, "Timeframe": null,
       "Properties": {"Type": "MobilePhone"}},

"a0_1": {"Action": "TakeOut",
         "Entities": ["a0", "spawn_phone_a0"],
         "Location": ["garden"], "Timeframe": null,
         "Properties": {}},
"a0_2": {"Action": "AnswerPhone",
         "Entities": ["a0", "spawn_phone_a0"],
         "Location": ["garden"], "Timeframe": null,
         "Properties": {}},
...
\end{verbatim}
\end{small}

\subsection*{Step 2: Proto-Graph Transformation}

The proto-graph exporter combines the output GEST with the event-frame mapping captured during simulation (Section~\ref{sec:temporal_alignment}) and applies three transformations to produce the proto-graph:

\begin{enumerate}
    \item \textbf{ID normalization}: internal entity identifiers are converted to a canonical format compatible with the text generation system. Actor IDs such as \texttt{a0}, \texttt{a1} become \texttt{actor0}, \texttt{actor1}. Action event IDs such as \texttt{a0\_1} become \texttt{action0\_1}. Spawnable object IDs such as \texttt{spawn\_phone\_a0} become sequential \texttt{object0}, \texttt{object1}, etc.

    \item \textbf{Timeframe population}: the \texttt{Timeframe} field of each action event, which is \texttt{null} in the input GEST, is populated with the exact start and end frame numbers from the event-frame mapping. For example, event \texttt{a0\_1} (TakeOut) which was captured at frames 29--59 receives \texttt{Timeframe: "29-59"}.

    \item \textbf{Object type annotation}: the \texttt{Type} property of object \texttt{Exists} nodes is reformatted with instance counters for disambiguation. \texttt{"Type": "MobilePhone"} becomes \texttt{"Type": "id:0.0-class:mobilephone"}, enabling the text generator to distinguish multiple objects of the same type.
\end{enumerate}

The same event after proto-graph transformation:

\begin{small}
\begin{verbatim}
"action0_1": {"Action": "TakeOut",
              "Entities": ["actor0", "object0"],
              "Location": ["garden"],
              "Timeframe": "29-59",
              "Properties": {}},
\end{verbatim}
\end{small}

The temporal section is preserved with transformed IDs (e.g., \texttt{starting\_actions: \{"actor0": "action0\_1", ...\}}), and the spatial, semantic, and camera sections are copied verbatim.

\subsection*{Step 3: Proto-Language Generation and LLM Prompt}

The proto-graph is converted to proto-language by iterating through each actor's event chain and assembling atomic sentences from the action's verb form, the involved entities, and temporal connectors. The proto-language is then wrapped in a prompt for the LLM, including rewriting instructions and a one-shot example. The full prompt sent to GPT-4o is shown below:

\begin{small}
\begin{quote}
Rewrite the following paragraph into a more natural, concise and accurate text. The text should be plausible and should not contain unrealistic information. Ensure that you preserve all actions (don't delete actions) and all relations (e.g.\ temporal, spatial order) between them. Consider the following example:

\textbf{Initial Paragraph:} In the garden the first man takeouts a cigarette. After that, the first man smokeins a cigarette. After that, the first man smokeouts a cigarette. After that, the first man and the second man talk to each other. After that, the second man moves from garden to classroom at the same time also the first man moves from garden to classroom. After that the second man pickups a drink. After that, the second man gives the first man a drink. After that, the first man drink a drink. After that, the first man and the second man hug each other.

\textbf{Rewritten paragraph:} In the garden, the first man takes out a cigarette, lights it, smokes it, and then puts it out. After that, the first man and the second man talk to each other. Then, at the same time, both men move from the garden to the classroom. Once there, the second man picks up a drink and gives it to the first man. The first man drinks it, and afterward, the two men hug each other.

\textbf{Initial Paragraph:} In the garden the first man taichis. Meanwhile, the first woman takeouts a mobilephone. Meanwhile, the second woman takeouts a cigarette. Meanwhile, the second man takeouts a mobilephone. After that, the second woman smokeins a cigarette. After that, the first woman answerphones a mobilephone. After that, the second man answerphones a mobilephone. After that, the second man talkphones a mobilephone. After that, the first woman talkphones a mobilephone. After that, the second woman smokes a cigarette. After that, the second woman smokeouts a cigarette. After that, the first woman hangups a mobilephone. After that, the second man hangups a mobilephone. After that, the first woman stashs a mobilephone. After that, the second woman stashs a cigarette. After that, the second man stashs a mobilephone. After that, the second man and the first woman hug to each other. After that, the second woman and the first man talk to each other. After that, the second woman taichis. After that, the second man and the second woman talk to each other. After that, the first man moves from garden to driveway. After that, in the driveway the first man takeouts a mobilephone. At the same time, the third woman takeouts a mobilephone. After that, the first man answerphones a mobilephone. After that, the third woman answerphones a mobilephone. After that, the first man talkphones a mobilephone. After that, the third woman talkphones a mobilephone. After that, the first man hangups a mobilephone. After that, the third woman hangups a mobilephone. After that, the first man stashs a mobilephone. After that, the third woman stashs a mobilephone. After that, the third woman and the first man laugh to each other. After that, the third woman and the first man hug to each other.
\end{quote}
\end{small}

Note the characteristic features of the proto-language: ungrammatical verb forms (\textit{takeouts}, \textit{smokeins}, \textit{answerphones}), explicit temporal connectors (\textit{After that}, \textit{Meanwhile}, \textit{At the same time}), region-level location references (\textit{from garden to driveway}, \textit{in the driveway}), and gender-aware actor references (\textit{the first woman}, \textit{the second man}).

\subsection*{Step 4: LLM-Refined Output (GPT-4o)}

The refined output produced by GPT-4o at temperature 1.0:

\begin{small}
\begin{quote}
In the garden, the first man practices tai chi while the first woman takes out a mobile phone, the second woman takes out a cigarette, and the second man takes out his mobile phone. The second woman lights and smokes the cigarette. Meanwhile, the first woman and the second man both answer their phones and start talking. The second woman continues to smoke and then puts out her cigarette. After finishing their calls, the first woman and the second man hang up and put their phones away. The second woman also puts away her cigarette. The second man and the first woman hug, while the second woman talks to the first man, and then practices tai chi. The second man and the second woman then have a conversation. The first man moves to the driveway, where he takes out his mobile phone. At the same time, the third woman also takes out her phone. They both answer their phones and have a conversation. After finishing their calls, they hang up, put their phones away, laugh together, and then share a hug.
\end{quote}
\end{small}

The LLM preserves all 32 actions and their temporal ordering while producing fluent, natural prose. Mechanical action names are replaced with natural language equivalents (\textit{takeouts a cigarette} $\to$ \textit{takes out a cigarette}; \textit{smokeins} $\to$ \textit{lights}; \textit{smokeouts} $\to$ \textit{puts out}; \textit{stashs} $\to$ \textit{puts away}; \textit{taichis} $\to$ \textit{practices tai chi}). Repetitive sequential actions are merged into compound sentences (\textit{they hang up, put their phones away, laugh together, and then share a hug}). Gender-appropriate pronouns (\textit{he}, \textit{she}, \textit{his}, \textit{her}) are used throughout, consistent with the actor genders specified in the GEST.


\section{Story Production Lifecycle}
\label{app:story_lifecycle}

\begin{figure}[p]
\centering
\includegraphics[height=0.95\textheight]{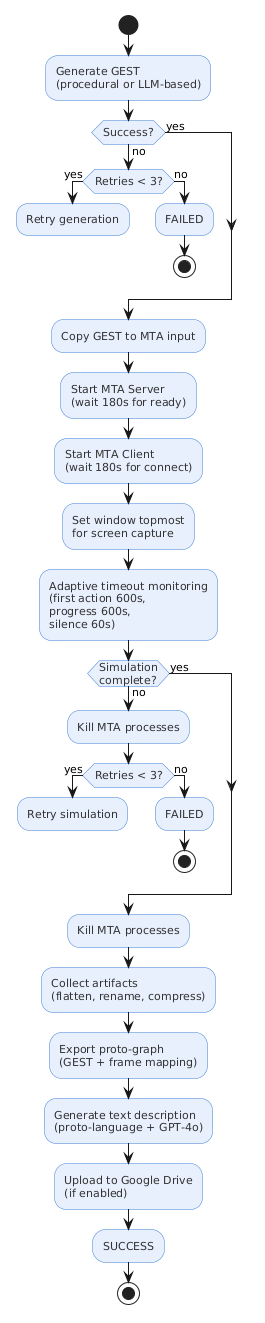}
\caption{\textbf{Per-story production lifecycle} (Section~\ref{sec:production}).}
\label{fig:story_lifecycle}
\end{figure}

This appendix provides a detailed walkthrough of the per-story production lifecycle illustrated in Figure~\ref{fig:story_lifecycle} and described in Section~\ref{sec:production}.

\subsection*{Step-by-Step Narration}

\textbf{1. GEST Generation.} The lifecycle begins with the generation of a GEST specification, either through the procedural random generator (Section~\ref{sec:procedural_gen}) or a companion LLM-based approach. The procedural generator reads the capability registry (Section~\ref{sec:capability_extraction}), selects episodes, creates actors, and builds action chains that are guaranteed to be executable by construction. If generation fails --- for instance, because the LLM produces an invalid graph or an API call times out --- the system retries up to three times with exponential backoff. If all retries are exhausted, the story is marked as failed and the batch controller moves to the next story.

\textbf{2. Copy GEST to MTA input.} The generated GEST is copied to the MTA server's input directory, where the engine will read it on startup. The batch controller also writes the engine's configuration file, specifying the input graph path, whether artifact collection is enabled, and whether segmentation capture is active.

\textbf{3. Start MTA Server.} The MTA Controller launches the MTA server executable as a subprocess. It then monitors the server's log output, waiting for the ``server started and is ready to accept connections'' message. If this message does not appear within 180 seconds, the server is assumed to have failed during startup (e.g., due to a corrupted configuration or missing resource files), and the simulation attempt is aborted.

\textbf{4. Start MTA Client.} Once the server is ready, the MTA Controller launches the MTA client, which connects to the local server. The controller waits for the client connection to appear in the server log. If the client does not connect within 180 seconds, the attempt is aborted --- this failure mode typically occurs when a UI dialog (such as a first-launch setup screen or an update prompt) blocks the client from connecting automatically.

\textbf{5. Set window topmost.} After the client connects, the controller sets the MTA game window to always-on-top using the Windows API. This is required because the Desktop Duplication API (Section~\ref{sec:rgb_capture}) captures the screen content at the GPU level, and the game window must be visible and unobstructed for the captured frames to contain the simulation rather than whatever window happens to be in front.

\textbf{6. Adaptive timeout monitoring.} The controller enters a polling loop that checks the simulation's health every two seconds. It monitors three signals in parallel: process health (are the server and client still running?), window responsiveness (is the game window responding to messages?), and log-based progress (are new actions being dispatched and completed?). Four adaptive timeouts guard against different failure modes: the first action must appear within 600 seconds of client connection; subsequent actions must appear within 600 seconds of the previous action (this timeout resets on each detected action, adapting to stories of varying length); total log silence must not exceed 60 seconds; and the absolute maximum simulation time is 3,600 seconds.

\textbf{7. Simulation complete or failed.} If the monitoring detects a completion signal in the logs, the simulation is marked as successful. If any timeout is triggered, or if the server or client process terminates unexpectedly, or if the game window becomes unresponsive for more than 180 seconds, the simulation is marked as failed. In either case, the controller kills all MTA processes (server, client, and any orphaned instances) to ensure a clean state for the next attempt.

\textbf{8. Retry on failure.} If the simulation failed, the system retries up to three times. Each retry starts from step 2 (copying the GEST and launching MTA from scratch). Retry timeouts may be configured differently from the first attempt --- for instance, using a longer absolute maximum on retries for stories that are known to be complex. If all retries are exhausted, the story is marked as failed.

\textbf{9. Collect artifacts.} After a successful simulation, the artifact collector processes the MTA output directory. The engine produces output in a UUID-named folder containing spectator subfolders; the collector flattens this structure into a clean layout with camera folders, copies the server and client logs for debugging, and checks for error markers (ERROR files or MAX\_STORY\_TIME\_EXCEEDED files). If an error marker is found despite the simulation reporting success, the success is overridden to a failure --- the error marker indicates that the engine detected an internal inconsistency during execution.

\textbf{10. Export proto-graph.} The proto-graph exporter (Section~\ref{sec:proto_language}) combines the output GEST with the event-frame mapping captured during simulation, normalizes entity identifiers, and populates each event's timeframe with the exact start and end frame numbers. The resulting proto-graph is saved alongside the simulation artifacts.

\textbf{11. Generate text description.} The text generation pipeline (Section~\ref{sec:text_gen}) converts the proto-graph to proto-language, wraps it in the LLM prompt with location context, and (if full mode is enabled) calls GPT-4o to produce the refined natural language description. The engine-generated text from the Logger (Section~\ref{sec:logger}) is also preserved alongside the LLM-refined description.

\textbf{12. Upload to Google Drive.} If cloud upload is enabled, the completed story --- including the GEST, video, compressed artifact archives, event-frame mapping, spatial relations, and textual descriptions --- is uploaded to Google Drive using resumable transfers. Error simulations are filtered out before upload.

\textbf{13. Mark success.} The story is marked as successful in the batch state, and the controller proceeds to the next story. The batch state is persisted to disk after each story completes, enabling the batch to resume from this point if interrupted.

\section*{Acknowledgements}
This work was supported by B\"uchi Labortechnik AG and by the project
``Romanian Hub for Artificial Intelligence --- HRIA'', Smart Growth,
Digitization and Financial Instruments Program, 2021--2027, MySMIS no.~351416.

\printbibliography

@article{ha2018worldmodels,
  title={World models},
  author={Ha, David and Schmidhuber, J{\"u}rgen},
  journal={arXiv preprint arXiv:1803.10122},
  volume={2},
  number={3},
  year={2018}
}

@article{hafner2023dreamerv3,
  title={Mastering diverse domains through world models},
  author={Hafner, Danijar and Pasukonis, Jurgis and Ba, Jimmy and Lillicrap, Timothy},
  journal={arXiv preprint arXiv:2301.04104},
  year={2023}
}

@inproceedings{bruce2024genie,
  title={Genie: Generative interactive environments},
  author={Bruce, Jake and Dennis, Michael D and Edwards, Ashley and Parker-Holder, Jack and Shi, Yuge and Hughes, Edward and Lai, Matthew and Mavalankar, Aditi and Steigerwald, Richie and Apps, Chris and others},
  booktitle={Forty-first International Conference on Machine Learning},
  year={2024}
}

@article{brooks2024sora,
  title={Video generation models as world simulators},
  author={Tim Brooks and Bill Peebles and Connor Holmes and Will DePue and Yufei Guo and Li Jing and David Schnurr and Joe Taylor and Troy Luhman and Eric Luhman and Clarence Ng and Ricky Wang and Aditya Ramesh},
  year={2024},
  url={https://openai.com/research/video-generation-models-as-world-simulators},
}

@article{veo3modelcard,
title={Veo 3 Model Card},
year={2025},
author={Google},
url={https://storage.googleapis.com/deepmind-media/Model-Cards/Veo-3-Model-Card.pdf},
note={Accessed: March 04, 2026}
}

@article{liu2025worldsurvey,
  title={Understanding world or predicting future? a comprehensive survey of world models},
  author={Ding, Jingtao and Zhang, Yunke and Shang, Yu and Zhang, Yuheng and Zong, Zefang and Feng, Jie and Yuan, Yuan and Su, Hongyuan and Li, Nian and Sukiennik, Nicholas and others},
  journal={ACM Computing Surveys},
  volume={58},
  number={3},
  pages={1--38},
  year={2025},
  publisher={ACM New York, NY}
}

@article{lecun2022path,
  title={A path towards autonomous machine intelligence version 0.9. 2, 2022-06-27},
  author={LeCun, Yann},
  journal={Open Review},
  volume={62},
  number={1},
  pages={1--62},
  year={2022}
}

@inproceedings{dosovitskiy2017carla,
  title={CARLA: An open urban driving simulator},
  author={Dosovitskiy, Alexey and Ros, German and Codevilla, Felipe and Lopez, Antonio and Koltun, Vladlen},
  booktitle={Conference on robot learning},
  pages={1--16},
  year={2017},
  organization={PMLR}
}

@inproceedings{black2023bedlam,
  title={Bedlam: A synthetic dataset of bodies exhibiting detailed lifelike animated motion},
  author={Black, Michael J and Patel, Priyanka and Tesch, Joachim and Yang, Jinlong},
  booktitle={Proceedings of the IEEE/CVF Conference on Computer Vision and Pattern Recognition},
  pages={8726--8737},
  year={2023}
}

@article{kolve2017ai2,
  title={Ai2-thor: An interactive 3d environment for visual ai},
  author={Kolve, Eric and Mottaghi, Roozbeh and Han, Winson and VanderBilt, Eli and Weihs, Luca and Herrasti, Alvaro and Gordon, Daniel and Zhu, Yuke and Gupta, Abhinav and Farhadi, Ali},
  journal={arXiv preprint arXiv:1712.05474},
  year={2017}
}

@inproceedings{puig2018virtualhome,
  title={Virtualhome: Simulating household activities via programs},
  author={Puig, Xavier and Ra, Kevin and Boben, Marko and Li, Jiaman and Wang, Tingwu and Fidler, Sanja and Torralba, Antonio},
  booktitle={Proceedings of the IEEE conference on computer vision and pattern recognition},
  pages={8494--8502},
  year={2018}
}

@inproceedings{qiu2023virtualhome,
  title={Virtualhome action genome: A simulated spatio-temporal scene graph dataset with consistent relationship labels},
  author={Qiu, Yue and Nagasaki, Yoshiki and Hara, Kensho and Kataoka, Hirokatsu and Suzuki, Ryota and Iwata, Kenji and Satoh, Yutaka},
  booktitle={Proceedings of the IEEE/CVF Winter Conference on Applications of Computer Vision},
  pages={3351--3360},
  year={2023}
}

@inproceedings{richter2016playing,
  title={Playing for data: Ground truth from computer games},
  author={Richter, Stephan R and Vineet, Vibhav and Roth, Stefan and Koltun, Vladlen},
  booktitle={European conference on computer vision},
  pages={102--118},
  year={2016},
  organization={Springer}
}

@article{jia2025evolutionary,
  title={The evolutionary disruption: A paradigm shift in film and animation industry driven by real-time rendering and virtual production},
  author={Jia, Xuguang and Berry, Adam and Johnston, Andrew},
  journal={Convergence},
  pages={13548565251356932},
  year={2025},
  publisher={SAGE Publications Sage UK: London, England}
}

@inproceedings{masala2023gest,
  title={Explaining vision and language through graphs of events in space and time},
  author={Masala, Mihai and Cudlenco, Nicolae and Rebedea, Traian and Leordeanu, Marius},
  booktitle={Proceedings of the IEEE/CVF International Conference on Computer Vision},
  pages={2826--2831},
  year={2023}
}

@article{allen1983maintaining,
  title={Maintaining knowledge about temporal intervals},
  author={Allen, James F},
  journal={Communications of the ACM},
  volume={26},
  number={11},
  pages={832--843},
  year={1983},
  publisher={ACM New York, NY, USA}
}

@article{floyd1962algorithm,
  title={Algorithm 97: shortest path},
  author={Floyd, Robert W},
  journal={Communications of the ACM},
  volume={5},
  number={6},
  pages={345--345},
  year={1962},
  publisher={ACM New York, NY, USA}
}

@article{warshall1962theorem,
  title={A theorem on boolean matrices},
  author={Warshall, Stephen},
  journal={Journal of the ACM (JACM)},
  volume={9},
  number={1},
  pages={11--12},
  year={1962},
  publisher={ACM New York, NY, USA}
}

@article{masala2025vision,
  title={From Vision To Language through Graph of Events in Space and Time: An Explainable Self-supervised Approach},
  author={Masala, Mihai and Leordeanu, Marius},
  journal={arXiv preprint arXiv:2507.04815},
  year={2025}
}

@article{weston2015towards,
  title={Towards ai-complete question answering: A set of prerequisite toy tasks},
  author={Weston, Jason and Bordes, Antoine and Chopra, Sumit and Rush, Alexander M and van Merri{\"e}nboer, Bart and Joulin, Armand and Mikolov, Tomas},
  journal={arXiv preprint arXiv:1502.05698},
  year={2015}
}

@inproceedings{varol2017learning,
  title={Learning from synthetic humans},
  author={Varol, Gul and Romero, Javier and Martin, Xavier and Mahmood, Naureen and Black, Michael J and Laptev, Ivan and Schmid, Cordelia},
  booktitle={Proceedings of the IEEE conference on computer vision and pattern recognition},
  pages={109--117},
  year={2017}
}

@inproceedings{lin2014microsoft,
  title={Microsoft coco: Common objects in context},
  author={Lin, Tsung-Yi and Maire, Michael and Belongie, Serge and Hays, James and Perona, Pietro and Ramanan, Deva and Doll{\'a}r, Piotr and Zitnick, C Lawrence},
  booktitle={European conference on computer vision},
  pages={740--755},
  year={2014},
  organization={Springer}
}

@article{SMPL:2015,
      author = {Loper, Matthew and Mahmood, Naureen and Romero, Javier and Pons-Moll, Gerard and Black, Michael J.},
      title = {{SMPL}: A Skinned Multi-Person Linear Model},
      journal = {ACM Trans. Graphics (Proc. SIGGRAPH Asia)},
      month = oct,
      number = {6},
      pages = {248:1--248:16},
      publisher = {ACM},
      volume = {34},
      year = {2015}
    }

@article{hu2024storyagent,
  title={Storyagent: Customized storytelling video generation via multi-agent collaboration},
  author={Hu, Panwen and Jiang, Jin and Chen, Jianqi and Han, Mingfei and Liao, Shengcai and Chang, Xiaojun and Liang, Xiaodan},
  journal={arXiv preprint arXiv:2411.04925},
  year={2024}
}

@inproceedings{wang2026mavis,
  title={MAViS: A multi-agent framework for long-sequence video storytelling},
  author={Wang, Qian and Huang, Ziqi and Jia, Ruoxi and Debevec, Paul and Yu, Ning},
  booktitle={Proceedings of the 19th Conference of the European Chapter of the Association for Computational Linguistics (Volume 1: Long Papers)},
  pages={2273--2295},
  year={2026}
}

@inproceedings{wang2026dreamrunner,
  title={Dreamrunner: Fine-grained compositional story-to-video generation with retrieval-augmented motion adaptation},
  author={Wang, Zun and Li, Jialu and Lin, Han and Yoon, Jaehong and Bansal, Mohit},
  booktitle={Proceedings of the AAAI Conference on Artificial Intelligence},
  volume={40},
  number={13},
  pages={10503--10511},
  year={2026}
}

@article{wan2025wan,
  title={Wan: Open and advanced large-scale video generative models},
  author={Wan, Team and Wang, Ang and Ai, Baole and Wen, Bin and Mao, Chaojie and Xie, Chen-Wei and Chen, Di and Yu, Feiwu and Zhao, Haiming and Yang, Jianxiao and others},
  journal={arXiv preprint arXiv:2503.20314},
  year={2025}
}

\end{document}